\pdfoutput=1
\RequirePackage{fix-cm}
\documentclass[10pt,twocolumn,letterpaper]{article}

\usepackage[pagenumbers]{cvpr}

\usepackage{cuted}      
\usepackage{currfile}   
\usepackage{caption}    
\usepackage{microtype}   
\usepackage{xspace}
\usepackage{pifont}   
\newcommand{\paper}{MatNWM\xspace}
\newcommand{\mono}{Monolithic\xspace}   
\DeclareRobustCommand{\M}[1]{\ensuremath{\mathrm{M}_{#1}}}   
\DeclareRobustCommand{\at}[2]{\ensuremath{\mathrm{M}_{#1}{@}#2}}  
\DeclareRobustCommand{\monoat}[1]{\ensuremath{\mathrm{Mono}{@}#1}}
\newcommand{\recon}{RECON\xspace}
\newcommand{\sacson}{SACSoN\xspace}
\newcommand{\scand}{SCAND\xspace}
\newcommand{\lpips}{LPIPS\xspace}
\newcommand{\fid}{FID\xspace}
\newcommand{\dino}{DINO-dist\xspace}
\newcommand{\ate}{ATE\xspace}
\newcommand{\rpe}{RPE\xspace}
\newcommand{\cem}{CEM\xspace}
\newcommand{\nwm}{NWM\xspace}
\newcommand{\raenwm}{RAE-NWM\xspace}
\newcommand{\bestval}[1]{\textbf{#1}}
\newcommand{\secval}[1]{\underline{#1}}

\newcommand{\sig}{\textsuperscript{*}}   
\newcommand{\tsig}{\rlap{\textsuperscript{*}}}   
\newcommand{\tdag}{\rlap{$^{\dagger}$}}          

\usepackage{amsmath,amssymb,bm}

\newcommand{\vel}{\bm{v}_\theta}
\newcommand{\R}{\mathbb{R}}

\usepackage{booktabs}
\usepackage{multirow}
\usepackage{array}
\usepackage{colortbl}
\newcolumntype{C}{>{\centering\arraybackslash}X}
\usepackage{tabularx}

\newcommand{\dn}{$\downarrow$}

\usepackage{pgfplots}
\pgfplotsset{compat=1.18}
\usepgfplotslibrary{groupplots}
\usetikzlibrary{arrows.meta,positioning,calc,fit,backgrounds,shapes.geometric,decorations.pathreplacing}
\pgfplotsset{
  every axis/.append style={font=\footnotesize, line width=0.6pt, tick style={line width=0.4pt}},
  every axis label/.append style={font=\footnotesize},
  legend style={font=\scriptsize, draw=none, fill=none},
  grid style={black!12},
}
\definecolor{cmono}{RGB}{90,90,90}
\definecolor{cM4}{RGB}{31,119,180}
\definecolor{cM1}{RGB}{214,39,40}
\definecolor{cTwo}{RGB}{148,103,189}
\colorlet{cM2}{cM1!66!cM4}   
\colorlet{cM3}{cM1!33!cM4}
\pgfplotscreateplotcyclelist{models}{
  {cmono, mark=square*, mark size=1.6pt},
  {cM4, mark=*, mark size=1.7pt},
  {cM1, mark=triangle*, mark size=2.1pt},
}

\AtEndPreamble{%
  \crefname{section}{Section}{Sections}%
  \crefname{table}{Table}{Tables}%
}

\definecolor{beatbase}{HTML}{DCEFD9}

\makeatletter\let\tabinput\@@input\makeatother
\providecommand{\ci}[1]{{\tiny[#1]}}
\newcommand{\RolloutT}{1.33}
\newcommand{\RolloutG}{0.035}
\newcommand{\RolloutS}{0.30}
\newcommand{\RolloutPage}[1]{%
\begin{tikzpicture}[x=1cm, y=1cm, every node/.style={inner sep=0pt, outer sep=0pt}]
\pgfmathsetmacro{\RolloutPW}{\RolloutS+4*\RolloutT+3*\RolloutG+0.12}
\pgfmathsetmacro{\RolloutPH}{6*\RolloutT+5*\RolloutG+0.72}
\foreach \pnl/\px/\py in {a/0/0, b/1/0, c/2/0, d/0/1, e/1/1, f/2/1}{
  \pgfmathsetmacro{\xo}{\px*\RolloutPW}
  \pgfmathsetmacro{\yo}{-\py*\RolloutPH}
  \pgfmathsetmacro{\xt}{\xo+\RolloutS}
  \node[anchor=base west, font=\footnotesize\bfseries] at (\xo, \yo+0.10) {(\pnl)};
  \foreach \c/\lab in {0/{0.25\,s}, 1/{4\,s}, 2/{8\,s}, 3/{16\,s}}{
    \node[anchor=base, font=\scriptsize] at ({\xt+\c*(\RolloutT+\RolloutG)+0.5*\RolloutT}, \yo+0.10) {\lab};
  }
  \node[rotate=90, anchor=south, font=\scriptsize] at ({\xt-0.07}, {\yo-0.5*\RolloutT}) {Ground truth};
  \foreach \c/\col in {0/t0, 1/t4, 2/t8, 3/t16}{
    \node[anchor=north west] at ({\xt+\c*(\RolloutT+\RolloutG)}, \yo)
      {\includegraphics[width=\RolloutT cm, height=\RolloutT cm]{fig_supp/rollout/#1_\pnl_gt_\col.jpg}};
  }
  \foreach \r/\key/\la/\lb/\lc in {1/m4roll/{\M{4}}/rollout/{49 NFEs}, 2/m1roll/{\M{1}}/rollout/{49 NFEs},
                                   3/m1d49/{\M{1}}/direct/{49 NFEs}, 4/m1d4/{\M{1}}/direct/{4 NFEs}, 5/m1d2/{\M{1}}/direct/{2 NFEs}}{
    \pgfmathsetmacro{\yr}{\yo-\r*(\RolloutT+\RolloutG)}
    \node[anchor=east, align=right, font=\fontsize{6.6}{7.6}\selectfont] at ({\xt+\RolloutT-0.10}, {\yr-0.5*\RolloutT}) {\la\\\lb\\\lc};
    \foreach \c/\col in {1/t4, 2/t8, 3/t16}{
      \node[anchor=north west] at ({\xt+\c*(\RolloutT+\RolloutG)}, \yr)
        {\includegraphics[width=\RolloutT cm, height=\RolloutT cm]{fig_supp/rollout/#1_\pnl_\key_\col.jpg}};
    }
  }
}
\end{tikzpicture}}

\definecolor{cvprblue}{rgb}{0.21,0.49,0.74}
\usepackage[pagebackref,breaklinks,colorlinks,allcolors=cvprblue]{hyperref}
\hypersetup{
  pdftitle={Fewer Steps, Better Plans: Sampling Variance in Flow-Matching Navigation World Models},
  pdfauthor={Po-Heng Chen, Maks Ovsjanikov, Leonidas Guibas, Chi-Que Pan, Ming-Hsuan Yang}
}

\newcommand{\suppline}[1]{\noindent\hyperref[#1]{\textbf{\ref*{#1}}\hspace{1em}\nameref*{#1}}\,\dotfill\,\pageref{#1}\par}

\title{Fewer Steps, Better Plans: Sampling Variance in Flow-Matching Navigation World Models}

\author{
Po-Heng Chen\textsuperscript{1}\quad
Maks Ovsjanikov\textsuperscript{2}\quad
Leonidas Guibas\textsuperscript{2}\quad
Chi-Que Pan\textsuperscript{1}\quad
Ming-Hsuan Yang\textsuperscript{2}\\[3pt]
{\normalsize \textsuperscript{1}Google \qquad \textsuperscript{2}Google DeepMind}
}

\makeatletter
\patchcmd{\@maketitle}{\vskip .375in}{\vskip .16in}{}{}
\patchcmd{\@maketitle}{\vspace*{24pt}}{\vspace*{6pt}}{}{}
\patchcmd{\@maketitle}{\vspace*{12pt}}{\vspace*{-6pt}}{}{}
\patchcmd{\abstract}{\vspace*{12pt}}{\vspace*{5pt}}{}{}
\makeatother

\begin{document}
\maketitle
\definecolor{m1col}{HTML}{E06666}
\definecolor{m2col}{HTML}{E69138}
\definecolor{m3col}{HTML}{F6B26B}
\definecolor{m4col}{HTML}{FFD966}
\begin{strip}
\noindent\begin{minipage}[b]{0.662\textwidth}
\noindent\resizebox{\linewidth}{!}{%
\begin{tikzpicture}[x=1cm, y=1cm,
  tile/.style={inner sep=0, outer sep=0, draw=black!55, line width=0.35pt},
  tag/.style={font=\sffamily\fontsize{5.8}{6.5}\selectfont, fill=white, fill opacity=0.92, text opacity=1, inner sep=1.2pt, outer sep=0, anchor=south west, draw=black!35, line width=0.2pt},
  farr/.style={-{Latex[length=1.45mm, width=1.25mm]}, line width=0.5pt, black!75},
  rowlab/.style={font=\sffamily\bfseries\scriptsize, text=black!85, inner sep=0},
]
\def\T{1.70}
\def\G{0.17}
\def\yt{0.94}
\node[tile] (in) at (0,0) {\includegraphics[width=\T cm, height=\T cm]{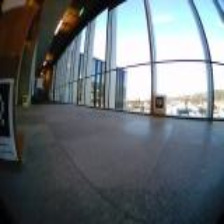}};
\node[tag] at (in.south west) {input};
\coordinate (bc) at (1.52,0);
\draw[draw=black!65, line width=0.45pt, rounded corners=2.4pt, fill=m4col!65] ($(bc)+(-0.30,-1.42)$) rectangle ($(bc)+(0.30,1.42)$);
\draw[draw=black!60, line width=0.4pt,  rounded corners=1.9pt, fill=m3col!65] ($(bc)+(-0.215,-1.14)$) rectangle ($(bc)+(0.215,1.14)$);
\draw[draw=black!60, line width=0.4pt,  rounded corners=1.4pt, fill=m1col!55] ($(bc)+(-0.135,-0.86)$) rectangle ($(bc)+(0.135,0.86)$);
\node[rotate=90, font=\sffamily\bfseries\scriptsize, text=black!90] at (bc) {\paper};
\draw[farr] (in.east) -- ($(bc)+(-0.30,0)$);
\begin{scope}[shift={(2.54,\yt+0.10)}]
  \fill[white] (-0.36,-0.11) rectangle (0.36,0.11); \fill[m1col!80] (-0.36,-0.11) rectangle (-0.27,0.11);
  \draw[black!30, line width=0.25pt] (-0.27,-0.11) -- (-0.27,0.11) (-0.18,-0.11) -- (-0.18,0.11) (0,-0.11) -- (0,0.11);
  \draw[black!60, line width=0.35pt, rounded corners=1pt] (-0.36,-0.11) rectangle (0.36,0.11);
\end{scope}
\begin{scope}[shift={(2.54,-\yt+0.10)}]
  \fill[m4col!65] (-0.36,-0.11) rectangle (0.36,0.11); \fill[m3col!70] (-0.36,-0.11) rectangle (0,0.11); \fill[m2col!75] (-0.36,-0.11) rectangle (-0.18,0.11); \fill[m1col!80] (-0.36,-0.11) rectangle (-0.27,0.11);
  \draw[black!30, line width=0.25pt] (-0.27,-0.11) -- (-0.27,0.11) (-0.18,-0.11) -- (-0.18,0.11) (0,-0.11) -- (0,0.11);
  \draw[black!60, line width=0.35pt, rounded corners=1pt] (-0.36,-0.11) rectangle (0.36,0.11);
\end{scope}
\node[rowlab, anchor=north] at (2.54,\yt-0.08)  {$\mathrm{M}_1$};
\node[rowlab, anchor=north] at (2.54,-\yt-0.08) {$\mathrm{M}_4$};
\draw[farr] ($(bc)+(0.30,0.22)$) -- (2.21,\yt+0.04);
\draw[farr] ($(bc)+(0.30,-0.22)$) -- (2.11,-\yt+0.04);
\foreach \row/\yy/\mdl in {t/\yt/M1, b/-\yt/M4} {
  \foreach \i/\f/\lab in {0/16/4s, 1/32/8s, 2/63/16s} {
    \node[tile] (\row\i) at ({4.08+\i*(\T+\G)},\yy) {\includegraphics[width=\T cm, height=\T cm]{fig/qual/sacson_id3134_\mdl_\f.jpg}};
    \node[tag] at (\row\i.south west) {\lab};
  }
  \draw[farr] (\row0.east) -- (\row1.west);
  \draw[farr] (\row1.east) -- (\row2.west);
}
\draw[farr] (2.86,\yt)  -- (t0.west);
\draw[farr] (2.96,-\yt) -- (b0.west);
\node[tile] (gt) at ({4.08+3*(\T+\G)},0) {\includegraphics[width=\T cm, height=\T cm]{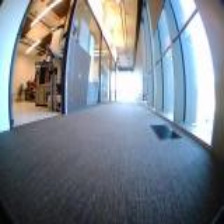}};
\node[tag] at (gt.south west) {16s GT};
\draw[white, line width=1.00pt] ($(t2.south west)+(0.34*\T,0.38*\T)$) rectangle ($(t2.south west)+(0.92*\T,0.88*\T)$);
\draw[cM1, densely dashed, line width=0.70pt] ($(t2.south west)+(0.34*\T,0.38*\T)$) rectangle ($(t2.south west)+(0.92*\T,0.88*\T)$);
\node[font=\sffamily\bfseries\fontsize{5.2}{5.8}\selectfont, text=cM1!90!black, fill=white, fill opacity=0.9, text opacity=1, inner sep=0.7pt, anchor=north] at ($(t2.south west)+(0.63*\T,0.36*\T)$) {hallucinated wall};
\draw[white, line width=1.00pt] ($(b2.south west)+(0.34*\T,0.38*\T)$) rectangle ($(b2.south west)+(0.92*\T,0.88*\T)$);
\draw[cM4, dash pattern=on 2.8pt off 1.1pt on 0.8pt off 1.1pt, line width=0.70pt] ($(b2.south west)+(0.34*\T,0.38*\T)$) rectangle ($(b2.south west)+(0.92*\T,0.88*\T)$);
\draw[white, line width=1.00pt] ($(gt.south west)+(0.34*\T,0.38*\T)$) rectangle ($(gt.south west)+(0.92*\T,0.88*\T)$);
\draw[cM4, dash pattern=on 2.8pt off 1.1pt on 0.8pt off 1.1pt, line width=0.70pt] ($(gt.south west)+(0.34*\T,0.38*\T)$) rectangle ($(gt.south west)+(0.92*\T,0.88*\T)$);
\draw[black!60, dashed, line width=0.45pt] ($(t2.north west)+(-0.08,0.08)$) rectangle ($(gt.east |- b2.south)+(0.08,-0.08)$);
\node[font=\sffamily\fontsize{5.6}{6.2}\selectfont, fill=white, inner sep=1.0pt, text=black!75, anchor=south] at ($0.5*(t2.north)+0.5*(gt.north |- t2.north)+(0,0.09)$) {16\,s vs.\ GT};
\end{tikzpicture}}\\[1pt]
\centering{\scriptsize (a) Long-horizon generation comparison (\sacson, $N{=}49$)}\par
\end{minipage}\hfill
\begin{minipage}[b]{0.324\textwidth}
\centering
\resizebox{\linewidth}{!}{%
\begin{tikzpicture}
\begin{axis}[name=mainb,
  scale only axis, width=3.55cm, height=1.42cm,
  xmode=log, log basis x=10, xmin=0.70, xmax=75, xtick={1,2,5,10,50}, xticklabels={},
  ylabel={\ate (\sacson-cal.) $\downarrow$}, ylabel style={yshift=-1pt},
  ymin=1.90, ymax=2.40, ytick={2.0,2.1,2.2,2.3,2.4}, yticklabel style={/pgf/number format/fixed, /pgf/number format/precision=1, /pgf/number format/fixed zerofill},
  grid=major, grid style={black!12},
  every axis plot/.append style={line width=0.92pt},
  tick label style={font=\scriptsize}, label style={font=\footnotesize},
  legend style={at={(0.50,-0.86)}, anchor=north, legend columns=3, font=\fontsize{6.4}{7.2}\selectfont, draw=none, fill=none, cells={anchor=west}, column sep=2pt, row sep=-0.6pt, /tikz/every even column/.append style={column sep=5pt}},
  legend image code/.code={\draw[mark repeat=2,mark phase=2,#1] plot coordinates {(0cm,0cm) (0.16cm,0cm) (0.32cm,0cm)};},
  legend cell align=left,
  error bars/y dir=both, error bars/y explicit, error bars/error bar style={line width=0.48pt}, error bars/error mark options={rotate=90, mark size=1.25pt},
]
\draw[black!45, densely dashed, line width=0.75pt] (axis cs:0.70,2.241) -- (axis cs:75,2.241);
\node[font=\sffamily\fontsize{6.2}{6.8}\selectfont, text=black!70, inner sep=0.6pt, anchor=south] at (axis cs:12,2.250) {Straight line (2.24)};
\addplot[cmono, mark=square*, mark size=1.45pt] coordinates {(52.1,2.154) +- (0,0.023) (4.40,2.091) +- (0,0.041) (2.26,2.084) +- (0,0.017) (1.19,2.182) +- (0,0.018)}; \addlegendentry{Mono.}
\addplot[cM4, mark=*, mark size=1.55pt] coordinates {(53.2,2.209) +- (0,0.059) (4.44,2.100) +- (0,0.045) (2.26,2.106) +- (0,0.011) (1.17,2.138) +- (0,0.038)}; \addlegendentry{\M{4}}
\addplot[cM1, mark=triangle*, mark size=1.95pt] coordinates {(39.8,2.229) +- (0,0.037) (3.35,2.130) +- (0,0.041) (1.72,2.142) +- (0,0.044) (0.91,2.249) +- (0,0.023)}; \addlegendentry{\M{1}}
\addplot[only marks, mark=diamond*, mark size=2.15pt, mark options={fill=black!22, draw=cmono!90!black, line width=0.62pt}] coordinates {(4.48,2.057) +- (0,0.038)}; \addlegendentry{Mono.\ ($2{\times}2$)}
\addplot[only marks, mark=diamond*, mark size=2.15pt, mark options={fill=cM4!30!white, draw=cM4!90!black, line width=0.62pt}] coordinates {(4.49,2.079) +- (0,0.028)}; \addlegendentry{\M{4} ($2{\times}2$)}
\addplot[only marks, mark=diamond*, mark size=2.35pt, mark options={fill=m4col, draw=cM1!85!black, line width=0.68pt}] coordinates {(3.39,2.092) +- (0,0.029)}; \addlegendentry{\M{1} ($2{\times}2$)}
\addplot[forget plot, only marks, mark=square*, mark size=1.45pt, mark options={fill=white, draw=cmono, line width=0.8pt}] coordinates {(52.1,2.154)};
\addplot[forget plot, only marks, mark=*, mark size=1.55pt, mark options={fill=white, draw=cM4, line width=0.8pt}] coordinates {(53.2,2.209)};
\addplot[forget plot, only marks, mark=triangle*, mark size=1.95pt, mark options={fill=white, draw=cM1, line width=0.8pt}] coordinates {(39.8,2.229)};
\end{axis}
\begin{axis}[name=stripb, at={($(mainb.south west)+(0,-0.12cm)$)}, anchor=north west,
  scale only axis, width=3.55cm, height=0.35cm,
  xmode=log, log basis x=10, xmin=0.70, xmax=75, xtick={1,2,5,10,50}, xticklabels={1,2,5,10,50},
  xlabel={Compiled latency per decision (s, log)}, xlabel style={font=\footnotesize, yshift=2pt},
  ymin=0.60, ymax=0.80, ytick={0.7}, yticklabel style={/pgf/number format/fixed, /pgf/number format/precision=1, /pgf/number format/fixed zerofill},
  grid=major, grid style={black!12}, tick label style={font=\scriptsize},
]
\fill[black!14] (axis cs:0.70,0.665) rectangle (axis cs:75,0.715);
\draw[black!55, line width=0.6pt] (axis cs:0.70,0.69) -- (axis cs:75,0.69);
\node[font=\sffamily\fontsize{6.2}{6.8}\selectfont, text=black!75, fill=white, fill opacity=0.85, text opacity=1, inner sep=0.6pt] at (axis cs:7,0.69) {Oracle (0.69)};
\end{axis}
\foreach \side in {west, east} {
  \draw[black!75, line width=0.45pt] ($(mainb.south \side)+(-0.07,-0.03)$) -- ($(mainb.south \side)+(0.07,0.03)$);
  \draw[black!75, line width=0.45pt] ($(stripb.north \side)+(-0.07,-0.03)$) -- ($(stripb.north \side)+(0.07,0.03)$);
}
\end{tikzpicture}}\\[1pt]
{\scriptsize (b) Calibrated \sacson planning vs.\ latency}
\end{minipage}
\captionsetup{hypcap=false}

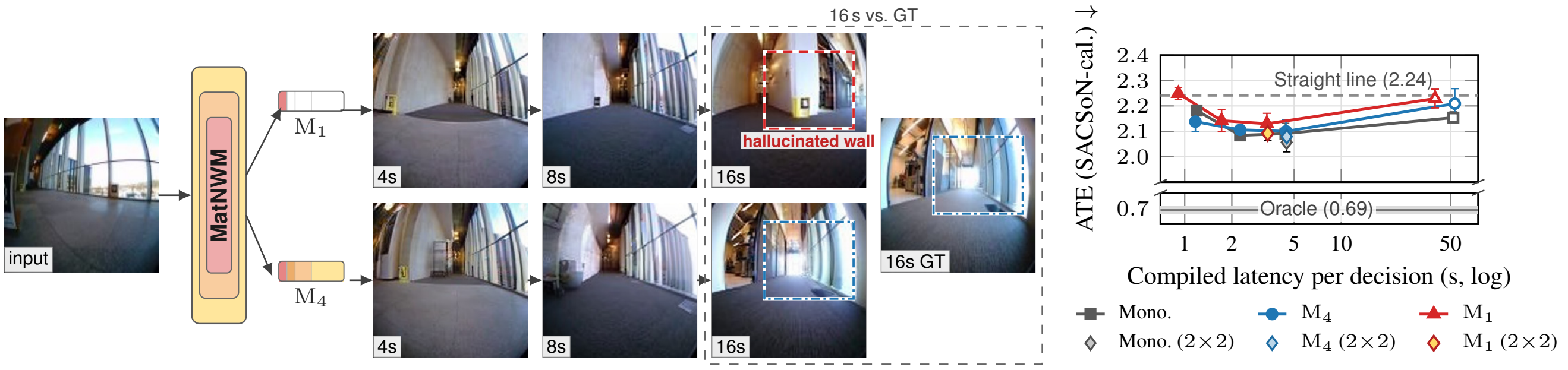
\captionof{figure}{\textbf{Generation needs full granularity and many NFEs. Planning needs low-variance rankings, which few-NFE Euler provides.} (a)~In a 16\,s \sacson rollout at 49 NFEs, \M{4} preserves corridor geometry (blue dash-dotted) whereas \M{1} hallucinates a wall (red dashed). (b)~\ate\ vs.\ latency per decision on calibrated \sacson; NFEs increase left to right and hollow markers are 49 NFEs. Dashed line: straight-line baseline without a world model (2.24). Lower strip: oracle ranking (0.69). After Holm correction, only the $2{\times}2$-NFE models are significantly below the line. \M{1} at $2{\times}2$ NFEs plans in 3.39\,s, $15.4\times$ faster than 49-NFE \mono\ ($11.6\times$ from the sampler alone).}
\captionsetup{hypcap=true}
\label{fig:teaser}
\end{strip}

\begin{abstract}
Navigation world models use one backbone and one dense multi-step sampler both to generate future frames and to rank candidate plans. We ask whether planning needs the capacity and integration accuracy that generation does. To answer this within one training run, we train \paper, whose feed-forward layers nest four granularities, alongside an identically trained monolithic baseline. Using \paper as a controlled probe, we find that the two tasks have asymmetric demands. Visual realism requires full granularity and many neural function evaluations (NFEs). In contrast, planning ranks candidates by expected outcome, meaning low-variance cost estimates are required rather than sharp samples. We trace this difference to the optimizer's curse: selecting the best of many noisy single-rollout scores rewards favorable noise. While a single NFE collapses to the conditional mean and distorts sharp turns, we show that averaging two 2-NFE rollouts balances both effects, and coarse Euler integration provides stable cost estimates without retraining. To this end, few-step Euler alone makes 49-NFE planning $11.6\times$ faster with no loss in accuracy. Utilizing \paper's smallest granularity further raises this speedup to $15.4\times$, while realistic future predictions can still be generated with \paper at full granularity. We plan to release the model weights and source code.
\end{abstract}

\section{Introduction}
\label{sec:intro}

Unlike navigation policies such as GNM, ViNT, and NoMaD~\cite{shah2023gnm,shah2023vint,sridhar2024nomad}, navigation world models~\cite{bar2025nwm,zhang2026raenwm} learn action-conditioned visual dynamics and serve two tasks by forward simulation. Generation is judged frame by frame for visual fidelity, whereas planning consumes only relative candidate rankings. Existing latent (\nwm~\cite{bar2025nwm}) and dense-token (\raenwm~\cite{zhang2026raenwm}) models serve both tasks with one fixed backbone and one dense ordinary differential equation (ODE) sampler. Planning therefore pays the full cost of generation. Scoring 120 eight-step candidates at 49 neural function evaluations (NFEs) per frame takes 52.1\,s per decision on an A100 GPU with compiled kernels. Planning ranks candidates by their expected outcome. It therefore needs low-variance cost estimates rather than sharp samples, and coarse Euler integration of a flow-matching world model provides them without retraining.
\begin{figure*}[t]
\centering
\resizebox{0.96\textwidth}{!}{\input{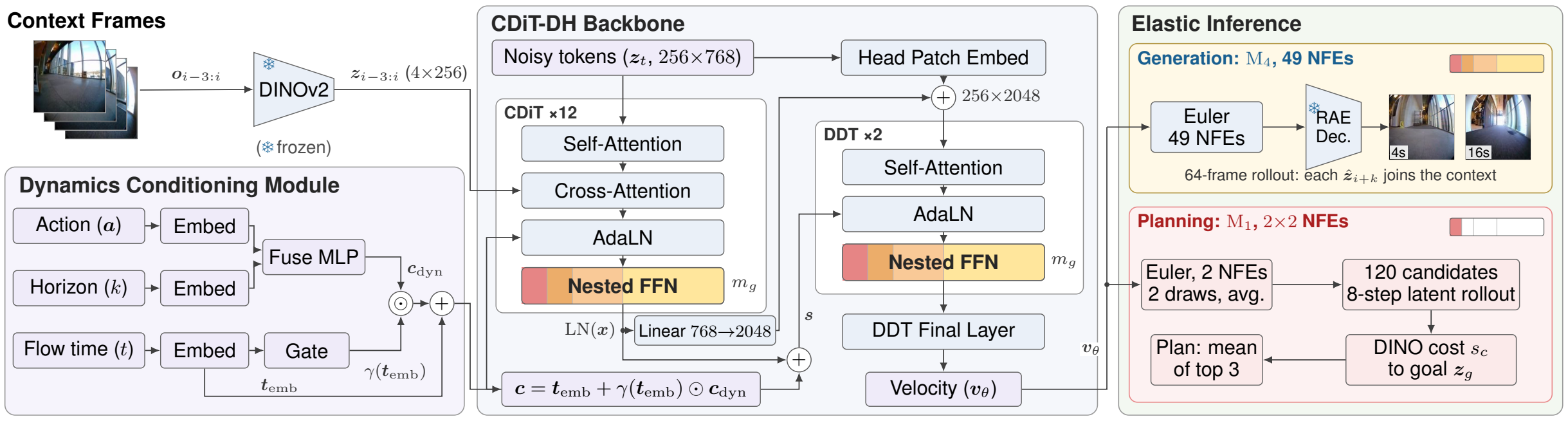}}
\vspace{2pt}
\caption{\textbf{Architecture and inference of \paper.} Frozen DINOv2 ({\color{cM4!85!black}\ding{100}}) encodes context frames into patch tokens $\bm{z}_{i-3:i}$, and the dynamics conditioning module gates $(\bm{a},k,t)$ into $\bm{c}$. The inherited CDiT-DH backbone predicts velocity $\bm{v}_\theta$ from $\bm{z}_t$. Normalized trunk tokens $\mathrm{LN}(\bm{x})$ reach the DDT head via a $768{\to}2048$ projection and per-token conditioning $\bm{s}=\mathrm{LN}(\bm{x})+\bm{c}$. Relative to \mono, only the nested FFN blocks differ. They share weights across four prefix hidden sizes $m_g$ ($\mathrm{M}_1\subset\mathrm{M}_2\subset\mathrm{M}_3\subset\mathrm{M}_4$; pre-attention AdaLN and block residuals omitted). Right: one checkpoint serves both tasks at different granularities, generation with \M{4} at 49 NFEs and open-loop \cem planning with \M{1} at $2{\times}2$ NFEs.}
\label{fig:arch}
\end{figure*}

Distilled~\cite{salimans2022progressive,bar2025nwm} and shortcut~\cite{frans2025shortcut,shen2026onestepwm} students accelerate generation by matching multi-step ODE endpoints from each noise draw. Lightweight predictors bypass pixel generation~\cite{liu2026litenwm,mei2026navwm}, and elastic diffusion transformers~\cite{zhao2024dydit,anagnostidis2025flexidit} trade compute against visual quality. The ODE demands of sampling-based candidate ranking remain uncharacterized.

Do generation and planning need the same capacity and the same sampler? We isolate capacity within one training run. \paper nests four Matryoshka granularities ($\M{1}\subset\M{2}\subset\M{3}\subset\M{4}$) in its feed-forward layers ($0.65\text{--}1.0\times$ multiply--accumulate operations, MACs)~\cite{kusupati2022mrl,devvrit2024matformer}, and we train it alongside an identically trained \mono baseline (\cref{fig:teaser}). \paper serves as a within-run capacity probe. We find that most of the speed-up comes from the ODE sampler; the nested granularity contributes only once the sampler is pushed to its limit. The two tasks have asymmetric demands. Perceptual realism (Fr\'echet inception distance, \fid) improves with granularity and collapses below 4 NFEs. Open-loop planning at the smallest granularity matches full granularity and improves on \recon and \sacson when the sampler drops from 49 to 4 NFEs.

Our controls trace this gain to sampling variance. At 49 NFEs per frame, each candidate rollout draws its own random realization; selecting the top 3 of 120 noisy scores consequently rewards favorable noise rather than better actions---a classic case of the optimizer's curse~\cite{smith2006optimizers}. Coarse Euler integration damps this variance without retraining, as evidenced by second-order solvers failing to beat 4-NFE Euler at equal compute. However, reducing this further to a single NFE collapses the sampler toward the conditional mean, which distorts sharp indoor turns. We show that averaging two 2-NFE rollouts balances the two effects.

The contributions of this work are threefold:
\begin{itemize}
\item We show that generation and open-loop planning impose asymmetric demands. Candidate ranking by the cross-entropy method (\cem) tolerates the smallest granularity and improves under coarse Euler where \fid\ degrades.
\item We trace the gain to the optimizer's curse. We measure selection optimism on held-out draws and separate variance contraction from single-step ranking shift with solver, time-grid, curvature, and multi-draw controls.
\item We introduce \paper as a within-run capacity probe that serves full-granularity generation and smallest-granularity planning from one trained model. It plans $15.4\times$ faster than standard 49-NFE planning with no loss in accuracy ($11.6\times$ from the sampler alone).
\end{itemize}

\section{Related Work}
\label{sec:related}

\paragraph{Navigation World Models.}
World models~\cite{ha2018worldmodels} such as Dreamer~\cite{hafner2020dreamer} and DreamerV3~\cite{hafner2023dreamerv3} support visual control via forward simulation. \nwm~\cite{bar2025nwm} trains a conditional diffusion transformer (CDiT) on variational autoencoder (VAE) latents for \cem navigation, while \raenwm~\cite{zhang2026raenwm} and FlowWM~\cite{porcher2026flowwm} model stochastic dynamics via flow matching in dense representation space~\cite{zheng2026rae}. Other designs change the predictor: \textsc{NavWM}~\cite{mei2026navwm} unifies foresight generation with policy planning, LiteNWM~\cite{liu2026litenwm} replaces candidate-wise autoregressive rollouts with multi-horizon latent scoring, and non-generative DINO-WM~\cite{zhou2025dinowm} and V-JEPA~2~\cite{assran2025vjepa2} plan without pixel generation. Within generative navigation models, the deterministic regressor of \raenwm trails flow matching~\cite{zhang2026raenwm}, whereas we show standard flow-matching weights plan accurately in 1--4 evaluations (\cref{sec:res-why}).

\paragraph{Elastic and Nested Networks.}
Slimmable~\cite{yu2019slimmable,yu2019usnets}, once-for-all~\cite{cai2020ofa}, and Matryoshka networks~\cite{kusupati2022mrl} train weight-sharing sub-models across compute budgets. MatFormer~\cite{devvrit2024matformer} nests Transformer feed-forward dimensions into prefix sub-models, and Flextron~\cite{cai2024flextron} adds learned routing. In diffusion transformers, DyDiT~\cite{zhao2024dydit}, FlexiDiT~\cite{anagnostidis2025flexidit}, elastic routing variants~\cite{wang2026edit,du2026elasticdit}, and Matryoshka Diffusion~\cite{gu2023mdm} optimize visual quality per FLOP. We adopt MatFormer nesting primarily as a controlled within-run probe to isolate sub-network granularity on candidate ranking without cross-run SGD variance.

\paragraph{Few-Step Sampling and Planning.}
Flow matching and rectified flow~\cite{lipman2023flow,liu2023rectified} straighten transport paths, shifted grids~\cite{esser2024sd3} and higher-order solvers~\cite{karras2022edm} cut evaluations, and distillation~\cite{salimans2022progressive,song2023consistency} or shortcut self-consistency~\cite{frans2025shortcut} accelerates generation. In navigation, \nwm~\cite{bar2025nwm} trains a 6-step distilled student, whereas One-Step WM~\cite{shen2026onestepwm} trains a non-autoregressive 3D U-Net via a single-stage shortcut objective~\cite{frans2025shortcut}. Interactive game world models also use 3--4 denoising steps~\cite{alonso2024diamond,valevski2025gamengen}. An ideal trajectory-matching student would preserve single-draw sampling variance unless its rollouts are averaged. Coarse Euler instead contracts this variance without retraining. We show that contraction mitigates optimizer's-curse selection optimism~\cite{smith2006optimizers}, echoing reinforcement-learning objective mismatch~\cite{lambert2020objective}.

\section{Method}
\label{sec:method}

Given recent visual observations $\bm{z}_{i-3:i}$, a relative planar action $\bm{a}=(\Delta x,\Delta y,\Delta\psi)$, and a prediction horizon $k$, a navigation world model learns $p(\bm{z}_{i+k}\mid\bm{z}_{i-3:i},\bm{a},k)$. \paper parameterizes this distribution with a CDiT over dense visual tokens whose feed-forward layers are nested across four granularities (\cref{fig:arch}), so one checkpoint serves both tasks at different granularities.

\subsection{Representation, Backbone, and~Objective}
\label{sec:method-base}

A frozen DINOv2-B encoder~\cite{oquab2024dinov2} maps each $224{\times}224$ frame into $L{=}256$ patch tokens ($d{=}768$) normalized by representation autoencoder (RAE) statistics~\cite{zheng2026rae}, whose frozen decoder reconstructs RGB frames for visualization and pixel metrics. As in \raenwm, dense semantic tokens improve rollout stability and action controllability over VAE latents~\cite{zhang2026raenwm}. The inherited CDiT-DH backbone~\cite{bar2025nwm,zhang2026raenwm} has a 12-block trunk ($d{=}768$, 12 heads) with target self-attention and cross-attention to $\bm{z}_{i-3:i}\in\R^{4\times 256\times 768}$, plus a 2-block decoupled diffusion transformer (DDT) head ($d{=}2048$, 16 heads)~\cite{wang2026ddt}. Standard DiT blocks~\cite{peebles2023dit} integrate $\bm{c}=\bm{t}_{\mathrm{emb}}+\gamma(\bm{t}_{\mathrm{emb}})\odot\bm{c}_{\mathrm{dyn}}\in\R^{768}$~\cite{zhang2026raenwm} via AdaLN. Normalized trunk tokens $\mathrm{LN}(\bm{x})\in\R^{256\times 768}$ enter the DDT head via both a $768{\to}2048$ linear projection added to the head patch embedding and per-token $\bm{s}=\mathrm{LN}(\bm{x})+\bm{c}$, where $\bm{c}$ is broadcast across all $L{=}256$ tokens before $768{\to}6{\times}2048$ AdaLN modulation (\cref{fig:arch}). We train $\vel$ via conditional flow matching~\cite{lipman2023flow} along $\bm{z}_t=(1-t)\,\bm{z}_{i+k}+t\,\bm{\epsilon}$, $\bm{\epsilon}\sim\mathcal{N}(\bm{0},\bm{I})$, under the Frobenius norm over $\R^{L\times d}$:
\begin{equation}
\mathcal{L}=\mathbb{E}_{t,\bm{\epsilon}}\,\big\lVert \vel(\bm{z}_t,t,\bm{z}_{i-3:i},\bm{a},k)-(\bm{\epsilon}-\bm{z}_{i+k})\big\rVert_F^2 ,
\label{eq:fm}
\end{equation}
with flow-time shift $\alpha = \sqrt{dL/4{,}096} = \sqrt{48} \approx 6.93$~\cite{esser2024sd3,zheng2026rae,zhang2026raenwm}. At inference, Euler takes $N$ steps over $N{+}1$ shifted time points $t_j=\alpha u_j/(1+(\alpha-1)u_j)$ with $u_j=1-j(1-10^{-3})/N$ ($j=0,\dots,N$, from $t_0{=}1$ to $t_N{\approx}0.007$). With Euler, one step costs one NFE, so each frame costs $N$ NFEs ($N{=}49$ by default~\cite{zhang2026raenwm}). At $N{=}4$, the grid $(1,0.95,0.87,0.70,0.007)$ places 69\,\% of $[0,1]$ in the final step (87\,\% at $N{=}2$).

\subsection{Nested World Model and Elastic Inference}
\label{sec:method-nested}

To isolate feed-forward capacity in one run while serving generation and planning from one checkpoint, \paper structures every trunk and head FFN into nested prefix sub-layers~\cite{kusupati2022mrl,devvrit2024matformer}. Given $\bm{x}\in\R^{L\times d}$, weights $\bm{W}_1\in\R^{d\times d_{\mathrm{ff}}}$, $\bm{W}_2\in\R^{d_{\mathrm{ff}}\times d}$, and biases $\bm{b}_1,\bm{b}_2$, granularity $g\in\{1,2,3,4\}$ selects the first $m_g$ hidden units (prefix width):
\begin{equation}
\mathrm{FFN}_g(\bm{x})=\phi\big(\bm{x}\bm{W}_1^{(g)}+\bm{b}_1^{(g)}\big)\bm{W}_2^{(g)}+\bm{b}_2 .
\label{eq:nested}
\end{equation}
Here $\phi$ is GELU, $\bm{W}_1^{(g)}\in\R^{d\times m_g}$ and $\bm{W}_2^{(g)}\in\R^{m_g\times d}$ hold the first $m_g$ columns of $\bm{W}_1$ and rows of $\bm{W}_2$, and $m_g\in\{384,768,\allowbreak 1536,3072\}$ in the trunk and $\{1024,2048,\allowbreak 4096,8192\}$ in the head. Selecting $g$ yields $\M{1}\subset\M{2}\subset\M{3}\subset\M{4}$, which share attention and projections. We call \M{1} the smallest granularity and \M{4} the full one. Each training step draws $g\sim\mathcal{U}\{1,2,3,4\}$ and optimizes \cref{eq:fm} on $\M{g}$~\cite{devvrit2024matformer}, activating 207.4--315.7\,M parameters (0.66--1.0$\times$) and 51.4--79.1\,GMACs (0.65--1.0$\times$). Sweeping $g\in\{1,\dots,4\}$ and $N\in\{49,19,9,6,\allowbreak 4,3,2,1\}$ spans 0.01--1.0$\times$ compute.

\subsection{Generation and Planning}
\label{sec:method-tasks}

Open-loop generation evaluates direct prediction at $k\in\{4,8,16,32,64\}$ frames (1--16\,s at 4\,Hz) and 64-frame rollouts ($k{=}1$ per step). Planning follows the benchmark \cem protocol~\cite{bar2025nwm,zhang2026raenwm} with $M{=}1$ rollout draw per candidate by default ($M\in\{1,\dots,4\}$ averaged draws analyzed in \cref{sec:res-why}). One \cem round samples $C{=}120$ candidates (three waypoints interpolated to eight steps) and rolls each out autoregressively for eight steps at $k{=}1$ ($0.25$\,s/step) in normalized DINOv2 token space with independent Gaussian noise and no RAE decoding. With one round, \cem reduces to sampling-based shooting with top-3 averaging. Following \raenwm~\cite{zhang2026raenwm}, standard \cem scores only the terminal token $\hat{\bm{z}}_{i+8}$ by cosine DINO patch distance to goal tokens $\bm{z}_g$ (a cost; lower is better; trajectory-averaged DINO in \cref{sec:res-why}) and returns the mean of the top-3 elites ($120\times 8\times N$ evaluations/decision for $M{=}1$). We evaluate plans open-loop against the recorded 2\,s path via \ate\ and relative pose error (\rpe) in dataset step-size units~\cite{sturm2012benchmark,bar2025nwm} ($1\,\text{unit}\approx 0.25$\,m on \recon/\sacson, $0.36$\,m on \scand).

\section{Experimental Setup}
\label{sec:setup}

\paragraph{Datasets and Implementation.}
We evaluate on \recon~\cite{shah2021recon} (off-road), \sacson~\cite{hirose2024sacson} (indoor crowds), and \scand~\cite{karnan2022scand} (social navigation), training one joint model on the combined training splits and testing on 897 held-out trajectories (9{,}561 samples: 4{,}262, 3{,}416, 1{,}883). \paper and \mono share architecture, data, and 50-epoch AdamW schedule (batch 96, four future goal offsets/trajectory sample, learning rate $2{\times}10^{-4}\to2{\times}10^{-6}$, zero weight decay, EMA, primary training seed on an A100-SXM4-40GB GPU), differing only in FFN nesting. Both models are also trained with a second seed, while the evaluation results are provided in the supplement.

\paragraph{Metrics and Baselines.}
Generation is evaluated by \fid~\cite{heusel2017fid}, learned perceptual image patch similarity (\lpips)~\cite{zhang2018lpips}, and cosine DINOv2 patch distance~\cite{zhang2026raenwm} on all 9{,}561 test samples; planning by \ate\ and \rpe~\cite{sturm2012benchmark} in dataset step-size units on 500 shared seeded test items per dataset. \mono is our primary paired baseline isolating nesting, with \nwm~\cite{bar2025nwm}, \raenwm~\cite{zhang2026raenwm} (and DINO-Reg), GNM~\cite{shah2023gnm}, and NoMaD~\cite{sridhar2024nomad} as full-split references.

\paragraph{Statistics.}
To evaluate performance, we analyze paired differences between models using shared samples or planning items before rounding (with intervals in the supplement), estimating 95\,\% bootstrap intervals over items (where ``\sig'' denotes intervals excluding zero). Because each evaluation seed re-draws candidate plans and sampler noise, our primary planning tables average per-item differences over three evaluation seeds before resampling. Consequently, a planner is deemed non-inferior if the upper bound of its interval lies below 5\,\% of the 49-NFE \mono error. This rigorous setup avoids the optimism of single-seed controls. Specifically, our primary comparisons assess \paper \M{1} versus the \mono baseline at 49 NFEs, alongside internal 4-NFE versus 49-NFE comparisons on \recon and \sacson.

\section{Experimental Results}
\label{sec:results}

\begin{table}[t]
\centering
\scriptsize
\caption{\textbf{Open-loop generation quality.} \lpips, cosine DINOv2 patch distance (\emph{DINO}, the \cem planning cost), and \fid on \sacson (4, 16\,s) and \recon/\scand (4\,s). Full horizons, further DINO columns, and standalone models are in the supplement. \emph{NFE}: neural function evaluations per frame (\nwm: 250 DDPM steps). Bottom: 64-frame autoregressive rollouts on all $9{,}561$ test anchors, following \raenwm's rollout protocol (frame 16 = 4.25\,s, frame 63 = 16\,s; non-antialiased $256{\to}224$ bicubic resize). \textbf{Bold}/\underline{underline}: best/second best direct-prediction row among our models (ties share rank).}
\label{tab:sacson}
\setlength{\tabcolsep}{1.45pt}\renewcommand{\arraystretch}{0.94}
\resizebox{\linewidth}{!}{%
\begin{tabular}{@{}lc ccccc cc cc@{}}
\toprule
 & & \multicolumn{5}{c}{\sacson} & \multicolumn{2}{c}{\recon (4\,s)} & \multicolumn{2}{c}{\scand (4\,s)} \\
\cmidrule(lr){3-7}\cmidrule(lr){8-9}\cmidrule(l){10-11}
 & & \multicolumn{2}{c}{\lpips \dn} & DINO \dn & \multicolumn{2}{c}{\fid \dn} & \lpips \dn & \fid \dn & \lpips \dn & \fid \dn \\
\cmidrule(lr){3-4}\cmidrule(lr){5-5}\cmidrule(lr){6-7}\cmidrule(lr){8-8}\cmidrule(lr){9-9}\cmidrule(lr){10-10}\cmidrule(l){11-11}
Model & NFE & 4\,s & 16\,s & 4\,s & 4\,s & 16\,s & 4\,s & 4\,s & 4\,s & 4\,s \\
\midrule
\nwm~\cite{bar2025nwm} & 250 & 0.407 & 0.470 & 0.402 & 26.2 & 33.1 & -- & -- & -- & -- \\
\raenwm~\cite{zhang2026raenwm} & 49 & 0.303 & 0.349 & 0.327 & 15.1 & 15.9 & -- & -- & -- & -- \\
\midrule
\mono & 49 & 0.323 & 0.361 & 0.345 & \bestval{16.5} & \bestval{17.8} & \secval{0.374} & \bestval{18.9} & 0.405 & \bestval{28.5} \\
\paper \M{4} & 49 & 0.333 & 0.369 & 0.354 & \secval{17.1} & \secval{18.7} & 0.379 & \secval{19.1} & 0.399 & \secval{29.1} \\
\paper \M{1} & 49 & 0.343 & 0.379 & 0.362 & 19.1 & 21.0 & 0.385 & 20.3 & 0.399 & 31.1 \\
\midrule
\mono & 4 & \bestval{0.310} & \bestval{0.347} & \secval{0.326} & 20.5 & 22.4 & \bestval{0.369} & 21.7 & \secval{0.397} & 33.3 \\
\paper \M{4} & 4 & 0.320 & \secval{0.355} & 0.335 & 21.6 & 23.7 & 0.375 & 22.9 & \bestval{0.389} & 34.1 \\
\paper \M{1} & 4 & 0.331 & 0.365 & 0.344 & 24.7 & 26.6 & 0.382 & 25.0 & \bestval{0.389} & 36.4 \\
\midrule
\mono & 2 & \secval{0.319} & \secval{0.355} & \bestval{0.323} & 32.7 & 38.3 & 0.376 & 25.1 & 0.406 & 49.9 \\
\paper \M{4} & 2 & 0.329 & 0.365 & 0.334 & 37.1 & 43.3 & 0.382 & 27.6 & 0.398 & 52.1 \\
\paper \M{1} & 2 & 0.344 & 0.378 & 0.346 & 44.7 & 52.0 & 0.389 & 31.0 & 0.401 & 55.1 \\
\midrule
\multicolumn{11}{@{}l}{\emph{64-frame autoregressive rollout}} \\
\mono & 49 & 0.371 & 0.450 & 0.384 & 17.1 & 19.1 & 0.445 & 19.3 & 0.448 & 27.5 \\
\paper \M{4} & 49 & 0.378 & 0.454 & 0.390 & 18.0 & 20.5 & 0.449 & 19.3 & 0.438 & 26.8 \\
\paper \M{1} & 49 & 0.390 & 0.471 & 0.400 & 20.2 & 22.7 & 0.454 & 21.4 & 0.441 & 27.3 \\
\bottomrule
\end{tabular}}
\end{table}

\begin{table}[t]
\centering
\scriptsize
\caption{\textbf{Open-loop planning accuracy (\ate $\downarrow$)} of 2\,s \cem plans in dataset step-size units~\cite{bar2025nwm} ($1\,\text{unit}\approx 0.25$\,m on \recon/\sacson, $0.36$\,m on \scand). Mean over three evaluation seeds on 500 items/dataset (sd and \rpe\ in supplement; single-draw 1- and 2-NFE results in \cref{fig:inference,tab:controls}). \sacson-cal.\ fits the \cem speed ($v_0{=}0.90$ vs.\ $0.32$ units/step) on training splits only. {\setlength{\fboxsep}{1pt}\colorbox{beatbase}{Shaded}}: mean below the best model-free baseline of the column. $^{\ddagger}$:~beats the \sacson-cal.\ line after Holm correction. \textbf{Bold}: best of our world models. \emph{$2{\times}2$}: two-draw 2-NFE average. \emph{Same-recipe DINO-Reg}: our 50-epoch regressor. \emph{Random shooting}: mean of three random candidates (20 repetitions).}
\label{tab:planning}
\setlength{\tabcolsep}{2.6pt}\renewcommand{\arraystretch}{0.94}
\newcommand{\bb}[1]{\cellcolor{beatbase}#1}
\begin{tabular}{@{}lccccc@{}}
\toprule
Planner & NFE & \recon & \sacson & \sacson-cal. & \scand \\
\midrule
GNM~\cite{shah2023gnm}   & -- & 1.87 & 3.71 & -- & 2.12 \\
NoMaD~\cite{sridhar2024nomad} & -- & 1.95 & 3.73 & -- & 2.24 \\
\nwm~\cite{bar2025nwm}   & 250 & 1.13 & 4.12 & -- & 1.28 \\
DINO-Reg~\cite{zhang2026raenwm} & -- & 1.46 & 3.14 & -- & 1.44 \\
\raenwm~\cite{zhang2026raenwm} & 49 & 1.36 & 2.91 & -- & 1.14 \\
Same-recipe DINO-Reg & -- & 2.23 & 3.09 & -- & 1.22 \\
\midrule
\mono & 49 & \bb{1.36} & 2.84 & \bb{2.15} & 1.04 \\
\paper \M{4} & 49 & \bb{1.36} & 2.83 & \bb{2.21} & 1.03 \\
\paper \M{1} & 49 & \bb{1.37} & 2.84 & \bb{2.23} & 1.03 \\
Standalone \M{1} & 49 & \bb{1.37} & 2.83 & -- & 1.02 \\
Standalone \M{0} & 49 & \bb{1.40} & 3.02 & 2.35 & 1.02 \\
\midrule
\mono & 4 & \bb{1.30} & 2.76 & \bb{2.09\rlap{$^{\ddagger}$}} & 1.03 \\
\paper \M{4} & 4 & \bb{1.27} & 2.77 & \bb{2.10\rlap{$^{\ddagger}$}} & 1.01 \\
\paper \M{1} & 4 & \bb{1.27} & 2.75 & \bb{2.13} & 1.01 \\
Standalone \M{1} & 4 & \bb{1.28} & 2.72 & -- & \bestval{0.99} \\
Standalone \M{0} & 4 & \bb{1.26} & 2.92 & 2.24 & 1.01 \\
\midrule
\mono (2 draws) & $2{\times}2$ & \bb{1.25} & \bestval{2.59} & \bb{\bestval{2.06}\rlap{$^{\ddagger}$}} & 1.05 \\
\paper \M{4} (2 draws) & $2{\times}2$ & \bb{\bestval{1.24}} & 2.64 & \bb{2.08\rlap{$^{\ddagger}$}} & \bestval{0.99} \\
\paper \M{1} (2 draws) & $2{\times}2$ & \bb{\bestval{1.24}} & 2.63 & \bb{2.09\rlap{$^{\ddagger}$}} & 1.00 \\
\midrule
\multicolumn{6}{@{}l}{\emph{No world model}} \\
\cem prior mean & -- & 1.64 & 3.78 & 2.24 & 0.83 \\
Straight line (train-fit speed) & -- & 1.59 & 2.24 & 2.24 & 0.83 \\
Random shooting & -- & 1.77 & 4.07 & 2.80 & 1.08 \\
Oracle$^{\dagger}$ & -- & 0.61 & 1.08 & 0.69 & 0.27 \\
\bottomrule
\end{tabular}\\[1pt]
\parbox{\linewidth}{\scriptsize $^{\dagger}$Mean \ate\ of the 3 ground-truth-closest candidates in an independent 120-candidate draw; bounds proposal quality.}
\vspace{-5pt}
\end{table}

We test whether generation (\cref{sec:res-gen}) and planning (\cref{sec:res-plan}) need full granularity at 49 NFEs, as well as how both respond as $N$ shrinks (\cref{sec:res-steps}), and why coarse Euler improves candidate ranking (\cref{sec:res-why}). Unless otherwise specified, all planning performance metrics are averaged over three evaluation seeds.

\subsection{Open-Loop Generation}
\label{sec:res-gen}

Generation quality scales monotonically with model granularity, demonstrating that our nested training imposes negligible performance overhead (\cref{tab:sacson}). Specifically, direct-prediction \fid\ improves consistently from \M{1} to \M{4} across all three datasets, notably dropping from 19.1 to 17.1 on the 4-second \sacson benchmark. This hierarchical ordering is further preserved during long-horizon (64-frame) rollouts on both \sacson and \recon. When operating at 49 NFEs, our highest-granularity \M{4} model performs comparably to the \mono baseline, trailing it by at most 1.5 \fid\ across 15 direct settings (with complete results provided in the supplement).

\subsection{CEM Planning}
\label{sec:res-plan}

In contrast, \cem planning yields the opposite observation: the smallest granularity is sufficient, as it already matches the performance of the full model (\cref{tab:planning}). On \recon, \sacson, and \scand, the three models (\mono, \M{4}, \M{1}) differ by at most 0.03 \ate\ at both 49 and 4 NFEs, which is well within our 5\,\% non-inferiority margin. Furthermore, the two-draw, 2-NFE averaging scheme ($2{\times}2$) produces the strongest overall world-model planners on \recon and \sacson, where the lightweight \M{1} model performs on par with the \mono baseline. On the \scand dataset, however, no active planner outperforms a straight-line baseline (0.83 \ate). Because open-loop navigation on \scand's smooth sidewalks favors a zero-variance prior, we do not consider \scand a win for any planner, focusing instead on the more challenging environments of \recon and \sacson.

\begin{figure}[t]
\centering
\includegraphics[width=\linewidth]{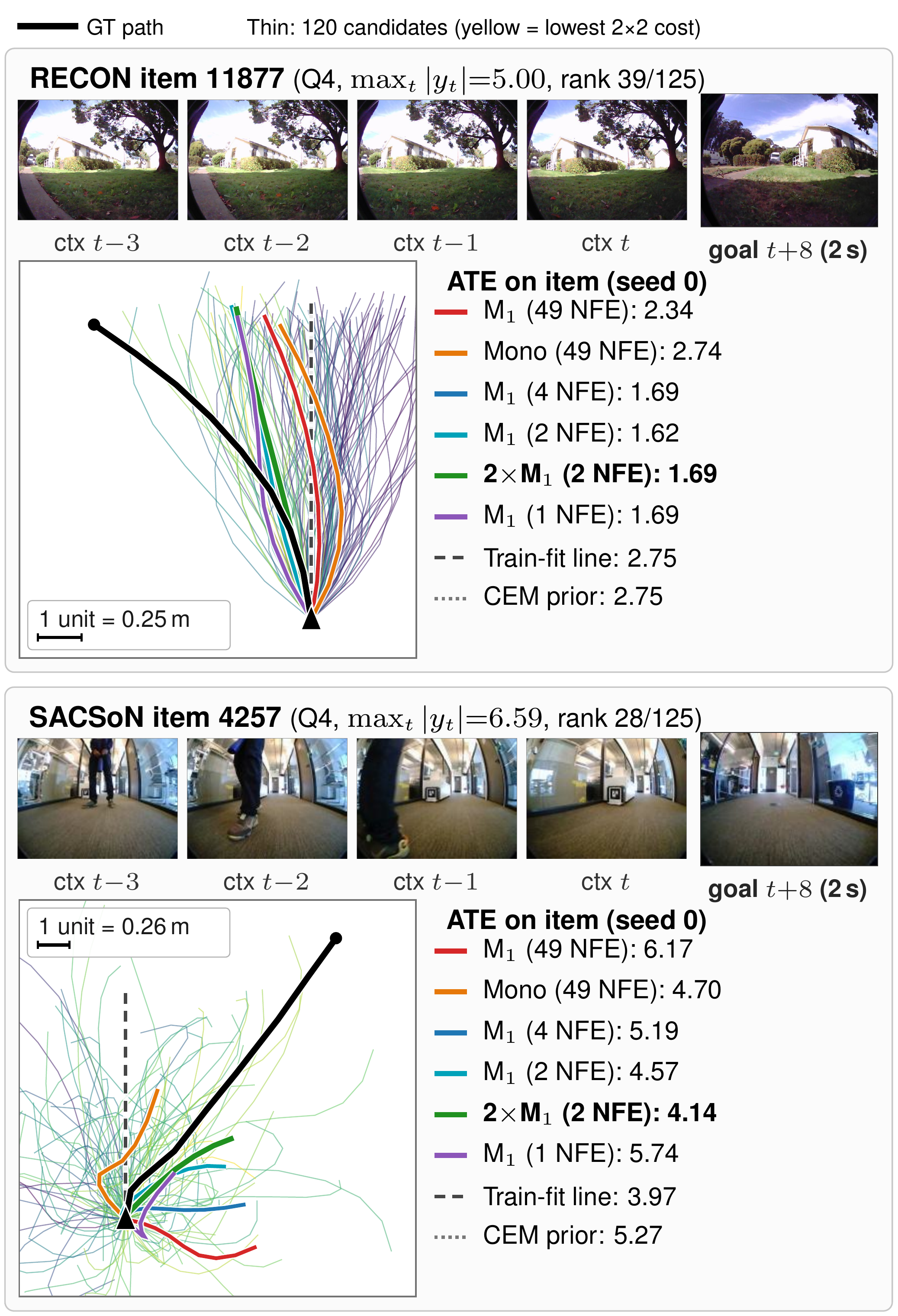}
\vspace{-12pt}
\caption{\textbf{Planning on two sharp turns.} Thin lines show the 120 \cem candidates colored by $2{\times}2$-NFE cost. With 49 NFEs, the \M{1} elites miss the turn on \recon and turn the wrong way on \sacson. With two 2-NFE draws ($2{\times}2$), they track the GT direction (\ate\ 2.34 $\to$ 1.69 on \recon item 11877 and 6.17 $\to$ 4.14 on \sacson item 4257). \sacson uses the default speed prior, so its fans stay short. More items appear in the supplement.}
\label{fig:planning}
\end{figure}

\paragraph{Comparison with Model-Free Baselines.}
World-model planning outperforms model-free straight-line baselines on curved trajectories. Specifically, on \recon, our $2{\times}2$-NFE \M{1} planner surpasses the train-fit line by 0.35 \ate\ overall (\cref{tab:planning}), and by 1.03 \ate\ (approx.\ 26\,cm) on the sharpest-turn quartile (Q4) where the line and \cem prior coincide (\cref{tab:turns,fig:planning}). On \sacson, planning efficacy is governed by the speed prior; the default \cem prior restricts forward progress, preventing any world model from outperforming the baseline. Calibrating a single speed scalar (\sacson-cal.) addresses this, lowering the \ate\ by 0.50--0.68 across all evaluations. In this calibrated setting (\cref{tab:planning}), no 49-NFE model outperforms the 2.24 straight-line baseline after Holm correction, whereas all $2{\times}2$-NFE configurations do (\cref{fig:teaser}(b)). These improvements are driven by curved paths, as the zero-jitter line remains superior on straight corridors (Q1 in \cref{tab:turns}). As a reference, the 1\,B \nwm~\cite{bar2025nwm} obtains 1.13 \ate\ on \recon with three 250-NFE rollouts, but scores 4.12 on \sacson\ (\cref{tab:planning}).

\newsavebox{\infleg}
\sbox{\infleg}{\scriptsize%
\tikz[baseline=-0.6ex]{\draw[cmono, line width=0.9pt] (0,0) -- (0.42,0); \fill[cmono] (0.21,0) +(-1.6pt,-1.6pt) rectangle +(1.6pt,1.6pt);}\,\mono\hspace{7pt}%
\tikz[baseline=-0.6ex]{\draw[cM4, line width=0.9pt] (0,0) -- (0.42,0); \fill[cM4] (0.21,0) circle (1.7pt);}\,\M{4} (full)\hspace{7pt}%
\tikz[baseline=-0.6ex]{\draw[cM3, densely dashed, line width=0.9pt] (0,0) -- (0.42,0); \fill[cM3] (0.21,0) +(0,2.3pt) -- +(1.9pt,0) -- +(0,-2.3pt) -- +(-1.9pt,0) -- cycle;}\,\M{3}, \M{2} (gen.\ only)\hspace{7pt}%
\tikz[baseline=-0.6ex]{\draw[cM1, line width=0.9pt] (0,0) -- (0.42,0); \fill[cM1] (0.21,0) +(0,2.3pt) -- +(-2.1pt,-1.4pt) -- +(2.1pt,-1.4pt) -- cycle;}\,\M{1} (smallest)\hspace{7pt}%
\tikz[baseline=-0.6ex]{\fill[black!22, draw=cmono!90!black, line width=0.62pt] (0,0) +(0,2.5pt) -- +(2.0pt,0) -- +(0,-2.5pt) -- +(-2.0pt,0) -- cycle; \fill[cM4!30!white, draw=cM4!90!black, line width=0.62pt] (0.15,0) +(0,2.5pt) -- +(2.0pt,0) -- +(0,-2.5pt) -- +(-2.0pt,0) -- cycle; \fill[m4col, draw=cM1!85!black, line width=0.68pt] (0.30,0) +(0,2.6pt) -- +(2.1pt,0) -- +(0,-2.6pt) -- +(-2.1pt,0) -- cycle;}\,$2{\times}2$ NFEs\hspace{7pt}%
\tikz[baseline=-0.6ex]{\draw[black!70, densely dashed, line width=0.72pt] (0,0) -- (0.42,0);}\,Straight line (train-fit)\hspace{7pt}%
\tikz[baseline=-0.6ex]{\draw[black!55, densely dotted, line width=0.70pt] (0,0) -- (0.42,0);}\,\cem prior (\recon only)}
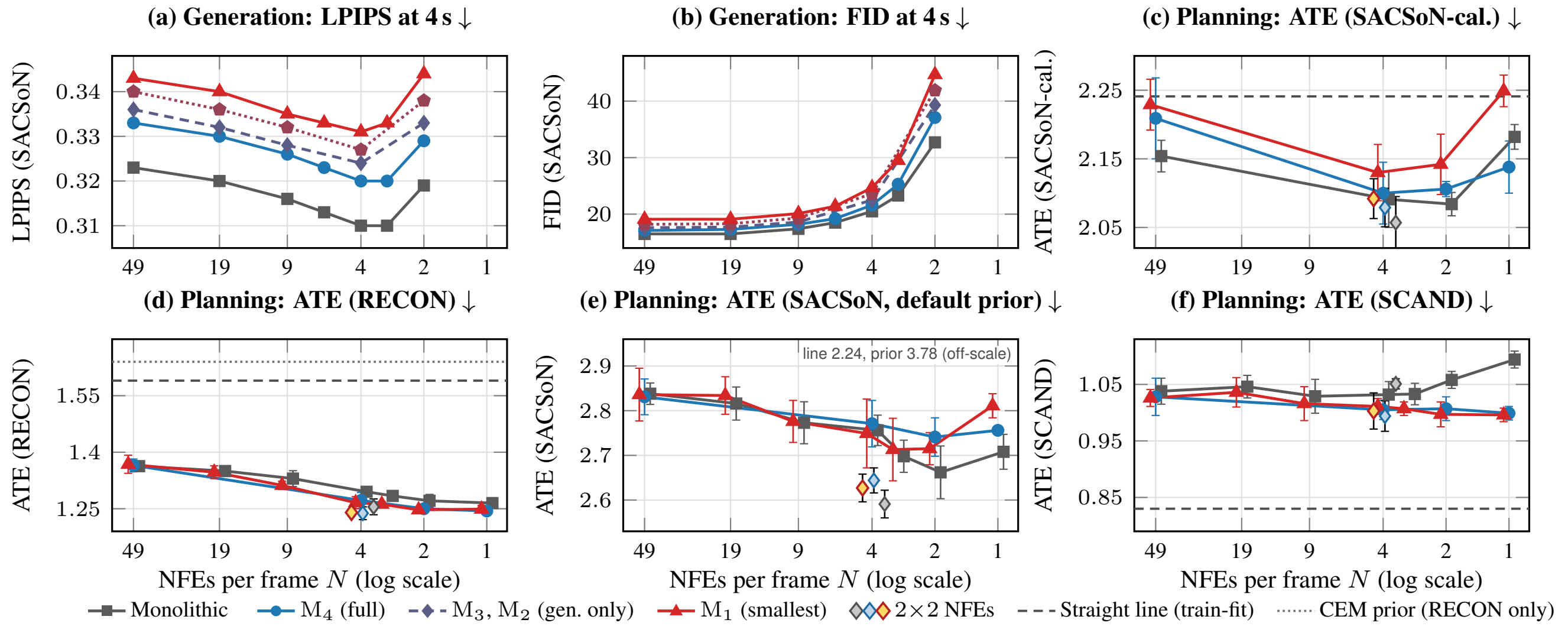
\begin{figure*}[t]
\centering
\begin{tikzpicture}
\begin{groupplot}[
  group style={group size=3 by 2, horizontal sep=1.30cm, vertical sep=1.02cm, xlabels at=edge bottom},
  scale only axis, width=4.42cm, height=2.15cm,
  xmode=log, log basis x=10, xmin=0.80, xmax=62,
  x dir=reverse, unbounded coords=discard,
  xlabel={NFEs per frame $N$ (log scale)}, grid=major, grid style={black!12},
  every axis plot/.append style={line width=0.90pt},
  tick label style={font=\scriptsize}, label style={font=\footnotesize},
  xlabel style={font=\footnotesize, yshift=3.0pt},
  title style={font=\footnotesize\bfseries, at={(0.5,1.0)}, anchor=south, yshift=1.0pt, inner sep=1pt},
  ylabel near ticks, ylabel style={font=\footnotesize},
  smono/.style={cmono, mark=square*, mark size=1.5pt},
  sMf/.style={cM4, mark=*, mark size=1.6pt},
  sMt/.style={cM3, densely dashed, mark=diamond*, mark size=1.9pt, mark options={solid}},
  sMw/.style={cM2, densely dotted, mark=pentagon*, mark size=1.8pt, mark options={solid}},
  sMo/.style={cM1, mark=triangle*, mark size=2.0pt},
  sTwoMono/.style={only marks, mark=diamond*, mark size=2.2pt, mark options={fill=black!22, draw=cmono!90!black, line width=0.62pt}},
  sTwoMf/.style={only marks, mark=diamond*, mark size=2.2pt, mark options={fill=cM4!30!white, draw=cM4!90!black, line width=0.62pt}},
  sTwoMo/.style={only marks, mark=diamond*, mark size=2.4pt, mark options={fill=m4col, draw=cM1!85!black, line width=0.68pt}},
  ebar/.style={error bars/y dir=both, error bars/y explicit, error bars/error bar style={line width=0.50pt}, error bars/error mark options={rotate=90, mark size=1.3pt}},
]
\nextgroupplot[title={(a) Generation: \lpips at 4\,s $\downarrow$}, ymin=0.305, ymax=0.348, ylabel={\lpips (\sacson)}, xtick={1,2,4,9,19,49}, xticklabels={1,2,4,9,19,49}]
\addplot[smono] table[x=nfe, y=lpips] {fig/data/steps_mono.dat};
\addplot[sMf]   table[x=nfe, y=lpips] {fig/data/steps_M4.dat};
\addplot[sMt]   table[x=nfe, y=lpips] {fig/data/steps_M3.dat};
\addplot[sMw]   table[x=nfe, y=lpips] {fig/data/steps_M2.dat};
\addplot[sMo]   table[x=nfe, y=lpips] {fig/data/steps_M1.dat};
\nextgroupplot[title={(b) Generation: \fid at 4\,s $\downarrow$}, ymin=14, ymax=48, ylabel={\fid (\sacson)}, xtick={1,2,4,9,19,49}, xticklabels={1,2,4,9,19,49}]
\addplot[smono] table[x=nfe, y=fid] {fig/data/steps_mono.dat};
\addplot[sMf]   table[x=nfe, y=fid] {fig/data/steps_M4.dat};
\addplot[sMt]   table[x=nfe, y=fid] {fig/data/steps_M3.dat};
\addplot[sMw]   table[x=nfe, y=fid] {fig/data/steps_M2.dat};
\addplot[sMo]   table[x=nfe, y=fid] {fig/data/steps_M1.dat};
\nextgroupplot[title={(c) Planning: \ate (\sacson-cal.) $\downarrow$}, ymin=2.02, ymax=2.30, ytick={2.05,2.15,2.25}, ylabel={\ate (\sacson-cal.)}, ebar, xtick={1,2,4,9,19,49}, xticklabels={1,2,4,9,19,49}]
\draw[black!70, densely dashed, line width=0.72pt] (axis cs:62,2.241) -- (axis cs:0.80,2.241);
\addplot[smono]    table[x expr=0.94*\thisrow{nfe}, y=ate, y error=err] {fig/data/planning_sacson_cal_mono_3seed.dat};
\addplot[sMf]      table[x=nfe, y=ate, y error=err] {fig/data/planning_sacson_cal_M4_3seed.dat};
\addplot[sMo]      table[x expr=1.06*\thisrow{nfe}, y=ate, y error=err] {fig/data/planning_sacson_cal_M1_3seed.dat};
\addplot[sTwoMono] table[x expr=0.87*\thisrow{nfe}, y=ate, y error=err] {fig/data/planning_sacson_cal_mono_2draw.dat};
\addplot[sTwoMf]   table[x expr=0.98*\thisrow{nfe}, y=ate, y error=err] {fig/data/planning_sacson_cal_M4_2draw.dat};
\addplot[sTwoMo]   table[x expr=1.11*\thisrow{nfe}, y=ate, y error=err] {fig/data/planning_sacson_cal_M1_2draw.dat};
\nextgroupplot[title={(d) Planning: \ate (\recon) $\downarrow$}, ymin=1.19, ymax=1.70, ytick={1.25,1.40,1.55}, ylabel={\ate (\recon)}, ebar, xtick={1,2,4,9,19,49}, xticklabels={1,2,4,9,19,49}]
\draw[black!55, densely dotted, line width=0.70pt] (axis cs:62,1.64) -- (axis cs:0.80,1.64);
\draw[black!70, densely dashed, line width=0.72pt] (axis cs:62,1.59) -- (axis cs:0.80,1.59);

\addplot[smono]    table[x expr=0.94*\thisrow{nfe}, y=ate, y error=err] {fig/data/planning_recon_mono_3seed.dat};
\addplot[sMf]      table[x=nfe, y=ate, y error=err] {fig/data/planning_recon_M4_3seed.dat};
\addplot[sMo]      table[x expr=1.06*\thisrow{nfe}, y=ate, y error=err] {fig/data/planning_recon_M1_3seed.dat};
\addplot[sTwoMono] table[x expr=0.87*\thisrow{nfe}, y=ate, y error=err] {fig/data/planning_recon_mono_2draw.dat};
\addplot[sTwoMf]   table[x expr=0.98*\thisrow{nfe}, y=ate, y error=err] {fig/data/planning_recon_M4_2draw.dat};
\addplot[sTwoMo]   table[x expr=1.11*\thisrow{nfe}, y=ate, y error=err] {fig/data/planning_recon_M1_2draw.dat};

\nextgroupplot[title={(e) Planning: \ate (\sacson, default prior) $\downarrow$}, ymin=2.53, ymax=2.96, ytick={2.6,2.7,2.8,2.9}, ylabel={\ate (\sacson)}, ebar, xtick={1,2,4,9,19,49}, xticklabels={1,2,4,9,19,49}]
\node[font=\sffamily\fontsize{5.0}{5.5}\selectfont, text=black!70, fill=white, fill opacity=0.88, text opacity=1, inner sep=0.5pt, anchor=north east] at (rel axis cs:0.99,0.97) {line 2.24, prior 3.78 (off-scale)};
\addplot[smono]    table[x expr=0.94*\thisrow{nfe}, y=ate, y error=err] {fig/data/planning_sacson_mono_3seed.dat};
\addplot[sMf]      table[x=nfe, y=ate, y error=err] {fig/data/planning_sacson_M4_3seed.dat};
\addplot[sMo]      table[x expr=1.06*\thisrow{nfe}, y=ate, y error=err] {fig/data/planning_sacson_M1_3seed.dat};
\addplot[sTwoMono] table[x expr=0.87*\thisrow{nfe}, y=ate, y error=err] {fig/data/planning_sacson_mono_2draw.dat};
\addplot[sTwoMf]   table[x expr=0.98*\thisrow{nfe}, y=ate, y error=err] {fig/data/planning_sacson_M4_2draw.dat};
\addplot[sTwoMo]   table[x expr=1.11*\thisrow{nfe}, y=ate, y error=err] {fig/data/planning_sacson_M1_2draw.dat};

\nextgroupplot[title={(f) Planning: \ate (\scand) $\downarrow$}, ymin=0.79, ymax=1.13, ytick={0.85,0.95,1.05}, ylabel={\ate (\scand)}, ebar, xtick={1,2,4,9,19,49}, xticklabels={1,2,4,9,19,49}]
\draw[black!70, densely dashed, line width=0.72pt] (axis cs:62,0.83) -- (axis cs:0.80,0.83);

\addplot[smono]    table[x expr=0.94*\thisrow{nfe}, y=ate, y error=err] {fig/data/planning_scand_mono_3seed.dat};
\addplot[sMf]      table[x=nfe, y=ate, y error=err] {fig/data/planning_scand_M4_3seed.dat};
\addplot[sMo]      table[x expr=1.06*\thisrow{nfe}, y=ate, y error=err] {fig/data/planning_scand_M1_3seed.dat};
\addplot[sTwoMono] table[x expr=0.87*\thisrow{nfe}, y=ate, y error=err] {fig/data/planning_scand_mono_2draw.dat};
\addplot[sTwoMf]   table[x expr=0.98*\thisrow{nfe}, y=ate, y error=err] {fig/data/planning_scand_M4_2draw.dat};
\addplot[sTwoMo]   table[x expr=1.11*\thisrow{nfe}, y=ate, y error=err] {fig/data/planning_scand_M1_2draw.dat};
\end{groupplot}
\node[anchor=north, inner sep=0] at ($(group c2r2.south)+(0,-0.78)$) {\usebox{\infleg}};
\end{tikzpicture}
\vspace{1pt}
\caption{\textbf{Generation and planning across Euler NFEs $N$ per frame (shared log axis).} (a,b)~4\,s \sacson direct prediction for $N\in\{49,19,9,6,4,3,2\}$ ($N{=}1$ off-scale at \lpips\ 0.41--0.44 and \fid\ 112--142; \M{2}, \M{3} generation only). (c--f)~Three-seed \cem \ate\ (mean\,$\pm$\,sd, 500 items/dataset) for $N\in\{49,19,9,4,3,2,1\}$ on \M{1} and \mono\ (seeds 1--2 fused at $N\in\{19,9,3\}$) and $N\in\{49,4,2,1\}$ on \M{4} and \sacson-cal., plus $2{\times}2$ NFEs (3{,}840 NFEs per decision). On \scand and \sacson-cal.\ the straight line equals the prior.}
\label{fig:inference}
\end{figure*}

\paragraph{Capacity Floor and Standalone Alternatives.}
Two alternatives could make nesting unnecessary, and neither holds. Slicing \mono to \M{1} without joint training collapses both tasks (supplement), so the small sub-network must be trained for. A separately trained \M{1} plans as well as nested \M{1}, but it cannot generate at full granularity, so serving both tasks would require two models. Nor can the planner shrink much further: a standalone \M{0} ($0.19\times$ MACs) matches \M{1} on \recon and \scand but loses 0.17 \ate\ on 4-NFE \sacson planning (\cref{tab:planning}). \M{1} therefore sits near the capacity floor.

\begin{table}[t]
\centering
\scriptsize
\caption{\textbf{Decomposition of the planning speed-up.} Median compiled latency per decision on one A100 (\texttt{torch.compile}, bf16, $C{=}120$, 50 \sacson items; other execution stacks in supplement). \emph{Granularity speed-up}: \mono latency over \M{1} (or \M{4}) latency at the same NFEs (1.31$\times$ for single draws and 1.32$\times$ for the batched $2{\times}2$ rows). The headline factorizes as $15.4\times = 11.6\times$ (steps: 52.1\,s $\to$ 4.48\,s on \mono) $\times\,1.32\times$ (granularity: 4.48\,s $\to$ 3.39\,s). \emph{2 draws}: both draws batched at 240. \M{1} also lowers peak forward memory from 4.71 to 4.24\,GB (fused attention, batch 120).}
\label{tab:latency}
\setlength{\tabcolsep}{3.2pt}\renewcommand{\arraystretch}{0.95}
\begin{tabular}{@{}lc cc r c@{}}
\toprule
 & & Params & Rel. & Latency (s) & Granularity \\
Planner & NFE & (M) & MACs & $\downarrow$ & speed-up \\
\midrule
\mono & 49 & 315.7 & 1.00 & 52.1 & -- \\
\paper \M{4} & 49 & 315.7 & 1.00 & 53.2 & 0.98$\times$ \\
\paper \M{1} & 49 & 207.4 & 0.65 & 39.8 & 1.31$\times$ \\
\midrule
\mono & 4 & 315.7 & 0.08 & 4.40 & -- \\
\paper \M{4} & 4 & 315.7 & 0.08 & 4.44 & 0.99$\times$ \\
\paper \M{1} & 4 & 207.4 & 0.05 & 3.35 & 1.31$\times$ \\
\midrule
\mono (2 draws) & $2{\times}2$ & 315.7 & 0.08 & 4.48 & -- \\
\paper \M{1} (2 draws) & $2{\times}2$ & 207.4 & 0.05 & \textbf{3.39} & \textbf{1.32$\times$} \\
\midrule
\mono & 2 & 315.7 & 0.04 & 2.26 & -- \\
\paper \M{1} & 2 & 207.4 & 0.03 & 1.72 & 1.31$\times$ \\
\midrule
\mono & 1 & 315.7 & 0.02 & 1.19 & -- \\
\paper \M{1} & 1 & 207.4 & 0.01 & 0.91 & 1.31$\times$ \\
\bottomrule
\end{tabular}
\end{table}

\subsection{Ablation on Sampling Schedule}
\label{sec:res-steps}

We next sweep $N\in\{49,19,9,6,4,3,2,1\}$ NFEs per frame across generation and planning (\cref{fig:inference}).

\begin{figure}[t]
\centering
\includegraphics[width=\linewidth]{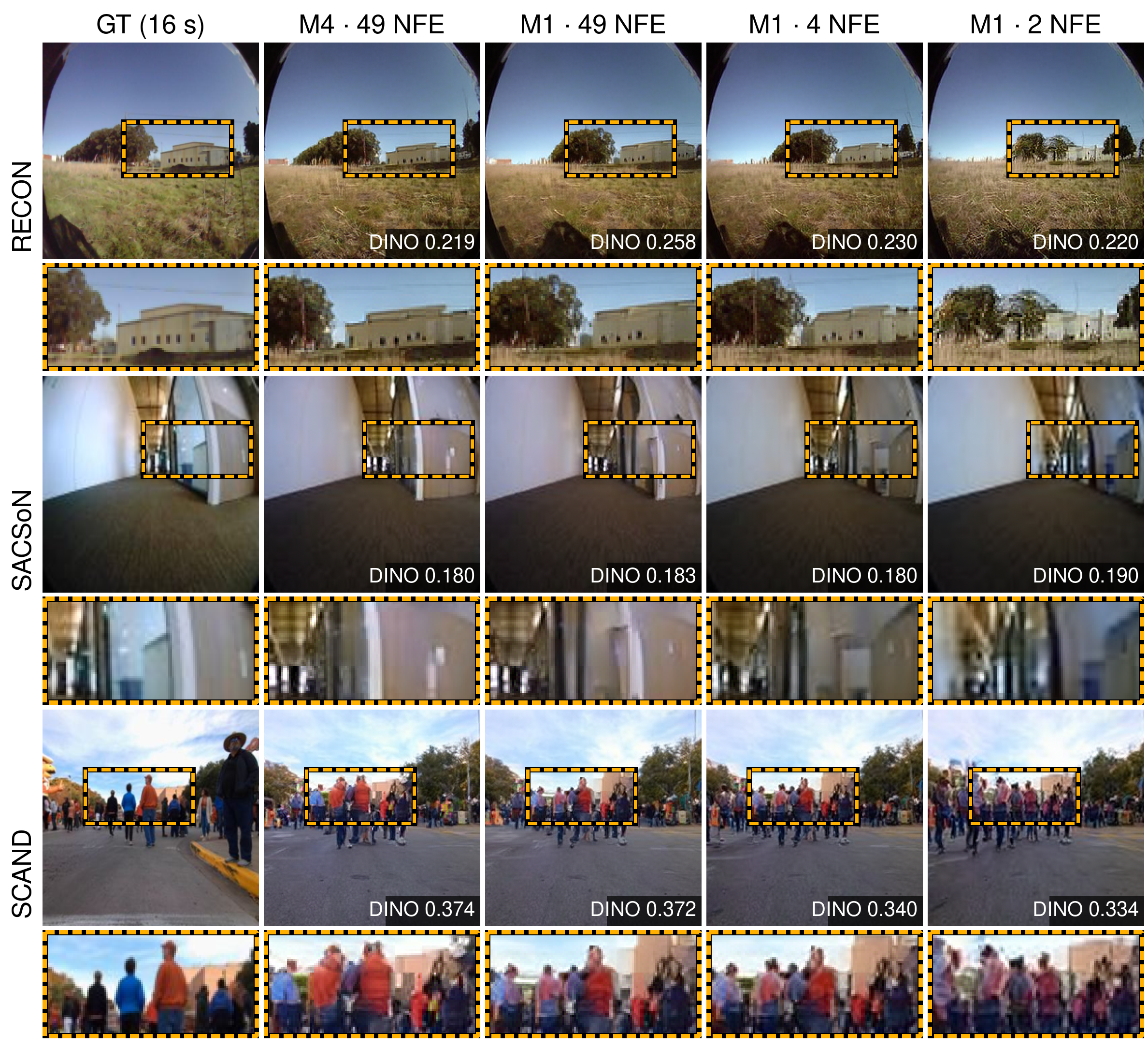}
\vspace{-12pt}
\caption{\textbf{Direct 16\,s prediction across granularity and NFEs} (four context frames; amber boxes: $2{\times}$ crops shown below each frame; insets: DINOv2 token distance to ground truth). Full and smallest granularity look alike at 49 NFEs. Reducing \M{1} to 4 NFEs preserves coarse geometry. At 2 NFEs, the setting used by the best planner, frames wash out (\M{1} \sacson \fid\ 44.7 at 4\,s and 52.0 at 16\,s; \cref{tab:sacson}).}
\label{fig:qual}
\end{figure}

\paragraph{Realism vs.\ Distortion Across Granularities.}
Fewer NFEs trade realism for lower distortion (\cref{tab:sacson,fig:inference}). 4\,s \sacson \fid\ holds down to 19 NFEs and collapses at 2 NFEs (\cref{fig:qual}). \lpips\ and \dino\ instead improve down to 2--4 NFEs. 4-NFE \M{1} reaches lower \dino\ than 49-NFE \M{4} (0.344 vs.\ 0.354) despite worse \fid. Coarse Euler steps from $t{=}1$ damp $\bm{\epsilon}$ toward the conditional expectation~\cite{blau2018perception}. The \M{1}--\M{4} offset stays nearly constant down to 4 NFEs and widens at 2 NFEs.

\paragraph{Fewer NFEs Improve Planning and Drive the Speed-Up.}
Coarse Euler improves planning on \recon and \sacson (\cref{fig:inference}). Cutting Euler from 49 to 4 NFEs lowers \ate\ by 0.06--0.14 (1.5--3.5\,cm) across models. Curvature quartiles rule out a straight-line bias. On \recon Q4, \M{1} gains 0.25 at 4 NFEs (\cref{tab:turns}). Two-round \cem\ and 4\,s horizons keep few-NFE parity or gains (supplement).

Step reduction drives the speed-up (\cref{tab:latency}). With compiled kernels, $2{\times}2$ NFEs cut the decision time of \mono from 52.1 to 4.48\,s ($11.6\times$). The $1.32\times$ granularity speed-up of \M{1} brings it to 3.39\,s, a total of $15.4\times$. At an iso-latency budget of about 3.4\,s, this granularity speed-up pays for the second draw. Within that budget \mono fits only a single 2-NFE draw (2.26\,s).

\begin{table}[t]
\centering
\scriptsize
\caption{\textbf{Inference controls of the sampler.} \cem \ate\ ($\downarrow$, 500 items/dataset). \emph{Fwd.}: network evaluations per decision. Top: three-seed means. \sig: paired 95\,\% CI vs.\ the shaded 4-NFE Euler reference excludes zero. Uniform-Euler seeds 1--2 use fused attention, and \mono \sacson mixes kernels (supplement). Bottom: seed-0 controls with common random numbers (CRN); $^{\dagger}$: vs.\ same-$N$ Euler.}
\label{tab:controls}
\setlength{\tabcolsep}{2.3pt}\renewcommand{\arraystretch}{0.95}
\resizebox{\linewidth}{!}{%
\begin{tabular}{@{}lr ccc ccc@{}}
\toprule
 & & \multicolumn{3}{c}{\paper \M{1}} & \multicolumn{3}{c}{\mono} \\
\cmidrule(lr){3-5}\cmidrule(l){6-8}
Sampler & Fwd. & \recon & \sacson & \scand & \recon & \sacson & \scand \\
\midrule
\multicolumn{8}{@{}l}{\emph{Three evaluation seeds (seeds 0--2)}} \\
\rowcolor{black!8} Euler ($N{=}4$, shift; reference) & 3{,}840 & 1.27 & 2.75 & 1.01 & 1.30 & 2.76 & 1.03 \\
Euler ($N{=}49$, shift) & 47{,}040 & 1.37\tsig & 2.84\tsig & 1.03 & 1.36\tsig & 2.84\tsig & 1.04 \\
Heun ($N{=}4$, 2 steps, shift) & 3{,}840 & 1.41\tsig & 2.80\tsig & 1.01 & 1.39\tsig & 2.81\tsig & 1.06\tsig \\
Midpoint ($N{=}4$, 2 steps, shift) & 3{,}840 & 1.37\tsig & 2.79\tsig & 1.03 & 1.38\tsig & 2.75 & 1.04 \\
Euler, $\alpha{=}1$ ($N{=}4$, unif.) & 3{,}840 & 1.28 & 2.73 & 1.00 & 1.31 & 2.69\tsig & 1.04 \\
Heun, $\alpha{=}1$ ($N{=}4$, unif.) & 3{,}840 & 1.34\tsig & 2.75 & 1.02 & 1.35\tsig & 2.71 & 1.04 \\
Midpoint, $\alpha{=}1$ ($N{=}4$, unif.) & 3{,}840 & 1.37\tsig & 2.74 & 1.03 & 1.39\tsig & 2.73 & 1.05 \\
Euler ($2{\times}N{=}2$, 2 draws) & 3{,}840 & 1.24\tsig & 2.63\tsig & 1.00 & 1.25\tsig & 2.59\tsig & 1.05 \\
Euler ($4{\times}N{=}1$, 4 draws) & 3{,}840 & 1.26 & 2.76 & 1.00 & 1.26\tsig & 2.66\tsig & 1.09\tsig \\
Euler ($N{=}1$, shift) & 960 & 1.25 & 2.81 & 1.00 & 1.26\tsig & 2.71 & 1.09\tsig \\
\midrule
\multicolumn{8}{@{}l}{\emph{Single-seed controls (seed 0, paired with seed-0 Euler)}} \\
Euler ($N{=}49$, s0) & 47{,}040 & 1.37\tsig & 2.78\tsig & 1.01 & 1.35\tsig & 2.85 & 1.01 \\
Euler ($N{=}49$, CRN) & 47{,}040 & 1.32 & 2.85 & 1.02 & 1.33 & 2.79 & 1.05 \\
Euler ($N{=}4$, s0) & 3{,}840 & 1.26 & 2.68 & 1.00 & 1.29 & 2.78 & 1.01 \\
Euler ($N{=}4$, CRN) & 3{,}840 & 1.26 & 2.75 & 1.02 & 1.29 & 2.65\tdag & 1.04 \\
\bottomrule
\end{tabular}}
\end{table}

\begin{table*}[t]
\centering
\scriptsize
\caption{\textbf{Planning accuracy (\ate $\downarrow$) stratified by trajectory curvature.} Straightest (Q1) vs.\ sharpest (Q4) quartiles of ground-truth lateral excursion $\max_{t\le 8}|y_t|$ in step-size units (125 items/quartile, three-seed mean). \emph{Prior}: straight-line \cem prior ($v_0{=}0.90$ on \sacson-cal.); $2{\times}2$: two-draw 2-NFE \M{1}; $\Delta_{4{-}49}$: $N{=}4$ minus $N{=}49$; $\Delta_{1{-}4}$: $N{=}1$ minus $N{=}4$; $\Delta_{2{\times}2{-}1}$: $2{\times}2$ minus $N{=}1$ (\sig: 95\,\% paired CI excludes zero; no paired intervals for calibrated Q1). Q2--Q3 are in the supplement.}
\label{tab:turns}
\setlength{\tabcolsep}{5.5pt}\renewcommand{\arraystretch}{1.0}
\makeatletter\let\tabinput\@@input\makeatother
\begin{tabular}{@{}ll c cc cccc ccc@{}}
\toprule
 & & & \multicolumn{2}{c}{\mono} & \multicolumn{4}{c}{\paper \M{1}} & \multicolumn{3}{c}{Paired $\Delta$ (\M{1})} \\
\cmidrule(lr){4-5}\cmidrule(lr){6-9}\cmidrule(l){10-12}
Dataset & Quartile ($\max|y|$, step units) & Prior & $N{=}49$ & $N{=}4$ & $N{=}49$ & $N{=}4$ & $2{\times}2$ & $N{=}1$ & $\Delta_{4{-}49}$ & $\Delta_{1{-}4}$ & $\Delta_{2{\times}2{-}1}$ \\
\midrule
\tabinput tab/gen/turns_main_body.tex
\bottomrule
\end{tabular}
\end{table*}

\subsection{Why Coarse Integration Helps Planning}
\label{sec:res-why}

To understand why coarse integration benefits planning, we first examine the search dynamics of \cem. When selecting the top 3 of 120 candidate trajectories based on costs derived from single stochastic rollouts, selecting the minimum cost inherently rewards lucky noise over truly superior actions---a classic manifestation of the optimizer's curse~\cite{smith2006optimizers}. The bias introduced by this effect scales directly with the noise-to-signal ratio of the estimated costs. A simple one-dimensional Gaussian model ($z\sim\mathcal{N}(0,\sigma^2)$) illustrates how ODE samplers can control this noise. Under a shifted Euler grid with $\sigma{=}0.5$, reducing the integration steps contracts the output variance to $0.049\sigma^2$ at 4 NFEs and to $0.007\sigma^2$ at 2 NFEs, whereas second-order solvers at 4 NFEs retain a much higher variance of $0.76\text{--}0.98\sigma^2$ (see supplement). This mathematical analysis leads to three key predictions: (1) coarse Euler integration should mitigate selection optimism and improve planning rankings; (2) second-order solvers at matched compute should fail to provide this benefit; and (3) an extreme reduction to 1-NFE should collapse the sampler to the conditional mean, exchanging variance for severe bias. Our experimental results systematically verify these predictions. Specifically, coarse Euler lowers selection optimism from 1.22 to 0.39 signal standard deviations (\cref{fig:mechanism}a,c). Meanwhile, Heun and midpoint solvers at matched compute never outperform 4-NFE Euler and merely reproduce standard 49-NFE rankings (\cref{tab:controls}). Finally, while 1-NFE uniform-scaling reduces overall costs, it alters critical trajectory rankings and consequently degrades performance on sharp indoor turns (\cref{tab:turns}).

\definecolor{cTeal}{HTML}{008080}
\newsavebox{\mechleg}
\sbox{\mechleg}{\fontsize{6.2}{7.0}\selectfont%
\tikz[baseline=-0.6ex]{\draw[cM1, line width=0.9pt] (0,0) -- (0.40,0); \fill[cM1] (0.20,0) +(0,2.1pt) -- +(-1.9pt,-1.3pt) -- +(1.9pt,-1.3pt) -- cycle;}\,Euler \sacson\hspace{5pt}%
\tikz[baseline=-0.6ex]{\draw[cM1!70!black, densely dashed, line width=0.9pt] (0,0) -- (0.40,0); \draw[cM1!70!black, fill=white, line width=0.6pt] (0.20,0) +(0,2.1pt) -- +(-1.9pt,-1.3pt) -- +(1.9pt,-1.3pt) -- cycle;}\,Euler \recon\hspace{5pt}%
\tikz[baseline=-0.6ex]{\fill[cM4, draw=cM4!70!black, line width=0.5pt] (0,0) +(-1.6pt,-1.6pt) rectangle +(1.6pt,1.6pt);}\,Heun (4 NFE)\hspace{5pt}%
\tikz[baseline=-0.6ex]{\fill[cTwo, draw=cTwo!70!black, line width=0.5pt] (0,0) +(0,2.3pt) -- +(1.9pt,0) -- +(0,-2.3pt) -- +(-1.9pt,0) -- cycle;}\,Midpoint (4 NFE)}
\begin{figure}[t]
\centering
\begin{tikzpicture}
\begin{groupplot}[
  group style={group size=2 by 2, horizontal sep=1.16cm, vertical sep=2.55cm},
  scale only axis, width=2.70cm, height=1.62cm,
  grid=major, grid style={black!12},
  every axis plot/.append style={line width=0.9pt},
  tick label style={font=\scriptsize}, label style={font=\footnotesize}, title style={font=\fontsize{7.6}{8.4}\selectfont\bfseries, yshift=-1.5pt},
  xlabel style={font=\footnotesize, yshift=2.5pt},
  ylabel near ticks, ylabel style={font=\footnotesize},
  legend image code/.code={\draw[mark repeat=2,mark phase=2,#1] plot coordinates {(0cm,0cm) (0.14cm,0cm) (0.28cm,0cm)};},
  legend style={font=\fontsize{6.2}{7.0}\selectfont, draw=none, fill=none, cells={anchor=west}, inner sep=0.6pt, row sep=-0.8pt, /tikz/every even column/.append style={column sep=3pt}},
  euS/.style={cM1, mark=triangle*, mark size=1.8pt},
  euR/.style={cM1!70!black, densely dashed, mark=triangle*, mark size=1.8pt, mark options={solid, fill=white}},
  heun/.style={only marks, mark=square*, mark size=1.6pt, mark options={fill=cM4, draw=cM4!70!black, line width=0.5pt}},
  midp/.style={only marks, mark=diamond*, mark size=2.2pt, mark options={fill=cTwo, draw=cTwo!70!black, line width=0.5pt}},
]
\nextgroupplot[title={(a) Rank $\rho$ (\sacson) $\uparrow$}, xmode=log, log basis x=10, xmin=0.82, xmax=62, xtick={1,2,4,49}, xticklabels={1,2,4,49}, x dir=reverse,
  xlabel={NFEs per frame $N$}, ylabel={Spearman $\rho$}, ymin=0.41, ymax=1.00, ytick={0.5,0.6,0.7,0.8,0.9,1.0},
  legend columns=1, legend style={at={(0.50,-0.53)}, anchor=north}]
\draw[black!55, densely dashed, line width=0.65pt] (axis cs:62,0.52) -- (axis cs:0.82,0.52);
\node[font=\sffamily\fontsize{6.0}{6.6}\selectfont, text=black!80, fill=white, fill opacity=0.90, text opacity=1, anchor=north, inner sep=0.6pt] at (axis cs:7.0,0.505) {49-NFE self: 0.52};
\addplot[cTwo!85!black, mark=diamond*, mark size=1.6pt] table[x=nfe, y=rho] {fig/data/mech_retest_sacson.dat}; \addlegendentry{Test--retest}
\addplot[cM4, densely dashed, mark=*, mark size=1.4pt, mark options={solid}] table[x=nfe, y=rho] {fig/data/mech_shared_M4.dat}; \addlegendentry{\M{4} vs.\ 49-NFE \M{4}}
\addplot[cM1!80!black, densely dashdotted, mark=star, mark size=1.8pt, mark options={solid}] table[x=nfe, y=rho] {fig/data/mech_shared_M1.dat}; \addlegendentry{\M{1} vs.\ 49-NFE \M{4}}

\nextgroupplot[title={(b) Multi-draw \ate\ vs.\ NFEs $\downarrow$}, xmode=log, log basis x=10, xmin=740, xmax=185000,
  xtick={960,3840,11520,47040,141120}, xticklabels={1k,3.8k,12k,47k,141k},
  xlabel={Total NFEs / decision}, ylabel={\ate (\sacson)}, ymin=2.53, ymax=2.85, ytick={2.6,2.7,2.8},
  error bars/y dir=both, error bars/y explicit, error bars/error bar style={line width=0.45pt}, error bars/error mark options={rotate=90, mark size=1.0pt},
  legend columns=2, legend style={at={(0.50,-0.53)}, anchor=north}]
\addplot[black!60, densely dotted, mark=o, mark size=1.35pt, mark options={solid}] table[x=nfe_tot, y=ate] {fig/data/kavg_M1at50_500.dat}; \addlegendentry{49 NFE (seed 0)}
\addplot[cTeal, densely dashed, mark=pentagon*, mark size=1.45pt, mark options={solid}] table[x expr=0.94*\thisrow{nfe_tot}, y=ate, y error=err] {fig/data/kavg_M1at5_500.dat}; \addlegendentry{4 NFE}
\addplot[cmono, mark=square*, mark size=1.35pt] table[x=nfe_tot, y=ate, y error=err] {fig/data/kavg_M1at2_500.dat}; \addlegendentry{1 NFE}
\addplot[cM1, line width=1.0pt, mark=triangle*, mark size=1.8pt] table[x expr=1.06*\thisrow{nfe_tot}, y=ate, y error=err] {fig/data/kavg_M1at3_500.dat}; \addlegendentry{2 NFE}
\addplot[only marks, mark=diamond*, mark size=2.1pt, mark options={fill=black!22, draw=cmono!90!black, line width=0.6pt}] coordinates {(3250,2.591) +- (0,0.031)}; \addlegendentry{Mono.\ $2{\times}2$}
\addplot[only marks, mark=diamond*, mark size=2.3pt, mark options={fill=m4col, draw=cM1!85!black, line width=0.65pt}] coordinates {(4150,2.627) +- (0,0.031)}; \addlegendentry{\M{1} $2{\times}2$}

\nextgroupplot[title={(c) Selection optimism $\downarrow$}, xmode=log, log basis x=10, xmin=0.82, xmax=62, xtick={1,2,4,49}, xticklabels={1,2,4,49}, x dir=reverse,
  xlabel={NFEs per frame $N$}, ylabel={$\Delta s_{\mathrm{curse}}/\hat{\sigma}_{\mathrm{signal}}$}, ymin=-0.05, ymax=1.35, ytick={0,0.4,0.8,1.2}]
\addplot[euS] coordinates {(49,1.22) (4,0.73) (2,0.39) (1,0.08)};
\addplot[euR] coordinates {(49,1.21) (4,0.69) (2,0.36) (1,0.04)};
\addplot[heun] coordinates {(4.75,0.49)};
\addplot[midp] coordinates {(3.40,0.76)};

\nextgroupplot[title={(d) $r_{\mathrm{true}}$ vs.\ 49-NFE \M{1} $\uparrow$}, xmode=log, log basis x=10, xmin=0.82, xmax=62, xtick={1,2,4,49}, xticklabels={1,2,4,49}, x dir=reverse,
  xlabel={NFEs per frame $N$}, ylabel={$r_{\mathrm{true}}$}, ymin=0.66, ymax=1.02, ytick={0.7,0.8,0.9,1.0}]
\addplot[euS] coordinates {(49,1.00) (4,0.95) (2,0.86) (1,0.75)};
\addplot[euR] coordinates {(49,1.00) (4,0.91) (2,0.79) (1,0.70)};
\addplot[heun] coordinates {(4.75,0.96)};
\addplot[midp] coordinates {(3.40,0.98)};
\end{groupplot}
\node[anchor=north, inner sep=0] at ($(group c1r2.south)!0.5!(group c2r2.south)+(0,-0.86)$) {\usebox{\mechleg}};
\end{tikzpicture}
\vspace{-4pt}
\caption{\textbf{Variance contraction and multi-draw efficiency} on default-prior \sacson. (a)~Coarse Euler makes independent draws rank the same 120 candidates more consistently (test--retest $\rho$). Agreement with 49-NFE \M{4} (100 items, shared initial noise) stays above the 49-NFE self-agreement. (b)~\ate\ of \M{1} (lines) vs.\ total NFEs per decision for 1--4 averaged draws (three seeds; seed 0 for 49-NFE \M{1}). At 3{,}840 NFEs, two 2-NFE draws beat one 4-NFE draw and four 1-NFE draws. They do not differ significantly from three 49-NFE draws, which cost $37\times$ more. (c)~Held-out selection optimism of \M{1} on \sacson and \recon falls toward zero as Euler coarsens, while 4-NFE Heun and midpoint re-inflate it (\sacson markers). (d)~Noise-corrected rank correlation with 49-NFE \M{1}. Coarse Euler departs from the dense ranking, while 4-NFE Heun and midpoint stay close to it.}
\label{fig:mechanism}
\end{figure}
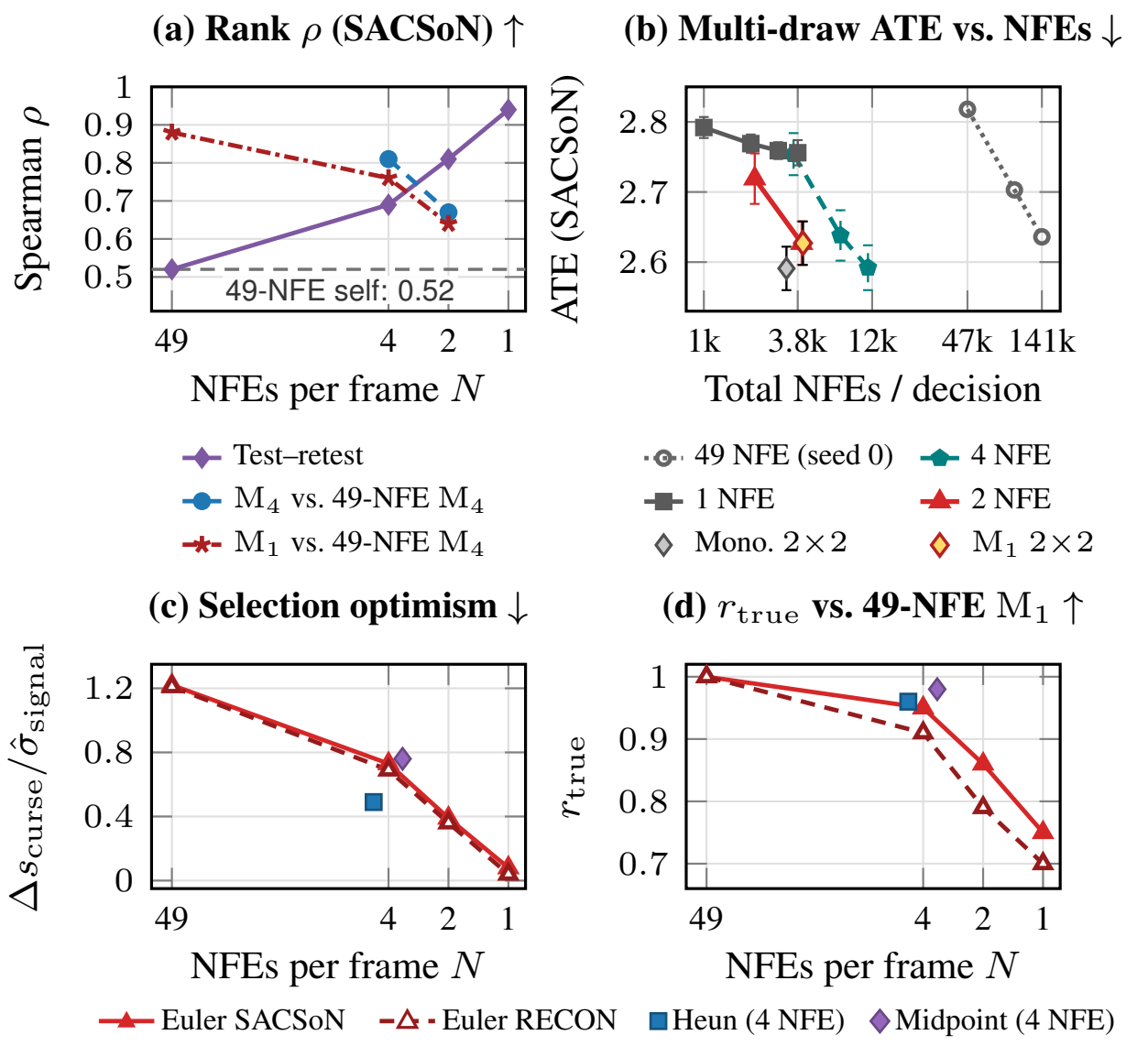

\paragraph{Coarse Euler Reduces Selection Optimism.}
Coarse Euler lowers both the noise-to-signal ratio and the measured selection optimism. We re-score the fixed pool of $C{=}120$ candidates of each item $i$ across $M$ draws. Let $s_{i,c,r}$ be the terminal cosine DINO distance of candidate $c$ in draw $r$, a cost where lower is better. With $\bar{s}_{i,c}=\frac{1}{M}\sum_{r}s_{i,c,r}$ and the Bessel-corrected variance $\widehat{\mathrm{Var}}_c(\bar{s}_{i,c})$ over candidates, we define
\begingroup\abovedisplayskip=6pt plus 2pt minus 3pt\belowdisplayskip=6pt plus 2pt minus 3pt\abovedisplayshortskip=6pt plus 2pt minus 3pt\belowdisplayshortskip=6pt plus 2pt minus 3pt
\begin{equation}
\begin{aligned}
\hat{\sigma}_{\mathrm{noise},i}^2 &= \frac{1}{C(M-1)}\sum_{c=1}^{C}\sum_{r=1}^{M}(s_{i,c,r}-\bar{s}_{i,c})^2, \\
\hat{\sigma}_{\mathrm{signal},i}^2 &= \widehat{\mathrm{Var}}_c(\bar{s}_{i,c})-\hat{\sigma}_{\mathrm{noise},i}^2/M,
\end{aligned}
\label{eq:ns}
\end{equation}
\endgroup
and the dataset-RMS ratio $\hat{\lambda}=\sqrt{\overline{\hat{\sigma}_{\mathrm{noise}}^2}/\overline{\hat{\sigma}_{\mathrm{signal}}^2}}$, where bars average over items. On \sacson \M{1}, $\hat{\lambda}$ drops from 0.90 at 49 NFEs to 0.44 at 2 NFEs. Consistent with this drop, two independent draws rank the same candidates more alike as NFEs decrease (test--retest $\rho$ from 0.52 to 0.81; \cref{fig:mechanism}a). Coarse-NFE rankings also stay closer to 49-NFE rankings than two 49-NFE draws are to each other. The high 1-NFE agreement ($\rho{=}0.94$) reflects consistency rather than correctness, because a single step collapses to the conditional mean (see Single-Step Prediction Shifts Rankings). We measure selection optimism~\cite{smith2006optimizers} $\Delta s_{\mathrm{curse}}/\hat{\sigma}_{\mathrm{signal}}$ as the held-out minus selecting-draw cost of the top-3 elites. On \sacson, it falls from $+1.22$ at 49 NFEs to $+0.39$ at 2 NFEs ($+1.21$ to $+0.36$ on \recon; \cref{fig:mechanism}c). A second-order correction on the 2-step grid re-inflates optimism (\cref{fig:mechanism}c). Sharing noise across candidates via common random numbers (CRN)~\cite{glasserman1992crn} does not help at 49 NFEs (\cref{tab:controls}). We attribute this to decorrelation. After $8{\times}49$ nonlinear action-conditioned evaluations, shared initial noise no longer produces shared errors.

\paragraph{Second-Order Solvers Keep the Variance.}
Heun~\cite{karras2022edm} and midpoint~\cite{lipman2023flow} improve realism but not ranking. They recover most of the 4-NFE \fid\ loss. In planning, they never significantly beat 4-NFE Euler across 24 pairs (2 models $\times$ 3 datasets $\times$ 2 solvers $\times$ 2 time grids; \cref{tab:controls}). They also reproduce 49-NFE rankings ($r_{\mathrm{true}}{=}0.94\text{--}0.99$; \cref{fig:mechanism}d). Here $r_{\mathrm{true}}$ is the noise-corrected correlation between costs from distinct noise seeds. On the uniform grid, 2-step midpoint evaluates the network at the same four times as 4-NFE Euler but gives zero weight to $t{=}1$.

\paragraph{Averaging Draws Is Less Efficient.}
Repeating dense rollouts contracts variance far less than coarse Euler does. On \recon, seed-0 3-draw 49-NFE \M{1} still trails 3-draw 4-NFE \M{1} significantly (0.09 \ate). Three draws cut the noise variance only by a factor of three, whereas 4-NFE Euler contracts it to $0.049\sigma^2$ in the Gaussian model. On \sacson, $2{\times}2$-NFE \M{1} matches 3-draw 49-NFE \M{1} at $37\times$ fewer NFEs. At fixed compute, two 2-NFE draws balance selection optimism against 1-NFE ranking shift (\cref{fig:mechanism}b).

\paragraph{Single-Step Prediction Shifts Rankings.}
One NFE returns the conditional mean and shifts rankings on turns. As $\bm{z}_1{=}\bm{\epsilon}$ is independent of $\bm{z}_{i+k}$ given $\bm{c}_{\mathrm{ctx}}{=}(\bm{z}_{i-3:i},\bm{a},k)$, the minimizer of \cref{eq:fm} at $t{=}1$ is $\bm{v}^\star(\bm{\epsilon},1,\bm{c}_{\mathrm{ctx}}){=}\bm{\epsilon}-\mathbb{E}[\bm{z}_{i+k}\mid\bm{c}_{\mathrm{ctx}}]$. A single Euler step of size 0.993 thus forms $\hat{\bm{z}}_{t_1}\approx 0.007\,\bm{\epsilon}+0.993\,\mathbb{E}[\bm{z}_{i+k}\mid\bm{c}_{\mathrm{ctx}}]$. In token space, this lowers costs uniformly across all 120 candidates. It also changes the ranking ($r_{\mathrm{true}}{=}0.75$ vs.\ 49-NFE \M{1}; \cref{fig:mechanism}d) and degrades indoor turns. On \sacson, $4{\times}1$-NFE \M{1} trails $2{\times}2$-NFE \M{1} by 0.13 at equal compute (\cref{tab:controls}). A same-recipe deterministic regressor, which targets the conditional mean directly, plans worse than 1-NFE \M{1} on all three datasets (\cref{tab:planning,tab:controls}; its \recon error rises during training). Few-NFE elites are not safer candidates. Re-scored on fresh 49-NFE draws unused in selection, their decoupled cost variance stays within $[-0.07,+0.05]$ of that of 49-NFE elites.

\section{Limitations}
\label{sec:limitations}

\smallskip\noindent\textbf{Open-Loop Evaluation.}
All planning results are open-loop against recorded 2\,s and 4\,s paths under terminal DINO distance. Closed-loop execution and collisions past 4\,s remain unmeasured. No model beats the straight-line prior on \scand, and on calibrated \sacson the straight line stays lower on Q1 corridors and step-wise \rpe\ (\cref{sec:res-plan}).

\smallskip\noindent\textbf{Statistical Scope.}
\mono itself shifts by up to 0.09 \ate\ across two training seeds, so smaller gaps between models lie within seed variation (supplement).

\smallskip\noindent\textbf{Architectural and Latency Scope.}
\paper nests only FFN granularity ($0.65\text{--}1.0\times$ MACs), while standalone \M{0} probes a joint reduction of depth and hidden size. At $C{=}120$ on one A100, compiled $2{\times}2$-NFE \M{1} takes 3.39\,s per 2\,s plan (\cref{tab:latency}), about 0.3\,Hz and far from real-time control. Using $C\le 64$ costs $+0.16$\sig\ to $+0.40$\sig\ indoor \ate.

\section{Conclusion}
\label{sec:conclusion}

Open-loop planning needs low-variance cost estimates rather than the sharp samples that generation requires. Coarse flow-matching integration supplies them at no training cost. Using \paper as a controlled probe, we show that visual realism requires full granularity and many NFEs, whereas planning at the smallest granularity matches full granularity and improves under a few Euler steps. We trace this gain to the optimizer's curse. As a result, one trained model renders full-granularity frames and plans $15.4\times$ faster than standard 49-NFE planning with no loss in accuracy ($11.6\times$ from the sampler alone). World-model planners can choose their sampler for planning rather than inherit it from generation.

\label{endofmain}
\par
{
    \small
    \bibliographystyle{ieeenat_fullname}
    \bibliography{ref}
}

\clearpage
\setcounter{section}{0}
\renewcommand{\thesection}{\Alph{section}}
\renewcommand{\theHsection}{supp.\Alph{section}}
\setcounter{table}{0}
\renewcommand{\thetable}{S\arabic{table}}
\renewcommand{\theHtable}{S\arabic{table}}
\setcounter{figure}{0}
\renewcommand{\thefigure}{S\arabic{figure}}
\renewcommand{\theHfigure}{S\arabic{figure}}
\setcounter{equation}{0}
\renewcommand{\theequation}{S\arabic{equation}}
\renewcommand{\theHequation}{S\arabic{equation}}
\makeatletter\setlength{\@fptop}{0pt}\makeatother
\renewcommand{\dbltopfraction}{0.90}
\renewcommand{\dblfloatpagefraction}{0.85}

\maketitlesupplementary

\noindent\textbf{Contents}\par\vspace{2pt}
{\small
\suppline{sec:s-impl}
\suppline{sec:s-gen}
\suppline{sec:s-plan}
\suppline{sec:s-controls}
\suppline{sec:s-noise}
\suppline{sec:s-nesting}
\suppline{sec:s-axes}
\suppline{sec:s-neg}
\suppline{sec:s-seed1}
\suppline{sec:s-reg}
\suppline{sec:s-qual}
}\medskip

This supplementary material complements the main paper. Unprefixed section, table, figure, and equation numbers refer to the main paper; in the generated tables below, All-\M{g} denotes single-granularity evaluation of \paper \M{g} across all rollout steps (contrasting with two-stage \M{1}$\to$\M{4} screening). The main paper indexes every sampling schedule by its number $N$ of Euler NFEs per frame, whereas the text and tables below index the shifted solver grid by its $S{=}N{+}1$ points (written $S{=}50$, $@50$, \at{1}{50}, or \monoat{50}): $S{=}50$ corresponds to 49 NFEs, $S{=}5$ to 4 NFEs, $S{=}3$ to 2 NFEs, the two-draw $2{\times}S{=}3$ planner (also $2{\times}@3$ or $2{\times}\at{1}{3}$) to the $2{\times}2$-NFE planner of the main paper, and $S{=}2$ to 1 NFE. Every number is computed from the per-sample or per-item outputs of the evaluation harness by the scripts released with the code; where the main text reports a rounded value, the table here gives the same quantity at full precision.

\medskip
\noindent\textbf{Guide from Main-Paper Claims to Supplement Tables:}
\begin{itemize}
\item \textbf{Architecture \& statistical calibration:} sub-network MACs (\cref{tab:s-widths}), single-seed null calibration (\cref{tab:s-nullcal}), and trajectory cluster bootstrap (\cref{tab:s-cluster}).
\item \textbf{Generation across granularities, horizons, \& solvers:} full-split 64-frame rollouts and resize audit (\cref{tab:s-rollout-1024}), 4\,s schedule sweep (\cref{tab:s-gen-full}), 1--16\,s horizons (\cref{tab:s-gen-horizon}), granularity vs.\ steps (\cref{tab:widthsteps}), nesting cost (\cref{tab:paired}), and solver controls (\cref{tab:s-gen-controls}).
\item \textbf{Planning seeds, paired CIs, \& task calibration:} seed-0 schedule grid (\cref{tab:s-plan-full}), three-seed runs (\cref{tab:s-seeds}), three-seed paired CIs (\cref{tab:s-pooled,tab:s-e12}), model-free baselines (\cref{tab:s-calib}), and speed-calibrated \sacson (\cref{tab:s-b2-cal}).
\item \textbf{Curvature, iterative \cem, \& 4\,s horizons:} difficulty halves (\cref{tab:s-difficulty}), all 12 curvature quartiles (\cref{tab:s-turns}), Q4 qualitative \cem proposals (\cref{fig:s-cem-qual}), and 2-round \cem / 16-step horizon (\cref{tab:s-horizon-cem}).
\item \textbf{Sampler controls \& GPU latency:} seed-0 solver/CRN/\,$\bm{\epsilon}{=}\bm{0}$ matrix and CIs (\cref{tab:s-controls-full,tab:s-controls-ci}), matched compiled and GPU-noise latency (\cref{tab:s-sdpa-shard}), kernel timings (\cref{tab:s-kernel-bench}), candidate-pool scaling (\cref{tab:s-realtime}), and alternative planning costs (\cref{tab:s-costfn}).
\item \textbf{Score variance, Gaussian ODE, \& ablations:} exact Gaussian ODE variance (\cref{tab:s-gauss-ode}), noise-to-signal and multi-draw \ate\ (\cref{tab:noise,tab:s-r4-r5-ci}), direct winner's-curse and decoupled uncertainty (\cref{tab:s-b0-curse}), 1-shot token vs.\ pixel error (\cref{tab:s-b0-oneshot}), standalone \M{1} and \M{0} (\cref{sec:s-nesting}), and negative results (\cref{tab:negatives}).
\item \textbf{Second training seed \& same-recipe DINO-Reg:} seed-1 planning and generation (\cref{tab:s-seed1-plan,tab:s-seed1-gen}) with within-seed, step-reduction, and cross-seed paired CIs (\cref{tab:s-seed1-ci}), regressor planning with sd, \rpe, and training progression (\cref{tab:s-reg-plan}), curvature quartiles (\cref{tab:s-reg-turns}), generation (\cref{tab:s-reg-gen}), and one-shot vs.\ rollout token error (\cref{tab:s-reg-tok}).
\item \textbf{Ranked qualitative panels:} 16\,s rollouts with few-NFE rows (\cref{fig:s-roll-sacson,fig:s-roll-recon,fig:s-roll-scand}) and Q4 \cem plans (\cref{fig:s-cem-recon,fig:s-cem-sacson,fig:s-cem-counter}).
\end{itemize}

\section{Implementation and Protocol Details}
\label{sec:s-impl}

\paragraph{Representation.}
Frames are resized to $224{\times}224$ and encoded by a frozen DINOv2-B encoder~\cite{oquab2024dinov2} into $16{\times}16{=}256$ patch tokens of dimension 768. Tokens are normalized by the per-channel statistics of a representation autoencoder (RAE)~\cite{zheng2026rae}, whose frozen decoder maps predicted tokens back to pixels for visualization and for the pixel-level metrics; the decoder is never trained or fine-tuned in this work.

\paragraph{Backbone and Nested Granularities.}
The trunk has 12 blocks of hidden size 768 with 12 attention heads, each with self-attention over the 256 target tokens, cross-attention to the $4{\times}256$ context tokens, and a feed-forward layer of width 3072; the decoupled head~\cite{wang2026ddt} has 2 blocks of hidden size 2048 with 16 heads and feed-forward width 8192. All blocks are standard DiT blocks~\cite{peebles2023dit} with LayerNorm and GELU, conditioned through adaptive layer normalization on the output of a dynamics conditioning module~\cite{zhang2026raenwm} that gates the embedded action $\bm{a}$, horizon $k$, and flow time $t$; normalized trunk tokens $\mathrm{LN}(\bm{x})$ enter the decoupled head via both a $768\to2048$ linear projection added to the head patch embedding and a per-token conditioning sum $\bm{s}=\mathrm{LN}(\bm{x})+\bm{c}$ feeding the head's adaptive layer normalization. The Matryoshka structure keeps the first $m_g$ hidden units of every feed-forward layer (\cref{tab:s-widths}); attention layers and linear projections are shared at full width by all sub-networks.

\begin{table}[t]
\centering\scriptsize
\caption{\textbf{Sub-networks of \paper.} Feed-forward widths of trunk and head, active parameters, and multiply--accumulate operations per network evaluation (one velocity evaluation at batch size 1).}
\label{tab:s-widths}
\setlength{\tabcolsep}{3.5pt}
\resizebox{\linewidth}{!}{%
\begin{tabular}{@{}lcccccc@{}}
\toprule
 & \multicolumn{2}{c}{FFN width $m_g$} & Active & & \multicolumn{2}{c}{Relative} \\
\cmidrule(lr){2-3}\cmidrule(lr){6-7}
Sub-network & trunk & head & params (M) & GMACs & params & MACs \\
\midrule
\M{1} & 384 & 1{,}024 & 207.4 & 51.4 & 0.66 & 0.65 \\
\M{2} & 768 & 2{,}048 & 222.8 & 55.3 & 0.71 & 0.70 \\
\M{3} & 1{,}536 & 4{,}096 & 253.8 & 63.2 & 0.80 & 0.80 \\
\M{4} (= monolithic) & 3{,}072 & 8{,}192 & 315.7 & 79.1 & 1.00 & 1.00 \\
\midrule
Standalone \M{0} (6L/384d, 2L/1024d) & 1{,}536 & 4{,}096 & 57.4 & 15.3 & 0.18 & 0.19 \\
\bottomrule
\end{tabular}}
\end{table}

\paragraph{Training.}
\paper and the monolithic baseline are trained with the identical recipe: 50 epochs of AdamW on the union of the training splits of \recon, \sacson, and \scand, a total batch of 96 context windows with four goal frames per window (384 targets per step), learning rate $2{\times}10^{-4}$ decayed linearly to $2{\times}10^{-6}$, no weight decay, conditional flow matching with the time distribution shifted by $\alpha{=}\sqrt{196{,}608/4{,}096}{=}\sqrt{48}{\approx}6.93$~\cite{esser2024sd3,zheng2026rae,zhang2026raenwm}, and an exponential moving average of the weights used for every evaluation. The only difference between the two models is the Matryoshka objective: at each optimizer step \paper draws one granularity $g\in\{1,2,3,4\}$ uniformly at random and back-propagates the loss of that sub-network, whereas the baseline always trains the full granularity. All \paper and \mono results outside \cref{sec:s-axes,sec:s-seed1} come from the primary runs (training seed 0), each trained on a single NVIDIA A100-SXM4-40GB GPU. Both models were also trained for the full 50 epochs with a second training seed (seed 1), evaluated in \cref{sec:s-seed1}. Every optimizer step accumulates the gradient of a 96-window batch (384 targets) over micro-batches whose size is set by the attention kernel and GPU memory: 12 windows in the primary runs (explicit attention kernel); 24 in the seed-1 baseline, the standalone \M{0}, and the DINO-Reg runs (fused attention, A100); and, in the seed-1 \paper run, 24 on an A100 for approximately the first 20 epochs and 48 on an NVIDIA H100 for the remainder. The batch per optimizer step, the optimizer, and the schedule are identical across runs, so cross-seed contrasts combine the training seed with these execution differences. The standalone \M{1} and \M{0} models (\cref{sec:s-nesting}) and the same-recipe DINO-Reg regressor (\cref{sec:s-reg}) have one training seed each.

\paragraph{Data Preparation.}
We follow the dataset splits of the benchmark~\cite{bar2025nwm,zhang2026raenwm}, partitioned at the trajectory level; the test split comprises 897 trajectories and 9{,}561 evaluation samples (4{,}262 \recon, 3{,}416 \sacson, 1{,}883 \scand). Our preparation differs from the published \raenwm pipeline in two respects that affect absolute numbers while leaving all paired comparisons in this paper unaffected: \sacson frames are stored at $160{\times}120$ pixels and upsampled to 224 (the published models use frames 320 pixels wide), and our backbone uses standard DiT blocks with LayerNorm and GELU in place of RMSNorm, SwiGLU and rotary embeddings (315.7\,M against 350\,M parameters).

\paragraph{Sampler.}
At inference the velocity field is integrated with a first-order Euler solver on the shifted grid $t_j=\alpha u_j/(1+(\alpha-1)u_j)$, $u_j=1-j\,(1-10^{-3})/(S-1)$, $j=0,\dots,S{-}1$, from $t_0{=}1$ to $t_{S-1}{\approx}0.007$, evaluating the velocity at the start point $t_j$ ($j{<}S{-}1$) of each interval without querying $t_{S-1}$, so that $S$ grid points execute $N{=}S{-}1$ neural function evaluations (NFEs) per predicted frame. On the unshifted uniform grid ($\alpha{=}1$, \texttt{time\_dist\_shift=1.0}), four-NFE Euler (\texttt{flow\_steps=4}) steps across $t\in\{1.00, 0.75, 0.50, 0.25\}$ to $t_{\mathrm{end}}{=}0$. Heun and the explicit midpoint rule (\cref{sec:s-controls}) evaluate twice per interval without terminal reuse (at both interval endpoints $(t_j, t_{j+1})$ for Heun and at $(t_j, \tfrac{1}{2}(t_j{+}t_{j+1}))$ for explicit midpoint) on both the shifted $S{=}3$ grid and the unshifted uniform grid ($\alpha{=}1$, \texttt{flow\_steps=2} intervals of size $1/2$ ending at $t_{\mathrm{end}}{=}0$), totaling four NFEs per frame ($3{,}840$ per decision).

\paragraph{Generation Protocol.}
Direct prediction conditions on four context frames and predicts the frame 1, 2, 4, 8, or 16\,s ahead (4 frames per second) in one shot; every test sample of each dataset serves as an evaluation start state. Rollouts predict 64 frames autoregressively by appending each predicted latent to the context window from every test sample across all three datasets ($9{,}561$ anchors total) for \mono, \paper \M{4}, and \paper \M{1}. Metrics are computed on decoded frames downsampled $256{\to}224$ under both non-antialiased bicubic resizing (Table~1, matching direct prediction) and antialiased bicubic resizing (\cref{tab:s-rollout-1024}): \lpips~\cite{zhang2018lpips} (AlexNet), \fid~\cite{heusel2017fid} over all predicted frames of a horizon against the ground-truth frames, and the DINO distance~\cite{zhang2026raenwm}, defined as the cosine distance between L2-normalized DINOv2 patch features of the predicted and ground-truth frames averaged over the 256 patch tokens.

\paragraph{Planning Protocol.}
Planning follows the \cem protocol of the benchmark~\cite{bar2025nwm,zhang2026raenwm}: 500 items per dataset are drawn once at random (seed 0) from the test trajectories and are shared across every planner; each item provides four context frames and a goal frame eight steps (2\,s) ahead. A decision draws 120 candidate action sequences from the \cem proposal (three control points per candidate, interpolated to eight one-frame steps), rolls every candidate out autoregressively for eight frames in normalized DINOv2 token space with independent Gaussian sampler noise at each frame (without RAE pixel decoding), scores the final predicted latent by cosine DINO token distance to the goal latent, and returns the mean action sequence of the three lowest-distance candidates as the plan, which is compared open-loop against the recorded trajectory without execution; a single \cem round is used, and each candidate is rolled out once (\nwm averages three rollouts per candidate). Evaluation seeds re-draw the 120 candidates and the sampler noise. \ate\ and \rpe~\cite{sturm2012benchmark} are computed in the normalized action space of the benchmark, where distances are divided by the average step size of each dataset (step-size units). Decision latency is measured on one NVIDIA A100-SXM4-40GB GPU across PyTorch scaled-dot-product attention (\emph{fused}), \texttt{torch.compile} (\texttt{reduce-overhead}, with both 8-thread pinned CPU staging and CUDA-native \texttt{randn}), and the reference explicit attention kernel on a shared 50-item \sacson shard (compiled latency with 8-thread staging in Table~3; all six execution stacks in \cref{tab:s-sdpa-shard}, \cref{sec:s-controls}), alongside 1{,}500-item pooled explicit-attention means across all three datasets (\cref{tab:s-plan-full}). When shared-cluster pre-emption interrupts an evaluation job, the resumed pass skips completed items deterministically because item random seeds derive solely from $(\text{seed}, \text{dataset}, \text{item index})$.

\paragraph{Statistics.}
All intervals are 95\,\% percentile bootstrap intervals with 1{,}000 resamples (seed 0; 10{,}000 resamples on speed-calibrated \sacson, \cref{tab:s-b2-cal}) over test samples (generation) or planning items, computed prior to rounding. Differences between our own models are paired on shared test samples or items, and \sig\ marks an interval that excludes zero. An evaluation seed re-draws the 120 candidates and the sampler noise of every item; at the same seed two planners receive identical candidates and initial noise, so a paired difference isolates the model or the sampler. Where three seeds exist, we average each item's paired difference over the seeds and bootstrap over items (\cref{tab:s-pooled}); the main text reports these three-seed estimates and notes auxiliary seed-0 controls explicitly. A planner is called non-inferior when the upper bound of the three-seed interval lies below 5\,\% of the three-seed error of the 49-NFE ($S{=}50$) monolithic planner (0.068, 0.142, and 0.052 on \recon, \sacson, and \scand). Primary comparisons test the two headline hypotheses of the paper (\paper \M{1} against \mono at 49 NFEs, $S{=}50$, and four NFEs, $S{=}5$, against 49 NFEs within each model) on \recon and \sacson; a 10{,}000-resample audit on the 12-test primary family (\M{1} minus \mono at 49 and four NFEs, plus four-vs-49 NFEs within \mono, \M{4}, \M{1}, and Standalone \M{1} on \recon and \sacson): 7 of the 8 step-count gains survive item-bootstrap Holm--Bonferroni at $\alpha{=}0.05$ ($p_{\mathrm{Holm}}\le 0.007$, with \M{4} \sacson at $p_{\mathrm{Holm}}{=}0.12$), while 0 of 4 granularity differences reject zero; and under a conservative two-way ($\text{item}{\times}\text{seed}$) bootstrap (3 seeds, inflating SEs by $1.35\text{--}1.43\times$), unadjusted intervals still exclude zero for 7 of 8 step-count gains and for $2{\times}\at{1}{3} - \at{1}{5}$ on \sacson ($-0.122$, CI [$-0.201$, $-0.040$]), with 5 of 12 surviving two-way Holm (\M{1} and \mono on \sacson at $p_{\mathrm{Holm}}{=}0.074$). Exploratory controls report uncorrected item-bootstrap intervals. \cref{tab:s-cluster} repeats single-seed comparisons with whole trajectories as the resampling unit.

\paragraph{Calibration of Single-Seed Tests.}
To measure how far one evaluation seed can be trusted, we form comparisons whose true mean is zero by construction: the same configuration at two seeds, and the difference between two models (or between two schedule lengths) at one seed minus the same difference at another seed. \cref{tab:s-nullcal} counts how many of these 243 comparisons exceed $|z|{=}1.96$, where $z$ is the mean over the 500 items divided by its standard error. Rejections are at the nominal rate when one configuration is evaluated twice (3 of 90), above it for differences between step counts (7 of 81, of which 4 on \scand), and far above it for differences between models on \sacson (9 of 24); the largest null statistic is $z{=}{-}3.6$ (\M{1} against the baseline at four NFEs, $S{=}5$, on \sacson, evaluation seeds 0 and 1). The 243 comparisons are built from three seeds and are not independent, so the counts indicate where single-seed tests fail without estimating an independent rejection rate. We therefore rest claims on three seeds wherever they exist. In the seeding scheme, the random generator of item $i$ at seed $s$ matches that of item $i{+}s$ at seed 0 (overlapping on 4--24\,\% of evaluated item indices across seed pairs); because item $i$ and item $i{+}s$ have distinct scene contexts and goal frames $(\bm{z}_{i-3:i}, \bm{z}_g)$, identical standard-normal draws map to unrelated trajectories while each individual item $i$'s three seeds $(i, i{+}1, i{+}2)$ remain strictly disjoint.

\begin{table}[t]
\centering\scriptsize
\caption{\textbf{Calibration of single-seed tests} on comparisons whose true mean difference is zero. Entries: comparisons with $|z|{>}1.96$ out of all comparisons (nominal rate 5\,\%), and the standard deviation of $z$ (nominal 1).}
\label{tab:s-nullcal}
\setlength{\tabcolsep}{2.5pt}
\resizebox{\linewidth}{!}{%
\begin{tabular}{@{}l cc cc cc cc@{}}
\toprule
 & \multicolumn{2}{c}{\recon} & \multicolumn{2}{c}{\sacson} & \multicolumn{2}{c}{\scand} & \multicolumn{2}{c}{All} \\
\cmidrule(lr){2-3}\cmidrule(lr){4-5}\cmidrule(lr){6-7}\cmidrule(l){8-9}
Null comparison & reject & sd & reject & sd & reject & sd & reject & sd \\
\midrule
\tabinput tab/gen/supp_nullcal_body.tex 
\bottomrule
\end{tabular}}
\end{table}

\section{Complete Generation Results}
\label{sec:s-gen}

\cref{tab:s-gen-full} gives direct prediction at 4\,s for every model and grid size $S$ ($N{=}S{-}1$ NFEs) on the three datasets, \cref{tab:s-gen-horizon} the horizon curves at $S{=}50$ (49 NFEs) and $S{=}5$ (4 NFEs), \cref{tab:widthsteps} the paired comparison of granularity against schedule length at matched computation, and \cref{tab:paired} the paired cost of nesting against the monolithic baseline. Three regularities hold across direct prediction. Realism, measured by \fid, worsens monotonically as granularity shrinks at every schedule and as $S$ shrinks below 20 (19 NFEs) for every granularity; at $S{=}3$ (2 NFEs), 16\,s direct-prediction \fid reaches 38.3 / 43.3 / 52.0 on \sacson, 26.5 / 29.4 / 33.7 on \recon, and 52.6 / 57.1 / 61.3 on \scand for \mono / \paper \M{4} / \paper \M{1}. Distortion (\lpips) improves down to $S\in\{4,5\}$ (3--4 NFEs) and worsens at $S{=}3$ (2 NFEs), while DINO distance keeps improving down to $S{=}3$ on \recon and \scand and reverses at $S{=}3$ only on \sacson because the coarse Euler step pulls predictions toward the conditional mean (the perception--distortion trade-off~\cite{blau2018perception}); and the gap between \M{1} and \M{4} remains a near-constant offset down to $S{=}5$ (4 NFEs) and widens below it. Over 64-frame autoregressive rollouts at $S{=}50$ across all $9{,}561$ test anchors (\cref{tab:s-rollout-1024}), unified non-antialiased $256{\to}224$ bicubic resizing (Table~1) yields frame-16 (4.25\,s) / frame-63 (16\,s) \fid of 17.1 / 19.1 (\mono), 18.0 / 20.5 (\M{4}), and 20.2 / 22.7 (\M{1}) on \sacson; 19.3 / 17.7, 19.3 / 18.4, and 21.4 / 20.9 on \recon; and 27.5 / 30.3, 26.8 / 32.5, and 27.3 / 32.0 on \scand (paired full-split \M{1}$-$\mono \lpips: $+0.019$\sig / $+0.021$\sig on \sacson, $+0.009$\sig / $+0.007$\sig on \recon, and $-0.008$\sig at 4.25\,s flipping to $+0.007$\sig at 16\,s on \scand). Unifying the resize kernel shifts 4\,s \recon \fid by $+2.4\text{--}2.6$ and resolves the 4\,s \recon rollout-vs.-direct inversion ($+0.2$ to $+1.1$ above 4\,s direct \fid under both kernels; $-0.8$ to $+0.6$ at 16\,s), whereas on \scand rollout \fid remains $1.0\text{--}3.8$ below 4\,s direct \fid under both kernels because resize accounts for only $0.3\text{--}0.7$ \fid. On a $1{,}024$-anchor seeded subset per dataset (antialiased), finite-sample \fid bias shifts \sacson 4.25\,s / 16\,s rollout \fid upward by $+12.8\text{--}13.4$ points (29.6 / 31.9, 31.1 / 33.5, and 33.4 / 36.1) while preserving model ordering. Ranked rollout panels are in \cref{fig:s-roll-sacson,fig:s-roll-recon,fig:s-roll-scand}.

\begin{table}[t]
\centering\scriptsize
\caption{\textbf{Full-split 64-frame autoregressive rollout quality} ($S{=}50$, 49 NFEs; frame 16 = 4.25\,s, frame 63 = 16\,s) across all $9{,}561$ test anchors ($4{,}262$ \recon, $3{,}416$ \sacson, $1{,}883$ \scand) scored under both $256{\to}224$ bicubic resize kernels.}
\label{tab:s-rollout-1024}
\setlength{\tabcolsep}{2.3pt}\renewcommand{\arraystretch}{0.95}
\resizebox{\linewidth}{!}{%
\begin{tabular}{@{}l cc cc cc@{}}
\toprule
 & \multicolumn{2}{c}{\mono} & \multicolumn{2}{c}{\paper \M{4}} & \multicolumn{2}{c}{\paper \M{1}} \\
\cmidrule(lr){2-3}\cmidrule(lr){4-5}\cmidrule(l){6-7}
Metric & 4.25\,s & 16\,s & 4.25\,s & 16\,s & 4.25\,s & 16\,s \\
\midrule
\multicolumn{7}{@{}l}{\emph{Full test splits ($9{,}561$ anchors; non-antialiased bicubic resize, matching Table~1)}} \\
\sacson \lpips $\downarrow$ & 0.371 & 0.450 & 0.378 & 0.454 & 0.390 & 0.471 \\
\sacson DINO $\downarrow$ & 0.384 & 0.451 & 0.390 & 0.456 & 0.400 & 0.471 \\
\sacson \fid $\downarrow$ & 17.1 & 19.1 & 18.0 & 20.5 & 20.2 & 22.7 \\
\recon \lpips $\downarrow$ & 0.445 & 0.542 & 0.449 & 0.548 & 0.454 & 0.548 \\
\recon DINO $\downarrow$ & 0.468 & 0.550 & 0.472 & 0.556 & 0.474 & 0.553 \\
\recon \fid $\downarrow$ & 19.3 & 17.7 & 19.3 & 18.4 & 21.4 & 20.9 \\
\scand \lpips $\downarrow$ & 0.448 & 0.518 & 0.438 & 0.524 & 0.441 & 0.525 \\
\scand DINO $\downarrow$ & 0.495 & 0.562 & 0.486 & 0.571 & 0.489 & 0.573 \\
\scand \fid $\downarrow$ & 27.5 & 30.3 & 26.8 & 32.5 & 27.3 & 32.0 \\
\midrule
\multicolumn{7}{@{}l}{\emph{Full test splits ($9{,}561$ anchors; antialiased bicubic resize, upstream rollout kernel)}} \\
\sacson \lpips $\downarrow$ & 0.372 & 0.451 & 0.379 & 0.456 & 0.392 & 0.472 \\
\sacson DINO $\downarrow$ & 0.384 & 0.451 & 0.390 & 0.456 & 0.400 & 0.470 \\
\sacson \fid $\downarrow$ & 16.8 & 18.8 & 17.7 & 20.2 & 20.1 & 22.7 \\
\recon \lpips $\downarrow$ & 0.432 & 0.530 & 0.436 & 0.536 & 0.440 & 0.536 \\
\recon DINO $\downarrow$ & 0.468 & 0.550 & 0.472 & 0.556 & 0.474 & 0.553 \\
\recon \fid $\downarrow$ & 16.9 & 15.5 & 16.8 & 16.3 & 18.9 & 18.7 \\
\scand \lpips $\downarrow$ & 0.443 & 0.513 & 0.433 & 0.519 & 0.436 & 0.520 \\
\scand DINO $\downarrow$ & 0.495 & 0.562 & 0.486 & 0.571 & 0.489 & 0.573 \\
\scand \fid $\downarrow$ & 26.9 & 29.5 & 26.1 & 31.6 & 27.0 & 31.4 \\
\bottomrule
\end{tabular}}
\end{table}

\begin{table*}[t]
\centering\scriptsize
\caption{\textbf{Direct prediction at 4\,s for every model and grid size $S$ ($N{=}S{-}1$ NFEs).} \lpips, DINO distance, and \fid on the full test split of each dataset (4{,}262 / 3{,}416 / 1{,}883 samples); the naive slice applies the \M{1} granularity to the monolithic model without nested training.}
\label{tab:s-gen-full}
\setlength{\tabcolsep}{3.5pt}
\begin{tabular}{@{}lc ccc ccc ccc@{}}
\toprule
 & & \multicolumn{3}{c}{\recon} & \multicolumn{3}{c}{\sacson} & \multicolumn{3}{c}{\scand} \\
\cmidrule(lr){3-5}\cmidrule(lr){6-8}\cmidrule(lr){9-11}
Model & $S$ & \lpips & DINO & \fid & \lpips & DINO & \fid & \lpips & DINO & \fid \\
\midrule
\tabinput tab/gen/supp_gen_full_body.tex 
\bottomrule
\end{tabular}
\end{table*}

\begin{table*}[t]
\centering\scriptsize
\caption{\textbf{Direct prediction along the horizon.} \lpips and \fid at 1, 2, 4, 8, and 16\,s for each model at $S{=}50$ (49 NFEs) and $S{=}5$ (4 NFEs).}
\label{tab:s-gen-horizon}
\setlength{\tabcolsep}{3.5pt}
\begin{tabular}{@{}llc ccccc ccccc@{}}
\toprule
 & & & \multicolumn{5}{c}{\lpips} & \multicolumn{5}{c}{\fid} \\
\cmidrule(lr){4-8}\cmidrule(lr){9-13}
Dataset & Model & $S$ & 1\,s & 2\,s & 4\,s & 8\,s & 16\,s & 1\,s & 2\,s & 4\,s & 8\,s & 16\,s \\
\midrule
\tabinput tab/gen/supp_gen_horizon_body.tex 
\bottomrule
\end{tabular}
\end{table*}

\begin{table}[t]
\centering
\scriptsize
\caption{\textbf{Granularity versus schedule length at matched computation.} Paired \lpips difference at 4\,s between the smallest granularity on a denser grid $S$ (A) and full granularity on a coarser grid (B, $N{=}S{-}1$ NFEs per frame), together with relative compute and 4\,s \sacson \fid; positive values favor B, and \sig\ marks a 95\,\% interval that excludes zero.}
\label{tab:widthsteps}
\setlength{\tabcolsep}{3.5pt}
\resizebox{\linewidth}{!}{%
\begin{tabular}{@{}lcccccc@{}}
\toprule
 & \multicolumn{2}{c}{MACs ($\times$)} & \multicolumn{3}{c}{$\Delta$\lpips at 4\,s (A$-$B)} & \fid 4\,s \\
\cmidrule(lr){2-3}\cmidrule(lr){4-6}
A vs.\ B & A & B & \recon & \sacson & \scand & A / B \\
\midrule
\at{1}{50} vs.\ \at{4}{20} & 0.65 & 0.39 & $+$0.007\sig & $+$0.014\sig & $+$0.003\sig & 19.1 / 17.3 \\
\at{1}{20} vs.\ \at{4}{10} & 0.25 & 0.18 & $+$0.007\sig & $+$0.014\sig & $+$0.003\sig & 19.1 / 18.2 \\
\at{1}{10} vs.\ \at{4}{5}  & 0.12 & 0.08 & $+$0.007\sig & $+$0.015\sig & $+$0.004\sig & 20.1 / 21.6 \\
\at{1}{5}  vs.\ \at{4}{3}  & 0.05 & 0.04 & $\phantom{+}$0.000      & $+$0.002\sig & $-$0.009\sig & \bestval{24.7} / 37.1 \\
\midrule
\at{1}{5}  vs.\ \at{4}{5}  & 0.05 & 0.08 & $+$0.007\sig & $+$0.011\sig & $+$0.001\sig & 24.7 / 21.6 \\
\at{1}{5}  vs.\ {\mono}@50 & 0.05 & 1.00 & $+$0.008\sig & $+$0.008\sig & $-$0.016\sig & 24.7 / 16.5 \\
\bottomrule
\end{tabular}}
\end{table}

\begin{table}[t]
\centering
\scriptsize
\caption{\textbf{Cost of nesting, measured against the monolithic baseline.} Mean per-sample (direct prediction at 4\,s and 64-frame rollout at frame 16 = 4.25\,s, \lpips across all $9{,}561$ test anchors under unified non-antialiased bicubic resizing) or per-item (planning, \ate in normalized action space, 500 items, seed 0) difference between a nested sub-network and the monolithic baseline at $S{=}50$ (49 NFEs), with 95\,\% bootstrap intervals for planning; \sig\ marks an interval that excludes zero. Generation pays a small, consistent price for nesting and a second one for granularity on \recon and \sacson, and gains on \scand; planning shows no detectable effect of either.}
\label{tab:paired}
\setlength{\tabcolsep}{2.5pt}\providecommand{\ci}[1]{{\tiny[#1]}}
\resizebox{\linewidth}{!}{%
\begin{tabular}{@{}llcc@{}}
\toprule
Use, metric & Dataset & \M{4} $-$ \mono & \M{1} $-$ \mono \\
\midrule
\multirow{3}{*}{Direct, \lpips 4\,s} & \recon  & $+$0.006\sig & $+$0.011\sig \\
                                                & \sacson & $+$0.010\sig & $+$0.020\sig \\
                                                & \scand  & $-$0.006\sig & $-$0.006\sig \\
\midrule
\multirow{3}{*}{Rollout, \lpips 4.25\,s} & \recon  & $+$0.005\sig & $+$0.009\sig \\
                                      & \sacson & $+$0.007\sig & $+$0.019\sig \\
                                      & \scand  & $-$0.010\sig & $-$0.008\sig \\
\midrule
\multirow{3}{*}{Planning, \ate} & \recon  & $+$0.025 \ci{$-$0.018, $+$0.065} & $+$0.020 \ci{$-$0.022, $+$0.060} \\
                                    & \sacson & $-$0.042 \ci{$-$0.135, $+$0.046} & $-$0.070 \ci{$-$0.159, $+$0.024} \\
                                    & \scand  & $+$0.012 \ci{$-$0.025, $+$0.053} & $+$0.002 \ci{$-$0.039, $+$0.046} \\
\bottomrule
\end{tabular}}
\end{table}

\paragraph{Generation Under the Sampler Controls.}
\cref{tab:s-gen-controls} reports direct prediction under the same sampler controls that \cref{sec:s-controls} applies to planning: one Euler evaluation ($S{=}2$, 1 NFE), Heun and the midpoint rule at $S{=}3$ (4 NFEs, matching Euler at $S{=}5$), Heun at $S{=}2$ (2 NFEs), and the unshifted uniform time grid ($\alpha{=}1$, 4 and 2 Euler NFEs). Across all nine model--dataset columns, one Euler evaluation ($S{=}2$) collapses realism (\fid 112--173, \lpips 0.41--0.55). At four NFEs, Heun improves over four-NFE Euler ($S{=}5$) in \lpips and \fid on \recon and \sacson (\sacson \M{4}: 0.318 / 19.5 vs.\ 0.320 / 21.6), while the midpoint rule recovers most of the 49-NFE \fid across all three datasets (\sacson 19.3 / 17.7 / 17.0 vs.\ 19.1 / 17.1 / 16.5 for \M{1} / \M{4} / \mono).

\begin{table*}[t]
\centering\scriptsize
\caption{\textbf{Direct prediction at 4\,s under the sampler controls} across \recon, \sacson, and \scand (L: \lpips, D: DINO distance, F: \fid; \emph{Evals}: NFEs per predicted frame).}
\label{tab:s-gen-controls}
\setlength{\tabcolsep}{5.0pt}
\begin{tabular}{@{}lc ccc ccc ccc@{}}
\toprule
 & & \multicolumn{3}{c}{\paper \M{1}} & \multicolumn{3}{c}{\paper \M{4}} & \multicolumn{3}{c}{\mono} \\
\cmidrule(lr){3-5}\cmidrule(lr){6-8}\cmidrule(l){9-11}
Sampler & Evals & L & D & F & L & D & F & L & D & F \\
\midrule
\multicolumn{11}{@{}l}{\emph{\recon (4\,s)}} \\
Euler, $S{=}50$ (protocol) & 49 & 0.385 & 0.422 & 20.3 & 0.379 & 0.418 & 19.1 & 0.374 & 0.413 & 18.9 \\
Euler, $S{=}5$ & 4 & 0.382 & 0.401 & 25.0 & 0.375 & 0.397 & 22.9 & 0.369 & 0.392 & 21.7 \\
Euler, $S{=}4$ & 3 & 0.383 & 0.396 & 26.4 & 0.376 & 0.392 & 24.2 & 0.370 & 0.387 & 22.7 \\
Euler, $S{=}3$ & 2 & 0.389 & 0.393 & 31.0 & 0.382 & 0.387 & 27.6 & 0.376 & 0.383 & 25.1 \\
Heun, $S{=}3$ & 4 & 0.377 & 0.404 & 23.6 & 0.371 & 0.399 & 22.0 & 0.365 & 0.393 & 21.7 \\
Midpoint, $S{=}3$ & 4 & 0.376 & 0.407 & 21.9 & 0.370 & 0.402 & 20.7 & 0.364 & 0.398 & 19.4 \\
Euler, $S{=}2$ (one evaluation) & 1 & 0.510 & 0.469 & 172.5 & 0.495 & 0.457 & 154.1 & 0.485 & 0.448 & 140.5 \\
Heun, $S{=}2$ & 2 & -- & -- & -- & 0.393 & 0.431 & 43.7 & 0.386 & 0.415 & 33.5 \\
Euler, uniform grid ($\alpha{=}1$) & 4 & -- & -- & -- & 0.377 & 0.388 & 24.7 & -- & -- & -- \\
Euler, uniform grid ($\alpha{=}1$) & 2 & -- & -- & -- & 0.383 & 0.388 & 36.1 & 0.377 & 0.384 & 32.8 \\
\midrule
\multicolumn{11}{@{}l}{\emph{\sacson (4\,s)}} \\
Euler, $S{=}50$ (protocol) & 49 & 0.343 & 0.362 & 19.1 & 0.333 & 0.354 & 17.1 & 0.323 & 0.345 & 16.5 \\
Euler, $S{=}5$ & 4 & 0.331 & 0.344 & 24.7 & 0.320 & 0.335 & 21.6 & 0.310 & 0.326 & 20.5 \\
Euler, $S{=}4$ & 3 & 0.333 & 0.342 & 29.5 & 0.320 & 0.332 & 25.3 & 0.310 & 0.323 & 23.3 \\
Euler, $S{=}3$ & 2 & 0.344 & 0.346 & 44.7 & 0.329 & 0.334 & 37.1 & 0.319 & 0.323 & 32.7 \\
Heun, $S{=}3$ & 4 & 0.330 & 0.345 & 22.5 & 0.318 & 0.336 & 19.5 & 0.307 & 0.326 & 17.9 \\
Midpoint, $S{=}3$ & 4 & 0.332 & 0.348 & 19.3 & 0.320 & 0.340 & 17.7 & 0.311 & 0.331 & 17.0 \\
Euler, $S{=}2$ (one evaluation) & 1 & 0.442 & 0.438 & 142.0 & 0.422 & 0.416 & 124.8 & 0.408 & 0.400 & 112.4 \\
Heun, $S{=}2$ & 2 & -- & -- & -- & 0.353 & 0.367 & 47.6 & 0.341 & 0.351 & 41.3 \\
Euler, uniform grid ($\alpha{=}1$) & 4 & -- & -- & -- & 0.319 & 0.334 & 26.6 & 0.312 & 0.324 & 24.5 \\
Euler, uniform grid ($\alpha{=}1$) & 2 & -- & -- & -- & 0.354 & 0.349 & 57.9 & 0.345 & 0.338 & 52.4 \\
\midrule
\multicolumn{11}{@{}l}{\emph{\scand (4\,s)}} \\
Euler, $S{=}50$ (protocol) & 49 & 0.399 & 0.456 & 31.1 & 0.399 & 0.455 & 29.1 & 0.405 & 0.458 & 28.5 \\
Euler, $S{=}5$ & 4 & 0.389 & 0.440 & 36.4 & 0.389 & 0.438 & 34.1 & 0.397 & 0.444 & 33.3 \\
Euler, $S{=}4$ & 3 & 0.390 & 0.436 & 40.5 & 0.388 & 0.434 & 37.8 & 0.397 & 0.441 & 36.8 \\
Euler, $S{=}3$ & 2 & 0.401 & 0.433 & 55.1 & 0.398 & 0.431 & 52.1 & 0.406 & 0.439 & 49.9 \\
Heun, $S{=}3$ & 4 & 0.399 & 0.445 & 39.4 & 0.397 & 0.443 & 37.7 & 0.406 & 0.448 & 36.4 \\
Midpoint, $S{=}3$ & 4 & 0.393 & 0.447 & 32.0 & 0.392 & 0.444 & 30.4 & 0.401 & 0.450 & 29.8 \\
Euler, $S{=}2$ (one evaluation) & 1 & 0.547 & 0.476 & 169.7 & 0.537 & 0.471 & 163.6 & 0.526 & 0.473 & 146.4 \\
Heun, $S{=}2$ & 2 & -- & -- & -- & 0.420 & 0.457 & 64.0 & 0.423 & 0.458 & 57.2 \\
Euler, uniform grid ($\alpha{=}1$) & 4 & -- & -- & -- & 0.390 & 0.431 & 39.8 & 0.399 & 0.439 & 37.2 \\
Euler, uniform grid ($\alpha{=}1$) & 2 & -- & -- & -- & 0.422 & 0.427 & 69.1 & 0.423 & 0.436 & 63.3 \\
\bottomrule
\end{tabular}
\end{table*}

\section{Complete Planning Results}
\label{sec:s-plan}

\cref{tab:s-plan-full} lists \ate, \rpe, and the 1{,}500-item pooled explicit-attention wall-clock time per decision of every planner and grid size $S$ ($N{=}S{-}1$ NFEs) at seed 0, \cref{tab:s-seeds} the three evaluation seeds (including intermediate grids $S\in\{20,10,4\}$ and uniform Euler on \M{1}/\mono, \paper \M{4} across $S\in\{50,5,3,2\}$, standalone \M{1} at $S\in\{50,5\}$, standalone \M{0} across $S\in\{50,5,3,2\}$, uniform-grid Heun/midpoint, and multi-draw score averaging across \M{1}, \M{4}, and \mono), \cref{tab:s-pooled,tab:s-e12} the three-seed paired differences behind the statements of the main text, \cref{tab:s-cluster} single-seed comparisons under item and trajectory resampling, \cref{tab:s-calib} the model-free task calibration, \cref{tab:s-b2-cal} the 24-cell leakage-free speed-calibrated \sacson benchmark, \cref{tab:s-turns} the full 12-row curvature stratification behind Table~5, and \cref{tab:s-horizon-cem} the two-round iterative \cem and 4\,s (16-step) horizon stress tests. \rpe\ varies by less than 0.02 across our flow-matching planners at any 2\,s protocol schedule and follows \ate\ everywhere, so the main text reports \ate\ primarily and notes calibrated \sacson \rpe\ explicitly.

\begin{table*}[t]
\centering\scriptsize
\caption{\textbf{Planning accuracy for every planner and grid size $S$ ($N{=}S{-}1$ NFEs)} at evaluation seed 0 (the same 500 items per dataset). \ate\ and translational \rpe\ in normalized action space; \emph{s/dec.}: pooled mean wall-clock time per decision across all 1{,}500 items (A100, explicit attention kernel; shared-shard fused SDPA and explicit medians in \cref{tab:s-sdpa-shard}).}
\label{tab:s-plan-full}
\setlength{\tabcolsep}{4pt}
\begin{tabular}{@{}lc cc cc cc c@{}}
\toprule
 & & \multicolumn{2}{c}{\recon} & \multicolumn{2}{c}{\sacson} & \multicolumn{2}{c}{\scand} & \\
\cmidrule(lr){3-4}\cmidrule(lr){5-6}\cmidrule(lr){7-8}
Planner & $S$ & \ate & \rpe & \ate & \rpe & \ate & \rpe & s/dec. \\
\midrule
\tabinput tab/gen/supp_plan_full_body.tex 
\bottomrule
\end{tabular}
\end{table*}

\begin{table*}[t]
\centering\scriptsize
\caption{\textbf{Evaluation seeds across grid sizes $S$ ($N{=}S{-}1$ NFEs) and solvers.} \ate\ at seeds 0 / 1 / 2 (each seed re-draws the 120 candidates of every item and the sampler noise) with mean and standard deviation over the three seeds (seeds 1--2 of $S\in\{20,10,4\}$ and unif.\ Euler use fused SDPA, \cref{tab:s-e12}; \emph{3, 2 draws} and \emph{2, 4 draws} average two $S{=}3$ or four $S{=}2$ noise draws offline on all $500$ items per dataset, \cref{sec:s-noise}).}
\label{tab:s-seeds}
\setlength{\tabcolsep}{4pt}
\begin{tabular}{@{}lc lll@{}}
\toprule
Planner & $S$ & \recon & \sacson & \scand \\
\midrule
\tabinput tab/gen/supp_seeds_body.tex 
\bottomrule
\end{tabular}
\end{table*}

\begin{table*}[t]
\centering\scriptsize
\caption{\textbf{Three-seed paired differences in \ate} (first minus second configuration; negative favors the first, $500$ items per dataset). For each dataset: the paired difference at seeds 0 / 1 / 2 (\sig: single-seed interval excludes zero), then the difference averaged over the three seeds with its 95\,\% bootstrap interval over items (\sig: excludes zero). Last row: the 5\,\% non-inferiority margin.}
\label{tab:s-pooled}
\setlength{\tabcolsep}{3pt}
\resizebox{\textwidth}{!}{%
\begin{tabular}{@{}l ll ll ll@{}}
\toprule
 & \multicolumn{2}{c}{\recon} & \multicolumn{2}{c}{\sacson} & \multicolumn{2}{c}{\scand} \\
\cmidrule(lr){2-3}\cmidrule(lr){4-5}\cmidrule(l){6-7}
Comparison & seeds 0 / 1 / 2 & three seeds & seeds 0 / 1 / 2 & three seeds & seeds 0 / 1 / 2 & three seeds \\
\midrule
\tabinput tab/gen/supp_pooled_body.tex 
\bottomrule
\end{tabular}}
\end{table*}

\begin{table*}[t]
\centering\scriptsize
\caption{\textbf{Paired \ate\ differences for intermediate Euler grids ($S\in\{20,10,4\}$) and three-seed uniform Euler ($\alpha{=}1$, 4 NFEs)} ($500$ items per dataset, 95\,\% item-bootstrap intervals; negative favors the first configuration). Seed 0 compares identical eager/explicit-attention runs on both sides; seeds 1--2 of $S\in\{20,10,4\}$ and uniform Euler use fused SDPA against eager-attention reference runs ($S\in\{50,5,3\}$ and uniform Heun/midpoint). On 500 \sacson items at $S{=}5$ (seed 0), the paired fused-minus-eager difference is $+0.011$ [$-0.002$, $+0.025$] for \M{1} and $-0.001$ [$-0.012$, $+0.011$] for \mono (neither significant; 479/500 top-3 plans identical in both arms).}
\label{tab:s-e12}
\setlength{\tabcolsep}{2.8pt}
\resizebox{\textwidth}{!}{%
\begin{tabular}{@{}l ll ll ll@{}}
\toprule
 & \multicolumn{2}{c}{\recon} & \multicolumn{2}{c}{\sacson} & \multicolumn{2}{c}{\scand} \\
\cmidrule(lr){2-3}\cmidrule(lr){4-5}\cmidrule(l){6-7}
Comparison & seed 0 (same kernel) & three seeds (seeds 1--2 fused) & seed 0 (same kernel) & three seeds (seeds 1--2 fused) & seed 0 (same kernel) & three seeds (seeds 1--2 fused) \\
\midrule
\tabinput tab/gen/supp_e12_body.tex
\bottomrule
\end{tabular}}
\end{table*}

\begin{table*}[t]
\centering\scriptsize
\caption{\textbf{Item against trajectory resampling.} Paired \ate\ differences (seed 0) across grid sizes $S$ with 95\,\% percentile intervals from 2{,}000 resamples of items and of whole trajectories (clusters of items from the same test trajectory; 391, 202, and 43 trajectories on \recon, \sacson, and \scand). No decision about the exclusion of zero changes.}
\label{tab:s-cluster}
\setlength{\tabcolsep}{5pt}
\begin{tabular}{@{}llc r ll c@{}}
\toprule
Comparison & $S$ & Dataset & Mean $\Delta$ & Item CI & Trajectory CI & Trajectories \\
\midrule
\tabinput tab/gen/supp_cluster_body.tex 
\bottomrule
\end{tabular}
\end{table*}

\begin{table}[t]
\centering\scriptsize
\caption{\textbf{Calibration of the planning task.} \ate\ of plans that use no world model on the same 500 items, against seed-0 world-model planners. The prior mean is the straight line at the nominal speed of the \cem proposal; the fitted straight line sets mean step progress $\bar{v}\in\{0.920, 0.896, 0.961\}$ strictly on training splits ($52{,}741 / 22{,}556 / 8{,}110$ windows, zero trajectory overlap; in-sample test fit $1.588 / 2.242 / 0.820$ is nearly identical); random shooting executes the mean of three random candidates (20 repetitions); the oracles select from a fresh draw of 120 candidates the candidate(s) closest to ground truth.}
\label{tab:s-calib}
\setlength{\tabcolsep}{4pt}
\resizebox{\linewidth}{!}{%
\begin{tabular}{@{}lccc@{}}
\toprule
Plan & \recon & \sacson & \scand \\
\midrule
\tabinput tab/gen/supp_calibration_body.tex 
\bottomrule
\end{tabular}}
\end{table}

\paragraph{Task Calibration, Train-Calibrated \sacson Prior, and Two-Stage Re-Scoring.}
In \cref{tab:s-calib}, all world-model planners outperform both the nominal \cem prior (1.637) and the leakage-free train-split constant-speed straight line (1.589; $2{\times}\at{1}{3}$ improves by $-0.349$\sig\ overall and $-1.031$\sig\ on Q4 sharp turns while trailing by $+0.242$\sig\ on straightest Q1) on \recon. On \sacson, the nominal \cem proposal substantially undershoots indoor pedestrian speed (0.32 vs.\ 0.92 step units, incurring 3.780 \ate\ vs.\ 2.241 for the train-split constant-speed straight line and 2.67 even on straightest Q1, \cref{tab:s-turns}). Selecting the top 3 among 120 laterally dispersed \cem candidates adds lateral variance on straight corridors (Q1--Q3), whereas on sharp turns (Q4) the slow proposal fan caps world-model forward progress at 0.60--0.65 so forward-speed RMSE (2.51 vs.\ 1.99) masks $2{\times}\at{1}{3}$'s lateral turn improvement (2.46 vs.\ 3.07 lateral RMSE), tying overall Q4 \ate\ at 3.74 vs.\ 3.82 ($-0.072$, CI [$-0.296$, $+0.160$]). To remove this proposal bottleneck without test-set leakage (\cref{tab:s-b2-cal}), we fit one scalar on the $22{,}556$ \sacson training windows ($\mu_x{=}0.014$, corresponding to $v_0{=}0.896$ step units, frozen before any test evaluation) and evaluate 24 world-model cells across all 500 items and three seeds (multi-draw cells executed online at batch 240 and 480). Calibration lowers \ate\ by $0.50\text{--}0.68$ across all 20 paired protocol counterparts ($p{=}0.0002$): 13 of 24 cells beat the $2.241$ straight line on item-bootstrap intervals, 9 on two-way ($\text{item}{\times}\text{seed}$) intervals, and 7 after 24-test Holm--Bonferroni correction, led by $2{\times}@3$ ($2.057 / 2.079 / 2.092$ on \mono / \M{4} / \M{1}; $-0.184$\sig / $-0.162$\sig / $-0.149$\sig\ vs.\ line) and $@5$ ($2.091 / 2.100 / 2.130$). Conversely, 49-NFE \at{4}{50} ($2.209$) and \at{1}{50} ($2.229$) tie the line, and \monoat{50} ($2.154$, raw $p{=}0.048$) includes zero under both two-way resampling and Holm correction. Every cell beats the line on item intervals in Q4 ($-0.43$\sig\ to $-0.80$\sig\ \ate; Q4 \rpe\ $-0.07$\sig\ to $-0.14$\sig) and 19 of 24 on Q3. In contrast, every cell trails the zero-jitter line on straight Q1 ($+0.28$\sig\ to $+0.59$\sig) and overall step-wise \rpe\ ($0.600\text{--}0.634$ vs.\ $0.585$; 16 of 24 significant). Within calibrated \sacson, $@5$ and $2{\times}@3$ significantly beat $@50$ on item intervals in all three arms ($-0.063$\sig\ to $-0.137$\sig), while $2{\times}@3{-}@5$ stays negative ($-0.021$ to $-0.038$) without reaching significance ($p\in[0.09,0.39]$). $2{\times}@3$ beats $4{\times}@2$ at equal $3{,}840$ evaluations on item intervals across all three arms ($-0.062$\sig\ to $-0.161$\sig), and Euler beats Heun/midpoint in direction across all 8 pre-registered 4-NFE contrasts (5 significant on item CIs, 3 two-way). Across granularities on item intervals, \M{1} stays within $+0.035\text{--}0.039$ (n.s.) of \mono at $2{\times}@3$ and $@5$, trails at $@50$ ($+0.075$\sig), $@3$ ($+0.058$\sig), and $@2$ ($+0.067$\sig), and beats standalone \M{0}@5 by $-0.115$\sig\ ($p{=}0.0014$). On \scand, sidewalk trajectories stay near the straight-line prior (0.834) while dynamic pedestrians corrupt terminal patch-wise DINO distance. At seed 0, screening the 120 candidates with \M{1} and re-scoring the best $k{=}24$ with \M{4} at $S{=}50$ gives 1.37, 2.80, and 1.03 \ate, within the paired interval of single-stage \paper \M{1} (1.37, 2.78, 1.01) at 1.24$\times$ the wall-clock cost.

\begin{table*}[t]
\centering\scriptsize
\caption{\textbf{Speed-calibrated \sacson planning benchmark} ($\mu_x{=}0.014$, $v_0{=}0.896$ step units fitted strictly on training windows; 500 test items $\times$ 3 seeds, fused SDPA; multi-draw rows executed online at batch 240 and 480). Columns report three-seed mean\,$\pm$\,sd of \ate\ and translational \rpe\ (\sig\ on \rpe: paired 95\,\% item CI vs.\ straight line excludes zero), paired $\Delta\text{\ate}$ against the $2.241$ train-fit straight line (\sig: item CI excludes zero; $^{\ddagger}$: survives 24-test Holm--Bonferroni at $\alpha{=}0.05$), curvature-quartile \ate\ (Q1--Q4, 125 items each), and overall forward / lateral trajectory RMSE ($e_x / e_y$).}
\label{tab:s-b2-cal}
\setlength{\tabcolsep}{3.0pt}\renewcommand{\arraystretch}{0.95}
\resizebox{\textwidth}{!}{%
\begin{tabular}{@{}lc cc l cccc c@{}}
\toprule
Planner & NFEs & \ate\ (mean\,$\pm$\,sd) & \rpe\ (mean\,$\pm$\,sd) & Paired $\Delta\text{\ate}$ vs.\ line [95\,\% CI] & Q1 & Q2 & Q3 & Q4 & $e_x$ / $e_y$ \\
\midrule
\tabinput tab/gen/supp_b2_cal_body.tex
\bottomrule
\end{tabular}}
\end{table*}

\begin{figure*}[tp]
\centering
\includegraphics[width=0.93\textwidth]{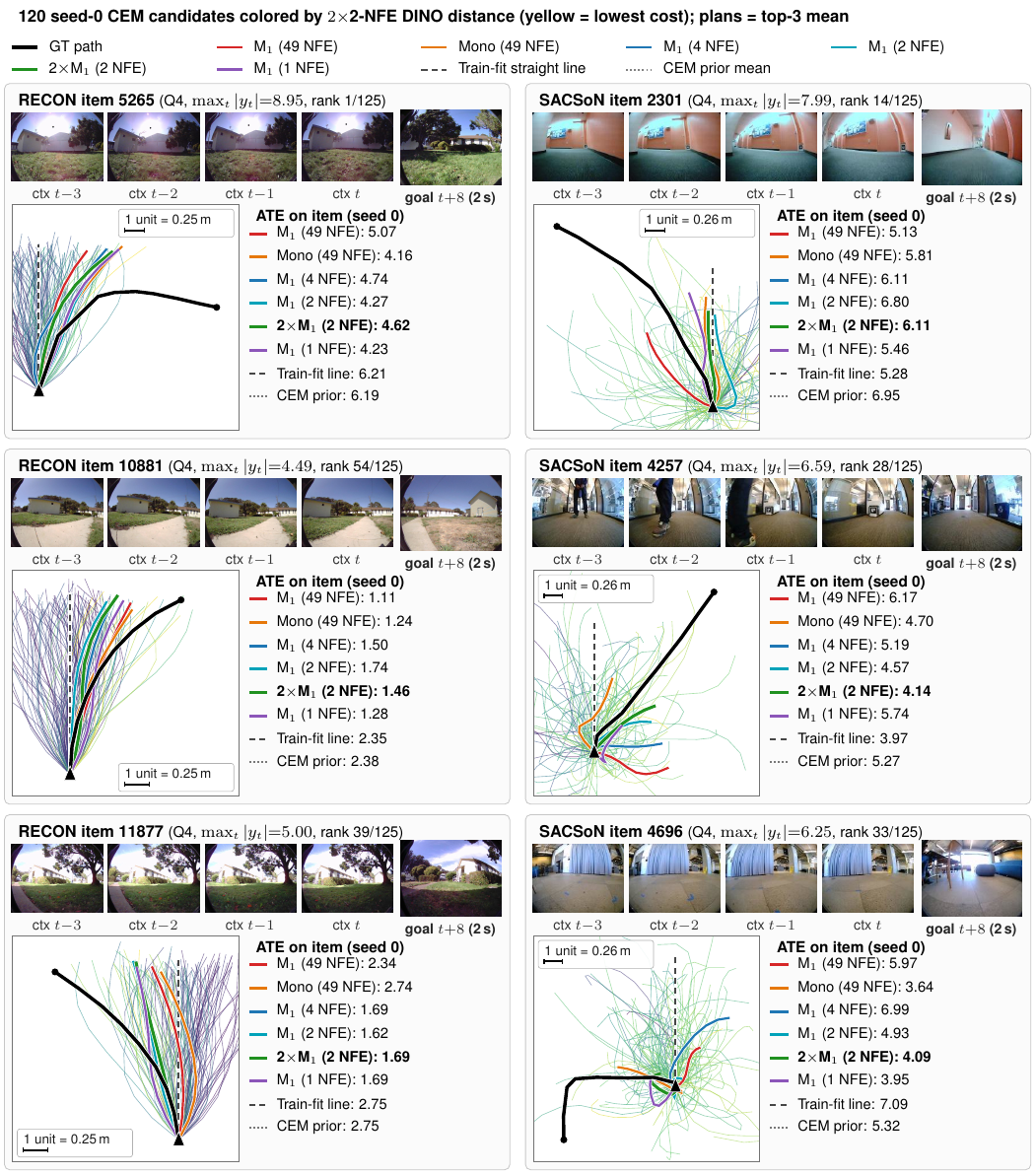}
\caption{\textbf{Qualitative open-loop \cem proposals on sharp turns (Q4).} Extends Fig.~3 of the main paper to six randomly sampled items from the sharpest quartile (seed 0; left column: three \recon items, right column: three \sacson items), including its two panels: \recon item 11877 and \sacson item 4257 (6.17 $\to$ 4.14 \ate\ from 49-NFE \M{1} to $2{\times}\at{1}{3}$). Candidate arcs are colored by terminal DINOv2 patch distance under the two-draw $S{=}3$ planner ($2{\times}\at{1}{3}$); black curves mark recorded ground-truth paths and dashed gray lines show the train-calibrated constant-speed straight line. On open terrain (item 11877), $2{\times}\at{1}{3}$ commits to the sharp left turn, beating both 49-NFE rollouts and the straight line (1.69 vs.\ 2.34 and 2.75 \ate), whereas extreme maneuvers outside the 120-sample proposal fan bound all selectors (item 5265). On indoor \sacson, the benchmark's slow \cem speed prior (0.32 vs.\ 0.92 recorded step units) restricts forward reach: even when $2{\times}\at{1}{3}$ selects the true turn direction and beats both 49-NFE \M{1} and the straight line (item 4696: 4.09 vs.\ 5.97 and 7.09 \ate), the short proposal fan leaves large residual forward error.}
\label{fig:s-cem-qual}
\end{figure*}

\paragraph{Results by Item Difficulty and Trajectory Curvature.}
\cref{tab:s-difficulty} splits the 500 items of each dataset at the median error of the nominal straight-line \cem prior, measuring how far the recorded path departs from the prior in speed or direction, and \cref{tab:s-turns} stratifies all 500 items per dataset into quartiles of ground-truth lateral excursion $\max_{t\le 8}|y_t|$ (125 items per quartile, three-seed mean). On \recon and \sacson the few-NFE gain is present in both prior-error halves and concentrates on sharp maneuvers: on \recon Q4 ($\max|y|\in[3.54,8.95]$), \paper \M{1} improves from 1.95 ($S{=}50$) to 1.70 ($S{=}5$, $-0.249$\sig), 1.67 ($S{=}3$), and 1.64 ($S{=}2$, $-0.314$\sig), accounting for 62\,\% of the overall gain and exceeding Q1--Q3 by $-0.197$\sig\ [$-0.263$, $-0.134$]. On \sacson, four-NFE and two-NFE Euler match or improve upon 49 NFEs across all four quartiles, whereas single-NFE \at{1}{2} degrades significantly against \at{1}{5} only on the sharpest quartile Q4 ($+0.160$\sig\ [$+0.008$, $+0.319$]), isolating where single-step conditional expectation smooths sharp indoor turns. On \scand, the planners trail the nominal straight-line prior across all four quartiles and both prior-error halves. \cref{fig:s-cem-recon,fig:s-cem-sacson,fig:s-cem-counter} add ranked Q4 items to the random draw of \cref{fig:s-cem-qual}.

\begin{table*}[t]
\centering\scriptsize
\caption{\textbf{Planning by item difficulty.} Items split at the dataset median of the straight-line prior error (250 items per half). \ate\ of the prior mean and three-seed \ate\ of three planners ($S{=}50$: 49 NFEs; $S{=}5$: 4 NFEs), followed by three-seed paired differences with 95\,\% intervals (\sig: excludes zero).}
\label{tab:s-difficulty}
\setlength{\tabcolsep}{4pt}
\resizebox{\textwidth}{!}{%
\begin{tabular}{@{}ll cccc lll@{}}
\toprule
Dataset & Half & Prior mean & \monoat{50} & \monoat{5} & \at{1}{5} & \at{1}{5} $-$ \at{1}{50} & \monoat{5} $-$ \monoat{50} & \at{1}{5} $-$ \monoat{50} \\
\midrule
\tabinput tab/gen/supp_difficulty_body.tex 
\bottomrule
\end{tabular}}
\end{table*}

\begin{table*}[t]
\centering\scriptsize
\caption{\textbf{Planning accuracy (\ate) across all trajectory curvature quartiles} (125 items per quartile, three-seed mean; extends Table~5). Quartiles partition ground-truth lateral excursion $\max_{t\le 8}|y_t|$ in dataset step-size units ($2{\times}@3$: two-draw \at{1}{3}); paired differences include 95\,\% bootstrap intervals (\sig: excludes zero).}
\label{tab:s-turns}
\setlength{\tabcolsep}{1.8pt}
\resizebox{\textwidth}{!}{%
\begin{tabular}{@{}ll cc cc ccccc cccccc@{}}
\toprule
 & & & & \multicolumn{2}{c}{\mono} & \multicolumn{5}{c}{\paper \M{1}} & \multicolumn{6}{c}{Three-seed paired $\Delta$ [95\,\% CI]} \\
\cmidrule(lr){5-6}\cmidrule(lr){7-11}\cmidrule(l){12-17}
Dataset & Quartile ($\max|y|$) & Prior & Line & $@50$ & $@5$ & $@50$ & $@5$ & $@3$ & $2{\times}@3$ & $@2$ & \at{1}{5}$-$\at{1}{50} & \at{1}{3}$-$\at{1}{50} & \at{1}{2}$-$\at{1}{5} & \at{1}{2}$-$\at{1}{50} & $2{\times}\at{1}{3}{-}\at{1}{2}$ & \monoat{5}$-$\monoat{50} \\
\midrule
\tabinput tab/gen/supp_turns_body.tex 
\bottomrule
\end{tabular}}
\end{table*}

\paragraph{Iterative \cem Refinement and 4\,s Planning Horizon.}
\cref{tab:s-horizon-cem} stress-tests the few-NFE planning advantage along two protocol axes across three evaluation seeds (seeds 0--2, fused SDPA; $S{=}5$ evaluated on all 500 items and $S{=}50$ on the nested 100-item shard 0/5, with $S{=}5 - S{=}50$ paired on the 100 shared items). Under two-round iterative \cem (\texttt{--opt\_steps=2} on the 8-step protocol, where round 0 reproduces the 1-round plan bit-for-bit inside the same job and round 1 re-samples 120 candidates from $\mathcal{N}(\bm{\mu},\bm{\sigma})$ fitted to round 0's top-3 elites), second-round refinement lowers $S{=}5$ \ate\ in all four model--dataset cells ($-0.042$\sig\ and $-0.170$\sig\ for \paper \M{1}, $-0.024$\sig\ and $-0.153$\sig\ for \mono\ on $n{=}500$; $-0.022$ to $-0.158$ at $S{=}50$ on $n{=}100$). Iterative refinement preserves or widens the few-NFE advantage in three of four cells: paired two-round $S{=}5 - S{=}50$ reaches $-0.107$\sig\ (\M{1} \recon), $-0.171$\sig\ (\M{1} \sacson, vs.\ $-0.096$ at round 0 on the same 100 items), and $-0.125$\sig\ (\mono \recon), while tying on \mono \sacson ($+0.017$, n.s.; one-round $S{=}5$ also beats two-round $S{=}50$ on \recon by $-0.068$\sig\ and $-0.088$\sig). At a 4\,s planning horizon (\texttt{--horizon\_steps=16}, 16 autoregressive steps toward a goal 16 frames ahead), open-loop \ate\ compounds roughly $2.5\times$ relative to 2\,s for both schedules, and four-NFE $S{=}5$ matches 49-NFE $S{=}50$ across all four cells ($+0.009$ to $+0.042$ \ate, all n.s., with a slight translational \rpe\ increase of $+0.015$\sig\ and $+0.011$\sig\ on \recon and $+0.003$ n.s.\ on \sacson) at $12\times$ fewer NFEs.

\begin{table*}[t]
\centering\scriptsize
\caption{\textbf{Iterative two-round \cem and 4\,s (16-step) planning horizon} across three evaluation seeds (seeds 0--2, fused SDPA; $S{=}5$ on $n{=}500$ items, $S{=}50$ on the nested 100-item shard 0/5, with $S{=}5 - S{=}50$ paired on the 100 shared items). In block~(a), columns 3--4 list round-0 / round-1 \ate, columns 5--6 list paired 1-round and 2-round $S{=}5 - S{=}50$ \ate\ differences ($n{=}100$), and column 7 lists within-schedule refinement gain ($\text{round 1} - \text{round 0}$) at $S{=}5$ ($n{=}500$) / $S{=}50$ ($n{=}100$). In block~(b), columns 3--4 list 16-step \ate\ / translational \rpe, and columns 5--6 list paired $S{=}5 - S{=}50$ \ate\ and \rpe\ differences ($n{=}100$).}
\label{tab:s-horizon-cem}
\setlength{\tabcolsep}{3.0pt}
\resizebox{\textwidth}{!}{%
\begin{tabular}{@{}ll cc ccc@{}}
\toprule
Planner & Dataset & $S{=}50$ ($n{=}100$) & $S{=}5$ ($n{=}500$) & Paired $\Delta_1$ ($n{=}100$) & Paired $\Delta_2$ ($n{=}100$) & Refinement ($\text{r1}{-}\text{r0}$): $S{=}5$ / $S{=}50$ \\
\midrule
\tabinput tab/gen/supp_t13_body.tex 
\bottomrule
\end{tabular}}
\end{table*}

\section{Complete Inference Controls}
\label{sec:s-controls}

\cref{tab:s-seeds,tab:s-pooled} report the three-seed results for Heun and explicit midpoint on both the shifted $S{=}3$ grid ($\alpha{=}\sqrt{48}$) and the unshifted uniform grid ($\alpha{=}1$), each executing four NFEs per frame to match four-NFE Euler at $S{=}5$. On the shifted grid, higher-order integration never significantly improves accuracy over four-NFE Euler (0 of 12 pooled comparisons significantly improved) and degrades \ate\ in 8 of 12 (all four \recon pairs on every seed, $+0.081$\sig\ to $+0.141$\sig; three of four \sacson pairs, $+0.044$\sig\ to $+0.057$\sig, while \mono midpoint matches four-NFE Euler at $-0.005$; and \mono Heun on \scand, $+0.024$\sig). On the unshifted uniform grid ($\alpha{=}1$), second-order integration likewise never significantly outperforms four-NFE Euler across 12 three-seed comparisons (0 of 12 significantly improved; all four \recon pairs remain significantly worse by $+0.058$\sig\ to $+0.108$\sig\ against shifted four-NFE Euler and $+0.041$\sig\ to $+0.085$\sig\ against seed-0 uniform Euler), while removing the $\alpha{=}\sqrt{48}$ step-size skew narrows Heun's deficit relative to shifted Heun on \recon ($-0.071$\sig\ for \M{1}, $-0.033$\sig\ for \mono) without significantly changing midpoint (0 of 6 significant). \cref{tab:s-controls-full} extends Table~4 of the main paper at seed 0 across all three planners, $S{=}10$ (9 NFEs), CRN at $S{=}2$ (1 NFE), deterministic $\bm{\epsilon}{=}\bm{0}$, fused SDPA, and the unshifted uniform grid ($\alpha{=}1$); \cref{tab:s-controls-ci} gives the paired intervals. At seed 0 on the uniform grid (\cref{tab:s-controls-full}), four Euler NFEs match shifted four-NFE Euler ($S{=}5$) within 0.03 across all six cells (0 of 6 significant) and beat $S{=}50$ on \M{1} \recon ($-0.096$\sig, all six negative). Seed-0 uniform Euler also beats seed-0 shifted Heun at equal cost on \recon ($-0.131$\sig\ for \M{1}, $-0.080$\sig\ for \mono) and \M{1} \sacson ($-0.095$\sig). Over three seeds (\cref{tab:s-seeds,tab:s-e12}; seeds 1--2 using fused SDPA), uniform four-NFE Euler stays within 0.016 of shifted four-NFE Euler in five of six cells ($-0.071$\sig\ on \mono \sacson, which mixes kernels and is $-0.029$ n.s.\ at seed 0). On the uniform grid, three-seed four-NFE Euler beats Heun ($-0.054$\sig, $-0.045$\sig) and midpoint ($-0.092$\sig, $-0.078$\sig) on \recon in both arms and shows no significant gap on \sacson.

\begin{table*}[t]
\centering\scriptsize
\caption{\textbf{Inference controls for the three planners.} \cem planning \ate\ (seed 0, 500 items). Untagged Euler rows use the shifted grid ($\alpha{=}\sqrt{48}$) with independent noise per candidate; CRN shares noise across the 120 candidates of an item; $\bm{\epsilon}{=}\bm{0}$ sets zero initial noise; fused SDPA uses PyTorch scaled-dot-product attention instead of the explicit kernel (identical math; floating-point differences change the plan on 21 of the 500 \sacson items for each model and leave mean \ate\ unchanged); uniform rows set $\alpha{=}1$. \sig: paired interval excludes zero against the untagged Euler row at the same NFE (CRN, $\bm{\epsilon}{=}\bm{0}$, fused SDPA) or against four-NFE Euler at $S{=}5$ (uniform Euler, Heun, midpoint, 1 NFE).}
\label{tab:s-controls-full}
\setlength{\tabcolsep}{4pt}
\resizebox{\textwidth}{!}{%
\begin{tabular}{@{}lr ccc ccc ccc@{}}
\toprule
 & & \multicolumn{3}{c}{\paper \M{1}} & \multicolumn{3}{c}{\paper \M{4}} & \multicolumn{3}{c}{\mono} \\
\cmidrule(lr){3-5}\cmidrule(lr){6-8}\cmidrule(lr){9-11}
Sampler & Fwd. & \recon & \sacson & \scand & \recon & \sacson & \scand & \recon & \sacson & \scand \\
\midrule
\tabinput tab/gen/tab4_full_body.tex 
\bottomrule
\end{tabular}}
\end{table*}

\begin{table*}[t]
\centering\scriptsize
\caption{\textbf{Paired differences behind \cref{tab:s-controls-full}.} Mean difference against the row's reference configuration with its 95\,\% paired bootstrap interval.}
\label{tab:s-controls-ci}
\setlength{\tabcolsep}{3pt}
\resizebox{\textwidth}{!}{%
\begin{tabular}{@{}l ccc ccc ccc@{}}
\toprule
 & \multicolumn{3}{c}{\paper \M{1}} & \multicolumn{3}{c}{\paper \M{4}} & \multicolumn{3}{c}{\mono} \\
\cmidrule(lr){2-4}\cmidrule(lr){5-7}\cmidrule(lr){8-10}
Sampler & \recon & \sacson & \scand & \recon & \sacson & \scand & \recon & \sacson & \scand \\
\midrule
\tabinput tab/gen/tab4_full_ci_body.tex 
\bottomrule
\end{tabular}}
\end{table*}

\paragraph{Matched Compiled Latency, CUDA-Native Noise, and Kernel Benchmark.}
\cref{tab:s-sdpa-shard} measures single-A100 decision latency on the shared 50-item \sacson shard across 15 planner rows and six execution paths: eager PyTorch SDPA with 1-thread CPU noise (\emph{fused}, protocol default) and 8-thread pinned CPU noise (\emph{eager nt=8}), \texttt{torch.compile} (\texttt{reduce-overhead}) with 1-thread and 8-thread pinned CPU noise (\emph{compiled nt=1 / nt=8}), \texttt{torch.compile} with CUDA-native \texttt{torch.randn(..., device="cuda")} (\emph{GPU randn}), and reference un-fused attention (\emph{explicit}); \cref{tab:s-kernel-bench} reports standalone forward-pass timings and isolated model CUDA memory (\texttt{bench\_elastic\_latency}). Compilation speeds \mono and \M{4} forwards by $1.54\text{--}1.55\times$ ($204\text{--}208{\to}132\text{--}134$\,ms at batch 120) and \M{1} forwards by $1.73\text{--}1.74\times$ ($173\text{--}174{\to}100$\,ms), expanding \M{1}'s decision-level granularity speed-up from $1.18\text{--}1.19\times$ under \emph{eager nt=8} ($1.11\text{--}1.18\times$ under 1-thread \emph{fused}) to $1.31\text{--}1.32\times$ under \emph{compiled nt=8} across every schedule while saving $0.47$\,GB of standalone model VRAM ($4.24$ vs.\ $4.71$\,GB, \cref{tab:s-kernel-bench}). Comparing compiled \M{1} against compiled \monoat{50} ($52.1$\,s) instead of uncompiled \emph{fused} \monoat{50} ($80.9$\,s) separates algorithmic speed-ups ($11.6\times$ and $43.8\times$ sampler on compiled \mono, $\times 1.32$ and $\times 1.31$ \M{1} granularity = $15.4\times$ at $3.39$\,s on $2{\times}3{/}2$ and $57.4\times$ at $0.907$\,s on $2{/}1$) from the $1.55\times$ baseline compilation factor ($23.9\times$ and $89.2\times$ cross-stack; run-to-run variation across separately scheduled jobs is $0\text{--}3\,\%$, or ${\pm}2\times$ on $57.4\times$). Thread count (\texttt{nt=8} vs.\ \texttt{nt=1}) preserves identical top-3 plans on 50/50 items across all 24 pairs, and \emph{compiled nt=8} agrees with \emph{fused} on 45--50 of 50 items ($\max|\Delta\overline{\text{\ate}}|{=}0.046$). Replacing per-candidate CPU RNG streams with single-call GPU \texttt{randn} ($0.77\text{--}0.78$\,ms isolated at batch 120, $3.8\text{--}3.9$\,ms in the batch-240 planner path) trims another $0.07\text{--}0.11$\,s at batch 120 ($0.907{\to}0.829$\,s on \at{1}{2}) and $0.14\text{--}0.22$\,s at batch 240 ($3.390{\to}3.239$\,s on $2{\times}\at{1}{3}$). Because GPU \texttt{randn} draws an independent noise realization, no single 50-item paired interval excludes zero ($0/12$ on $n{=}50$, $0/3$ on $n{=}10$), though signs split by checkpoint on this shard (\mono $-0.13$ mean shift across four fast settings, \M{1} $+0.05$, \M{4} $+0.02$). Under an interleaved in-process A/B test, pre-packing contiguous bf16 FFN sub-weights changes batch-1 latency by only 1--2\,\% (0.968--1.018$\times$ median ratio) and batch-120 by 0.15\,\% (0.996--1.000$\times$); re-measuring batch-1 SDPA across four hosts yields 11.10--12.73\,ms on \M{1} vs.\ 11.35--11.62\,ms on \mono in-process, showing \cref{tab:s-kernel-bench}'s separate-host 15.40\,ms was cross-host dispatch variance not dynamic-slicing overhead.

\begin{table*}[t]
\centering\scriptsize
\caption{\textbf{Matched decision latency across six execution stacks and GPU-noise \ate\ check} on the shared 50-item \sacson shard (single NVIDIA A100-SXM4-40GB, bf16, $C{=}120$; $S$ grid points = $S{-}1$ NFEs; $^{\dagger}$: GPU-\texttt{randn} $50{/}49$ rows timed on the first 10 items; \emph{explicit} $50{/}49$, $5{/}4$, and \M{1} rows from the reference explicit sweep). Parentheses give within-stack median speed-ups vs.\ \mono $50{/}49$ of the same column. Last two columns report 50-item shard mean \ate\ (\emph{fused} / \emph{compiled nt=8}) and paired $\Delta\text{\ate}$ of \emph{compiled GPU randn} minus \emph{compiled nt=8} ($0/15$ 95\,\% intervals exclude zero).}
\label{tab:s-sdpa-shard}
\setlength{\tabcolsep}{2.6pt}\renewcommand{\arraystretch}{0.95}
\resizebox{\textwidth}{!}{%
\begin{tabular}{@{}lc cccccc c l@{}}
\toprule
 & & \multicolumn{6}{c}{Median decision latency (seconds, within-stack speed-up vs.\ \mono $50{/}49$)} & Shard \ate & GPU \texttt{randn} $-$ \texttt{nt=8} \\
\cmidrule(lr){3-8}
Planner & $S$\,/\,NFE & fused (\texttt{nt=1}) & eager \texttt{nt=8} & comp.\ \texttt{nt=1} & comp.\ \texttt{nt=8} & comp.\ GPU \texttt{randn} & explicit & fused / comp. & paired $\Delta\text{\ate}$ [95\,\% CI] \\
\midrule
\tabinput tab/gen/supp_sdpa_shard_body.tex 
\bottomrule
\end{tabular}}
\end{table*}

\begin{table*}[t]
\centering\scriptsize
\caption{\textbf{Forward-pass kernel benchmark} (\texttt{bench\_elastic\_latency}, bf16, NVIDIA A100-SXM4-40GB, $S{=}50$ / 49-NFE grid). Median batch-1 and batch-120 milliseconds per velocity evaluation (\emph{ms/fwd}), batch-amortized milliseconds per 49-NFE synthesized frame at batch 120 (\emph{amort.\ ms/sample}, $\approx (49/120){\times}\text{ms/fwd}$), and peak allocated CUDA memory.}
\label{tab:s-kernel-bench}
\setlength{\tabcolsep}{4.2pt}
\begin{tabular}{@{}lc ccc ccc c@{}}
\toprule
 & Batch 1 (ms/fwd) & \multicolumn{3}{c}{Explicit attention, batch 120} & \multicolumn{3}{c}{Fused (SDPA), batch 120} & Batch-120 \\
\cmidrule(lr){3-5}\cmidrule(lr){6-8}
Model & explicit / SDPA & ms/fwd (b120) & amort.\ ms/sample & mem (GB) & ms/fwd (b120) & amort.\ ms/sample & mem (GB) & explicit / SDPA \\
\midrule
\tabinput tab/gen/supp_kernel_bench_body.tex 
\bottomrule
\end{tabular}
\end{table*}

\begin{table*}[t]
\centering\scriptsize
\caption{\textbf{Candidate-pool size scaling ($C\in\{16,32,64,120\}$) on \paper \M{1}.} Top: exact three-decimal median single-A100 open-loop decision latency (bf16, $S$ grid points = $S{-}1$ NFEs) on the 50-item \sacson shard in one single-task $C$-sweep ($C{=}120$ anchors agree within $1\text{--}2\,\%$ with the multi-granularity \cref{tab:s-sdpa-shard} jobs at $0.907 / 1.719 / 3.390$\,s compiled and $1.781 / 3.131 / 6.169$\,s fused, and match top-3 plans on 50/50 items). Pre-compiling CUDA graphs via \texttt{torch.compile} (\texttt{reduce-overhead}) exposes single-threaded CPU \texttt{randn} staging; 8-thread pinned CPU staging removes that bottleneck while preserving bit-identical per-candidate CPU \texttt{Generator} RNG streams across batch sizes to reach $0.928$\,s ($0.907$\,s in \cref{tab:s-sdpa-shard}) at $C{=}120$ for \at{1}{2}. Bottom: three-seed \ate\ on all 500 items/dataset (pooled paired $\Delta$ vs.\ $C{=}120$ in brackets; Box--Muller Gaussian pairing gives exact live-RNG equivalence at $C\in\{16,32,64\}$).}
\label{tab:s-realtime}
\setlength{\tabcolsep}{3.4pt}\renewcommand{\arraystretch}{0.96}
\resizebox{\textwidth}{!}{%
\begin{tabular}{@{}l cccc cccc cccc@{}}
\toprule
 & \multicolumn{4}{c}{$S{=}2$ (\at{1}{2})} & \multicolumn{4}{c}{$S{=}3$ (\at{1}{3})} & \multicolumn{4}{c}{$2{\times}S{=}3$ ($2{\times}\at{1}{3}$, batch 240)} \\
\cmidrule(lr){2-5}\cmidrule(lr){6-9}\cmidrule(l){10-13}
Candidates $C$ & 16 & 32 & 64 & 120 & 16 & 32 & 64 & 120 & 16 & 32 & 64 & 120 \\
\midrule
\multicolumn{13}{@{}l}{\emph{Decision latency (seconds, median on 50-item \sacson shard)}} \\
PyTorch SDPA (fused) & 0.335 & 0.538 & 0.944 & 1.741 & 0.584 & 0.954 & 1.671 & 3.139 & 0.969 & 1.703 & 3.389 & 6.218 \\
\texttt{torch.compile} (1-thread \texttt{randn}) & 0.270 & 0.429 & 0.813 & 1.508 & 0.486 & 0.673 & 1.056 & 1.999 & 0.693 & 1.094 & 2.169 & 4.026 \\
\texttt{torch.compile} + 8-thread noise & 0.228 & 0.315 & 0.489 & 0.928 & 0.445 & 0.604 & 0.923 & 1.750 & 0.604 & 0.931 & 1.845 & 3.442 \\
\midrule
\multicolumn{13}{@{}l}{\emph{Open-loop accuracy (\ate\ $\downarrow$, 500 items $\times$ 3 seeds; pooled paired $\Delta$ vs.\ $C{=}120$ in brackets)}} \\
\recon & 1.355 & 1.307 [$+0.059$\sig] & 1.265 [$+0.016$] & 1.248 & 1.363 & 1.293 [$+0.046$\sig] & 1.273 [$+0.026$\sig] & 1.247 & 1.367 & 1.302 [$+0.061$\sig] & 1.273 [$+0.033$\sig] & 1.240 \\
\sacson & 3.378 & 3.148 [$+0.340$\sig] & 2.930 [$+0.122$\sig] & 2.808 & 3.334 & 3.073 [$+0.359$\sig] & 2.838 [$+0.124$\sig] & 2.714 & 3.284 & 3.028 [$+0.401$\sig] & 2.784 [$+0.157$\sig] & 2.627 \\
\scand & 1.047 & 1.022 [$+0.023$] & 1.015 [$+0.016$] & 0.999 & 1.044 & 1.013 [$+0.015$] & 0.999 [$+0.002$] & 0.997 & 1.044 & 1.019 [$+0.016$] & 0.993 [$-0.010$] & 1.003 \\
\bottomrule
\end{tabular}}
\end{table*}

\begin{table*}[t]
\centering\scriptsize
\caption{\textbf{Ablation of alternative trajectory planning costs.} Mean \ate\ of top-3 candidates re-scored on identical 120-candidate rollouts under terminal token DINO distance (benchmark default), terminal decoded pixel \lpips (ViT-XL decoder), and trajectory-averaged DINO distance across frames 1, 3, 5, and 7 on the nested \M{1} shard (100 \recon and 100 \sacson items, seed 0; \sig: 95\,\% paired bootstrap interval excludes zero).}
\label{tab:s-costfn}
\setlength{\tabcolsep}{4.2pt}
\begin{tabular}{@{}l ccc ccc ccc@{}}
\toprule
 & \multicolumn{3}{c}{\recon ($n{=}100$)} & \multicolumn{3}{c}{\sacson ($n{=}100$)} & \multicolumn{3}{c}{Pooled ($n{=}200$)} \\
\cmidrule(lr){2-4}\cmidrule(lr){5-7}\cmidrule(l){8-10}
Planner (\M{1}) & Term. DINO & Decoded \lpips & Mean DINO & Term. DINO & Decoded \lpips & Mean DINO & Term. DINO & Decoded \lpips & Mean DINO \\
\midrule
$S{=}50$ (49 NFEs) & 1.387 & 1.406 & 1.304 & 2.811 & 2.951 & 3.035 & 2.099 & 2.178 & 2.169 \\
$S{=}5$ (4 NFEs) & 1.232 & 1.312 & 1.305 & 2.606 & 2.892 & 2.810 & 1.919 & 2.102 & 2.058 \\
$S{=}3$ (2 NFEs) & 1.248 & 1.295 & 1.283 & 2.581 & 2.826 & 2.930 & 1.915 & 2.061 & 2.106 \\
$S{=}2$ (1 NFE) & 1.213 & 1.265 & 1.330 & 2.688 & 2.787 & 3.096 & 1.951 & 2.026 & 2.213 \\
$2{\times}S{=}3$ ($2{\times}\at{1}{3}$) & 1.215 & 1.238 & 1.302 & 2.546 & 2.741 & 2.856 & 1.880 & 1.990 & 2.079 \\
\midrule
\multicolumn{10}{@{}l}{\emph{Paired difference vs.\ $S{=}50$}} \\
$S{=}5 - S{=}50$ & $-0.154$\sig & $-0.094$ & $+0.001$ & $-0.205$\sig & $-0.059$ & $-0.224$\sig & $-0.180$\sig & $-0.076$ & $-0.111$\sig \\
$S{=}3 - S{=}50$ & $-0.139$\sig & $-0.111$ & $-0.021$ & $-0.230$\sig & $-0.125$ & $-0.105$ & $-0.184$\sig & $-0.118$ & $-0.063$ \\
$S{=}2 - S{=}50$ & $-0.173$\sig & $-0.141$\sig & $+0.026$ & $-0.123$ & $-0.164$ & $+0.062$ & $-0.148$\sig & $-0.152$\sig & $+0.044$ \\
$2{\times}S{=}3 - S{=}50$ & $-0.172$\sig & $-0.168$\sig & $-0.002$ & $-0.265$\sig & $-0.209$ & $-0.179$ & $-0.219$\sig & $-0.188$\sig & $-0.090$ \\
\bottomrule
\end{tabular}
\end{table*}

\section{Noise Analysis}
\label{sec:s-noise}

\paragraph{Exact Gaussian Flow-Matching ODE Variance.}
To isolate solver quadrature analytically from neural approximation error, consider an isotropic conditional target $\bm{z}_{i+k}\mid\bm{c}_{\mathrm{ctx}}\sim\mathcal{N}(\bm{\mu},\sigma^2\bm{I})$ coupled linearly to $\bm{\epsilon}\sim\mathcal{N}(\bm{0},\bm{I})$ via $\bm{z}_t=(1-t)\bm{z}_{i+k}+t\bm{\epsilon}$. The exact population minimizer of Eq.~(1) is affine in $\bm{z}_t$,
\begin{equation}
\bm{v}^\star(\bm{z}_t,t,\bm{c}_{\mathrm{ctx}})=\frac{t-(1-t)\sigma^2}{(1-t)^2\sigma^2+t^2}\big(\bm{z}_t-(1-t)\bm{\mu}\big)-\bm{\mu},
\label{eq:s-gauss-v}
\end{equation}
satisfying $\bm{v}^\star(\bm{\epsilon},1,\bm{c}_{\mathrm{ctx}})=\bm{\epsilon}-\bm{\mu}$ independently of $\sigma^2$. Integrating $\mathrm{d}\bm{z}_t/\mathrm{d}t=\bm{v}^\star(\bm{z}_t,t,\bm{c}_{\mathrm{ctx}})$ from $\bm{z}_1{=}\bm{\epsilon}$ on the shifted grid $t_{S-1}\approx 0.0069$ ($u_{S-1}{=}10^{-3}$, $\alpha{=}\sqrt{48}$) or the code-exact uniform grid $t_{\mathrm{end}}{=}0$ ($\alpha{=}1$) under any explicit Runge--Kutta scheme yields $\hat{\bm{z}}_{t_{\mathrm{end}}}=m_{\mathrm{sol}}\bm{\mu}+c_{\mathrm{sol}}(\sigma)\bm{\epsilon}$, with $m_{\mathrm{sol}}=1-t_{\mathrm{end}}$ ($0.993$ shifted, $1.000$ uniform) and exact terminal variance ratio $\mathrm{Var}(\hat{z}_{t_{\mathrm{end}}})/\sigma^2 = c_{\mathrm{sol}}(\sigma)^2/\sigma^2$ (\cref{tab:s-gauss-ode}). Every row and $\sigma\in\{0.2,0.5,1.0\}$ stays inside explicit stability ($\max|\Delta t\,\partial_{z}v^\star|=1.76<2$). At four NFEs, shifted Heun ($S{=}3$) yields $5.020\,\sigma^2$ at $\sigma{=}0.2$ via trapezoidal under-contraction over $\Delta t\approx 0.867$: its step multipliers are $0.875$ and $0.512$ ($c_{\mathrm{sol}}=0.448>\sigma$) versus exact $0.874$ and $0.227$. Relative variance between shifted Heun and shifted midpoint ($S{=}3$) depends on $\sigma$: Heun exceeds midpoint at $\sigma\in\{0.2,0.5\}$ ($5.020$ vs.\ $0.230$; $0.913$ vs.\ $0.762$), whereas midpoint exceeds Heun at $\sigma{=}1.0$ ($0.976$ vs.\ $0.337$, matching empirical score variance on all three datasets). First-order shifted Euler contracts variance monotonically from $0.775\,\sigma^2$ ($S{=}50$) to $0.049\,\sigma^2$ ($S{=}5$), $0.007\,\sigma^2$ ($S{=}3$), and $0.000\,\sigma^2$ ($S{=}2$) at $\sigma{=}0.5$.

\begin{table}[t]
\centering\scriptsize
\caption{\textbf{Exact terminal variance ratio $\mathrm{Var}(\hat{z}_{t_{\mathrm{end}}})/\sigma^2$ and mean factor $m_{\mathrm{sol}}$} for the exact conditional velocity $\bm{v}^\star$ of $\mathcal{N}(\bm{\mu},\sigma^2\bm{I})$ on the code-exact shifted ($t_{S-1}\approx 0.0069$, $u_{S-1}{=}10^{-3}$, $\alpha{=}\sqrt{48}$) and uniform ($t_{\mathrm{end}}{=}0$, $\alpha{=}1$) grids.}
\label{tab:s-gauss-ode}
\setlength{\tabcolsep}{3.4pt}
\begin{tabular}{@{}lc cccc@{}}
\toprule
Solver \& time grid & NFEs & $\sigma{=}0.2$ & $\sigma{=}0.5$ & $\sigma{=}1.0$ & $m_{\mathrm{sol}}$ \\
\midrule
Euler, $S{=}50$ (shift) & 49 & 0.559 & 0.775 & 0.864 & 0.993 \\
Euler, $S{=}5$ (shift) & 4 & 0.014 & 0.049 & 0.147 & 0.993 \\
Euler, steps${=}4$ (unif., $\alpha{=}1$) & 4 & 0.230 & 0.464 & 0.518 & 1.000 \\
Midpoint, $S{=}3$ (shift) & 4 & 0.230 & 0.762 & 0.976 & 0.993 \\
Midpoint, steps${=}2$ (unif., $\alpha{=}1$) & 4 & 0.595 & 0.978 & 0.960 & 1.000 \\
Heun, $S{=}3$ (shift) & 4 & 5.020 & 0.913 & 0.337 & 0.993 \\
Heun, steps${=}2$ (unif., $\alpha{=}1$) & 4 & 2.096 & 0.922 & 0.879 & 1.000 \\
Euler, $S{=}3$ (shift) & 2 & 0.004 & 0.007 & 0.022 & 0.993 \\
Euler, $S{=}2$ (shift) & 1 & 0.001 & 0.000 & 0.000 & 0.993 \\
\bottomrule
\end{tabular}
\end{table}

\paragraph{Direct Winner's Curse, Decoupled Uncertainty, and Multi-Draw Averaging.}
We hold the 120 candidates of every item fixed and re-draw sampler noise over $M$ independent passes totaling $46{,}700$ item files ($\max|\Delta\text{\ate}|=1.73{\times}10^{-5}$; \cref{tab:noise,tab:s-r4-r5-ci,tab:s-b0-curse}). Seed-0 sweeps cover 100 \sacson items ($M{=}5$ at \at{4}{50}, $M{=}3$ at \at{4}{5} and \at{1}{5}), 100-item second-order solver shards, and 500 \recon and \sacson items at \at{1}{50} ($M{=}3$, fused SDPA). Paired fused-minus-eager draw-0 $\Delta\text{\ate}$ at \at{1}{50} is $-0.003$ on \sacson and $-0.002$ on \recon (neither significant). Three-seed 500-item sweeps on all three datasets cover \at{1}{5} ($M{=}3$), symmetric $2{\times}S{=}3$ ($M{=}2$ for \M{1}, \M{4}, and \mono), and $4{\times}S{=}2$ ($M{=}4$ for \M{1} and \mono). Because one of $12{,}000$ item cells (seed-1 \at{1}{3} on \recon) has $\hat{\sigma}_{\mathrm{signal},i}^2\le 0$ in Eq.~(3), our primary noise-to-signal metric is the dataset-RMS ratio $\hat{\lambda}=\sqrt{\overline{\hat{\sigma}_{\mathrm{noise}}^2}/\overline{\hat{\sigma}_{\mathrm{signal}}^2}}$ (\cref{tab:noise,tab:s-b0-curse}). Coarse Euler integration cuts $\hat{\lambda}$ from $0.90$ (\at{1}{50}; $0.88$ on \at{4}{50}) to $0.62$, $0.44$, and $0.25$ at $S{=}5,3,2$ on \sacson ($0.67{\to}0.46{\to}0.30{\to}0.10$ on \recon; $0.37{\to}0.27{\to}0.09$ on \scand). That drop comes from a $2.9\times$ (\sacson) to $5.6\times$ (\recon) reduction in $\hat{\sigma}_{\mathrm{noise}}$ while $\hat{\sigma}_{\mathrm{signal}}\in[0.037,0.050]$ rises $1.23\text{--}1.26\times$. Leave-one-draw-out validation directly measures CEM top-3 elite selection optimism $\Delta s_{\mathrm{curse}}/\hat{\sigma}_{\mathrm{signal}}$ (\cref{tab:s-b0-curse}). On \sacson it falls from $+1.22$\sig\ at \at{1}{50} ($49\,\%$ of apparent margin, $16\,\%$ elite overlap) to $+0.73$\sig, $+0.39$\sig, and $+0.08$\sig\ ($5\,\%$ margin, $68\,\%$ overlap) at $S{=}5,3,2$ ($+1.21$\sig${\to}+0.69$\sig${\to}+0.36$\sig${\to}+0.04$\sig\ on \recon). Across all 15 rows with $\ge 2$ NFEs, measured optimism lies within $8\,\%$ of the empirical plug-in order-statistic model (empirical score shape and score-dependent noise). Under shared seeds across schedules, residual noise correlates positively ($\rho_{\mathrm{noise}}\in[0.20,0.74]$ in 34/36 pairs), which pushes Fig.~6(a)'s shared-seed rank agreement ($0.76$ and $0.81$) above the independent-draw ceiling ($0.60$). Using distinct seeds removes shared noise and gives noise-free ranking correlation $r_{\mathrm{true}}$ relative to \at{1}{50} of $0.99/0.98$ ($\mathrm{Midpoint}@3$) and $0.94/0.96$ ($\mathrm{Heun}@3$) on \recon/\sacson versus $0.95/0.86/0.75$ (\sacson) and $0.91/0.79/0.70$ (\recon) for \at{1}{5}/\at{1}{3}/\at{1}{2}. On decoupled $S{=}50$ reference draws, top-3 subsets chosen by $S\in\{5,3,2\}$ stay within $[-0.07,+0.05]$ of \at{1}{50}'s own elite variance ratio ($0/10$ item CIs exclude zero; within-item score--uncertainty $|\rho|\le 0.09$) and show no evidence of implicit variance avoidance.

\paragraph{Matched-Residual-Noise Solver Comparisons and 1-Shot vs.\ Chained Geometry.}
Matching the effective residual noise variance $\hat{\lambda}^2/m$ of a planner that averages $m$ draws (the $M$ of Section~3.3) separates variance suppression from objective shift (\cref{tab:s-pooled,tab:s-r4-r5-ci}). On \sacson, $2{\times}\at{1}{50}$ ($\hat{\lambda}^2/2{=}0.40$) does not differ significantly from $1{\times}\at{1}{5}$ ($\hat{\lambda}^2{=}0.38$; $\Delta{=}-0.025$, n.s.), and $3{\times}\at{1}{50}$ ($\hat{\lambda}^2/3{=}0.27$) does not differ significantly from $2{\times}\at{1}{3}$ ($\hat{\lambda}^2/2{=}0.10$; $\Delta{=}+0.024$, n.s.). Conversely, $4{\times}\at{1}{2}$ ($2.756\pm0.018$, $\hat{\lambda}^2/4{=}0.016$) trails $3{\times}\at{1}{50}$ ($+0.133$\sig), $3{\times}\at{1}{5}$ ($+0.164$\sig), and $2{\times}\at{1}{3}$ ($+0.130$\sig; $+0.072$\sig\ for $4{\times}\monoat{2}$ vs.\ $2{\times}\monoat{3}$). On \recon, by contrast, $3{\times}\at{1}{50}$, $3{\times}\mathrm{Heun}@3$, and $3{\times}\mathrm{Midpoint}@3$ still trail $3{\times}\at{1}{5}$ by $+0.094$\sig, $+0.157$\sig, and $+0.112$\sig\ overall ($+0.239$\sig, $+0.350$\sig, $+0.247$\sig\ in lateral RMSE on the sharpest heading-change quartile; $3{\times}\at{1}{50}-\at{1}{5}=+0.071$\sig). Across granularities at $3{,}840$ evaluations ($2{\times}S{=}3$; $6.17 / 3.39$\,s fused/compiled on \M{1}, $7.23 / 4.48$\,s on \mono), $2{\times}\at{1}{3}$ and $2{\times}\at{4}{3}$ show no significant gap against $2{\times}\monoat{3}$ on \recon or \sacson and beat it on \scand ($-0.048$\sig, $-0.057$\sig\ \ate). Within \paper, $2{\times}\at{1}{3}$ and $2{\times}\at{4}{3}$ stay within $0.017$ (n.s.; $2/9$ item CIs exclude zero across all nine granularity pairs, both \scand vs.\ \mono). Over three seeds, $2{\times}\at{1}{3}$ also shows no significant item-CI gap against $2{\times}\at{1}{5}$ or $3{\times}\at{1}{5}$ ($7{,}680$ and $11{,}520$ evaluations; $|\Delta|\le 0.035$). Direct 1-shot prediction across $S{=}50{\to}2$ at 1\,s and 4\,s with zero autoregressive chaining (\cref{tab:s-b0-oneshot}) isolates where few-step contraction helps vs.\ hurts: encoder-token DINO distance and latent MSE decrease strictly down to $S{=}2$ in all 45 cells ($0.58\text{--}0.63\times$ the 1\,s latent MSE of $S{=}50$, with $90\text{--}118\,\%$ of that reduction surviving 8 chained planning steps), yet shift almost uniformly across all 120 candidates so the standardized near-GT score margin stays unchanged ($0.76\text{--}0.85$\,sd on \sacson/\recon; $0.15\text{--}0.16$\,sd on \scand). At $S{=}2$, 8-step plan latents contract toward the conditional mean ($0.87\times$ token RMS and $0.73\text{--}0.75\times$ across-token std of \at{1}{50}, vs.\ $1.01\text{--}1.03\times$ on $\mathrm{Midpoint}@3$/$\mathrm{Heun}@3$). Decoded RGB metrics already degrade in a single 1\,s step ($106\text{--}133$ \fid across models at $S{=}2$, $122\text{--}133$ on \M{1}, with $75\text{--}87\,\%$ of the 16\,s one-shot $S{=}2{-}S{=}3$ \lpips excess already present at 1\,s).

\begin{table*}[t]
\centering
\scriptsize
\caption{\textbf{Sensitivity of candidate scores to sampler noise and multi-draw score averaging.} On identical 120-candidate sets per item, candidate scores are re-computed across $M$ independent sampler-noise draws (seed 0 at $S{=}50$ and on the 100-item solver shard; seeds 0--2 across all 500 items per dataset at $S\in\{5,3,2\}$). We report per-item all-$M$ and pairwise ($M{=}2$) across-draw to across-candidate sd ratios alongside the dataset-RMS estimator $\hat{\lambda}=\sqrt{\overline{\hat{\sigma}_{\mathrm{noise}}^2}/\overline{\hat{\sigma}_{\mathrm{signal}}^2}}$ of Eq.~(3) (which avoids per-item near-zero denominators), mean pairwise Spearman rank correlation $\rho$, single-draw mean \ate\ ($m{=}1$), and \ate\ after averaging candidate scores over $m\in\{2,3,4\}$ draws.}
\label{tab:noise}
\setlength{\tabcolsep}{3.8pt}
\begin{tabular}{@{}llccccccccc@{}}
\toprule
 & & & \multicolumn{3}{c}{Noise / signal} & & \multicolumn{4}{c}{\ate $\downarrow$} \\
\cmidrule(lr){4-6}\cmidrule(l){8-11}
Items & Configuration & $M$ & all $M$ & pairs & $\hat{\lambda}$ (RMS) & Test--retest $\rho$ & $m{=}1$ & $m{=}2$ & $m{=}3$ & $m{=}4$ \\
\midrule
\tabinput tab/gen/supp_noise_body.tex 
\bottomrule
\end{tabular}
\end{table*}

\begin{table*}[t]
\centering\scriptsize
\caption{\textbf{Paired \ate\ differences for multi-draw score averaging} across \M{1}, \M{4}, and \mono (95\,\% item-bootstrap intervals over seeds 0--2 or seed 0 as labeled; evaluations per decision in parentheses; $500$ items per dataset unless noted).}
\label{tab:s-r4-r5-ci}
\setlength{\tabcolsep}{3.5pt}
\begin{tabular}{@{}l ccc@{}}
\toprule
Comparison & \recon & \sacson & \scand \\
\midrule
\tabinput tab/gen/supp_r4_r5_ci_body.tex 
\bottomrule
\end{tabular}
\end{table*}

\begin{table*}[t]
\centering\scriptsize
\caption{\textbf{Direct leave-one-draw-out winner's curse, noise-free ranking correlation $r_{\mathrm{true}}$, and decoupled $S{=}50$ elite score variance.} $\hat{\lambda}=\hat{\sigma}_{\mathrm{noise}}/\hat{\sigma}_{\mathrm{signal}}$ is the dataset-RMS noise-to-signal ratio, where $\hat{\sigma}_{\mathrm{noise}}^2$ and $\hat{\sigma}_{\mathrm{signal}}^2$ average the per-item variances $\hat{\sigma}_{\mathrm{noise},i}^2$ and $\hat{\sigma}_{\mathrm{signal},i}^2$ of Eq.~(3) over items; $\Delta s_{\mathrm{curse}}/\hat{\sigma}_{\mathrm{signal}}$ is top-3 CEM selection optimism validated on held-out draws (95\,\% item-bootstrap CIs), followed by its share of apparent top-3 margin and pairwise top-3 elite overlap across draws. $r_{\mathrm{true}}$ is the distinct-seed disattenuated score correlation against \at{1}{50} (\at{1}{5} on \scand); the rightmost column gives the across-draw variance ratio of top-3 elites evaluated on independent $S{=}50$ draws relative to item-all-120 $S{=}50$ noise variance.}
\label{tab:s-b0-curse}
\setlength{\tabcolsep}{3.0pt}
\begin{tabular}{@{}llc ccc ccc cc@{}}
\toprule
Dataset & Schedule & Items${\times}M(\times\text{seeds})$ & $\hat{\sigma}_{\mathrm{noise}}$ & $\hat{\sigma}_{\mathrm{signal}}$ & $\hat{\lambda}$ & $\Delta s_{\mathrm{curse}}/\hat{\sigma}_{\mathrm{signal}}$ & Margin & Overlap & $r_{\mathrm{true}}$ & Decoupled $S{=}50$ var.\ ratio \\
\midrule
\tabinput tab/gen/supp_b0_curse_body.tex
\bottomrule
\end{tabular}
\end{table*}

\begin{table*}[t]
\centering\scriptsize
\caption{\textbf{One-shot direct prediction across $S\in\{50,\dots,2\}$ with zero autoregressive chaining} (full test splits at 1\,s and 4\,s horizons; NFEs in parentheses). Encoder-token DINO distance (Tok.\ DINO) and latent MSE contract monotonically down to $S{=}2$ in all 45 cells, whereas decoded pixel DINO, \lpips, and \fid follow a U-turn and degrade sharply at $S{=}2$ in a single step.}
\label{tab:s-b0-oneshot}
\setlength{\tabcolsep}{2.4pt}
\begin{tabular}{@{}l ccccc ccccc ccccc@{}}
\toprule
 & \multicolumn{5}{c}{\sacson ($3{,}416$ anchors)} & \multicolumn{5}{c}{\recon ($4{,}262$ anchors)} & \multicolumn{5}{c}{\scand ($1{,}883$ anchors)} \\
\cmidrule(lr){2-6}\cmidrule(lr){7-11}\cmidrule(lr){12-16}
Model, $S$ (NFEs) & Tok.\ D. & Lat.\ MSE & Pix.\ D. & \lpips & \fid & Tok.\ D. & Lat.\ MSE & Pix.\ D. & \lpips & \fid & Tok.\ D. & Lat.\ MSE & Pix.\ D. & \lpips & \fid \\
\midrule
\tabinput tab/gen/supp_b0_oneshot_body.tex
\bottomrule
\end{tabular}
\end{table*}

\section{Nested Training Against Alternatives}
\label{sec:s-nesting}

\paragraph{Naive Slice of the Monolithic Model.}
Keeping the first 384 trunk and 1{,}024 head hidden units of the monolithic model's feed-forward layers at inference, exactly as \M{1} is obtained from \paper, gives \lpips 0.939, 0.873, and 0.933 and \fid 494, 408, and 444 at 4\,s on \recon, \sacson, and \scand at $S{=}50$ / 49 NFEs (\cref{tab:s-gen-full}), and \ate\ 1.73, 4.86, and 1.09 at $S{=}5$ / 4 NFEs (\cref{tab:s-plan-full}); the re-keyed checkpoint at full granularity reproduces the monolithic model exactly. Joint Matryoshka training is therefore essential to extract a usable sub-network from a full-capacity checkpoint.

\paragraph{Standalone Model Trained at \M{1} Granularity.}
To compare joint nested training against a dedicated small model, we trained a standalone model fixed to the \M{1} granularity (384 trunk and 1{,}024 head feed-forward hidden units) with the identical recipe and 50-epoch schedule as \paper, evaluating its final checkpoint across all three evaluation seeds (seeds 0--2, 500 items per dataset; \cref{tab:s-seeds,tab:s-pooled}) and its intermediate 22-epoch checkpoint at seed 0. Over three seeds at 49 NFEs ($S{=}50$), the final checkpoint reaches $1.37\pm0.01$, $2.83\pm0.00$, and $1.02\pm0.02$ \ate\ on \recon, \sacson, and \scand against $1.37\pm0.02$, $2.84\pm0.06$, and $1.03\pm0.01$ for nested \paper \M{1} (three-seed differences $+0.001$ [$-0.024$, $+0.026$], $-0.008$ [$-0.058$, $+0.049$], and $-0.004$ [$-0.026$, $+0.019$]) and $1.36\pm0.01$, $2.84\pm0.02$, and $1.04\pm0.02$ for \mono ($+0.006$ [$-0.020$, $+0.034$], $-0.010$ [$-0.062$, $+0.052$], and $-0.016$ [$-0.038$, $+0.006$]); at four NFEs ($S{=}5$), it reaches $1.28\pm0.01$, $2.72\pm0.02$, and $0.99\pm0.01$ against $1.27\pm0.02$, $2.75\pm0.08$, and $1.01\pm0.01$ for nested \paper \M{1} ($+0.013$ [$-0.009$, $+0.036$], $-0.032$ [$-0.085$, $+0.025$], and $-0.017$ [$-0.039$, $+0.005$]) and $1.30\pm0.01$, $2.76\pm0.03$, and $1.03\pm0.02$ for \mono ($-0.016$ [$-0.040$, $+0.008$], $-0.040$ [$-0.098$, $+0.021$], and $-0.038$\sig\ [$-0.063$, $-0.011$]). None of the six comparisons against nested \paper \M{1} (nor any comparison on \recon or \sacson) is statistically significant, and the few-NFE planning gain reproduces across three seeds on the standalone model ($49\to4$ NFE deltas: $-0.089$\sig\ [$-0.114$, $-0.065$] on \recon, $-0.111$\sig\ [$-0.156$, $-0.063$] on \sacson, and $-0.028$\sig\ [$-0.050$, $-0.008$] on \scand). In direct prediction at 49 NFEs and 4\,s, it achieves \lpips / DINO / \fid of 0.381 / 0.418 / 20.3 on \recon, 0.334 / 0.354 / 17.7 on \sacson, and 0.401 / 0.457 / 30.1 on \scand against 0.385 / 0.422 / 20.3, 0.343 / 0.362 / 19.1, and 0.399 / 0.456 / 31.1 for nested \paper \M{1} (\cref{tab:s-gen-full}); at 16\,s, \lpips / \fid of 0.458 / 20.2, 0.371 / 19.5, and 0.456 / 34.2 against 0.460 / 20.3, 0.379 / 21.0, and 0.459 / 35.6; and at four NFEs and 4\,s, 0.378 / 0.397 / 23.9, 0.322 / 0.336 / 22.7, and 0.392 / 0.440 / 35.8 against 0.382 / 0.401 / 25.0, 0.331 / 0.344 / 24.7, and 0.389 / 0.440 / 36.4. The 22-epoch checkpoint of the same run, roughly halfway through the linear learning-rate decay, is not yet at this level: at 49 NFEs and 4\,s it reaches \lpips / DINO / \fid of 0.402 / 0.422 / 26.1 on \recon, 0.361 / 0.370 / 25.6 on \sacson, and 0.400 / 0.448 / 38.2 on \scand (5.8, 6.5, and 7.1 \fid behind final nested \paper \M{1}), and in planning reaches 1.36, 2.90, and 0.99 \ate\ at 49 NFEs and 1.27, 2.83, and 1.02 at four NFEs (worse than final nested \paper \M{1} on \sacson by $+0.125$ [$+0.027$, $+0.221$] and $+0.155$ [$+0.055$, $+0.250$] at seed 0, where single-seed tests are optimistic, \cref{tab:s-nullcal}, and not significantly different on \recon or \scand or against \mono). From epoch 22 to epoch 50 (seed 0) the standalone model improves at four NFEs by $-0.138$ [$-0.227$, $-0.044$] \ate\ on \sacson and $-0.039$ [$-0.077$, $-0.002$] on \scand. \paper retains only its last five checkpoints, so nested \paper \M{1} was not evaluated at 22 epochs and this comparison does not measure relative convergence speed. A dedicated \M{1} model therefore matches nested \M{1} in planning across three seeds with slightly better realism after a second full training run that stopping at 22 epochs does not replace, while offering no full-granularity synthesis mode: at 49 NFEs nested \paper \M{4} remains 0.6--1.4 \fid ahead at equal \lpips and DINO (within 0.002; primary runs, one training seed for the standalone model). \paper isolates capacity within one training run and serves both configurations from one checkpoint.

\paragraph{Sub-\M{1} Capacity Probe (Standalone \M{0}).}
To push beyond the $0.65\times$ capacity floor of \paper \M{1}, we trained a standalone \M{0} model (57.4\,M parameters, 0.19$\times$ MACs; 6 trunk blocks of hidden size 384 and 2 head blocks of hidden size 1{,}024) under the identical recipe. At 50 epochs, \M{0} matches nested \paper \M{1} in planning on \recon (1.26 vs.\ 1.27 \ate\ at four NFEs) and \scand (1.01) and reproduces the 49-to-4 NFE step-count gain ($-0.14$\sig\ \recon, $-0.10$\sig\ \sacson), yet degrades \sacson planning (2.92 vs.\ 2.75 at four NFEs, $+0.17$\sig, \cref{tab:s-pooled}) and has the worse mean than nested \paper \M{1} in all 36 generation cells (\lpips, DINO, and \fid at 4\,s and 16\,s, at 49 and four NFEs, on three datasets; e.g., 20.9 vs.\ 19.1 \fid on 4\,s \sacson at 49 NFEs, \cref{tab:s-gen-full,tab:s-gen-horizon}). Its intermediate 25-epoch checkpoint has higher \sacson \ate\ at every step count (e.g., 3.14 vs.\ 2.92 at four NFEs) and higher \fid in each of the 12 \fid cells (e.g., 28.2 vs.\ 20.9 on 4\,s \sacson at 49 NFEs), while its \recon and \scand \ate\ differ from the 50-epoch values by at most 0.02 and 0.04; the second half of the schedule therefore mainly improves \sacson planning and \fid. While sub-$0.65\times$ compression suffices for open-terrain dynamics, indoor crowd planning and high-fidelity synthesis demand the capacity of \M{1} or \M{4}.

\begin{table*}[tp]
\makeatletter\def\@captype{figure}\makeatother\centering
\includegraphics[width=\textwidth]{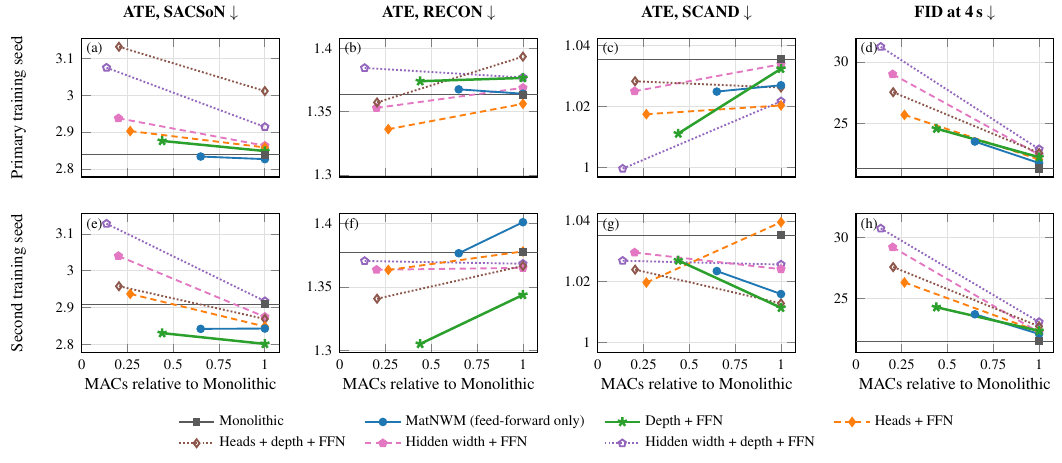}
\caption{\textbf{Quality against compute for the nested axes.} Each line joins \M{1} (left end) and \M{4} (right end) of one model; the horizontal line marks \mono. (a)--(c), (e)--(g): planning \ate\ at $S{=}50$ (49 NFEs), mean of evaluation seeds 0--2 on 500 items per dataset. (d), (h): \fid at 4\,s and $S{=}50$, mean of the three datasets (one evaluation seed). Top row: primary training seed; bottom row: second training seed. Lower is better.}
\label{fig:s-axes-compute}
\end{table*}

\begin{table*}[t]
\centering\scriptsize
\caption{\textbf{Nesting further axes on top of the feed-forward layers: \M{1} and \M{4} on two training seeds.} MACs relative to \mono; ms: compiled latency per forward pass at batch 120, measured once per training run and averaged over the two runs. Per training seed, over 12 planning cells (\recon, \sacson, \scand at $S\in\{50,5,2\}$, i.e., 49, 4, and 1 NFEs, and the two-draw $2{\times}S{=}3$ planner; $500$ items, evaluation seeds 0--2): \emph{worse}, cells in which the 95\,\% item-bootstrap interval (5{,}000 resamples) of the paired \ate\ difference against \paper at the same granularity and training seed lies above zero; \emph{non-inf.}, cells in which the upper interval bound against \mono of the same training seed is below 5\,\% of the \mono \ate\ at $S{=}50$. $\Delta$\fid: \fid at 4\,s and $S{=}50$ minus that of \paper at the same granularity and training seed, mean over the three datasets (one evaluation seed). One training run per model and training seed.}
\label{tab:s-axes}
\setlength{\tabcolsep}{4pt}
\begin{tabular}{@{}ll cc ccc ccc@{}}
\toprule
 & & & & \multicolumn{3}{c}{Primary training seed} & \multicolumn{3}{c}{Second training seed} \\
\cmidrule(lr){5-7}\cmidrule(l){8-10}
Nested axes & & MACs & ms & worse & non-inf. & $\Delta$\fid & worse & non-inf. & $\Delta$\fid \\
\midrule
\tabinput tab/gen/supp_axes_body.tex
\bottomrule
\end{tabular}
\end{table*}

\begin{table*}[t]
\centering\scriptsize
\caption{\textbf{Intermediate granularities \M{2} and \M{3} of the models in \cref{tab:s-axes}} (same columns and training seeds; one evaluation seed for every entry, so the planning counts are noisier than in \cref{tab:s-axes}).}
\label{tab:s-axes-mid}
\setlength{\tabcolsep}{4pt}
\begin{tabular}{@{}ll cc ccc ccc@{}}
\toprule
 & & & & \multicolumn{3}{c}{Primary training seed} & \multicolumn{3}{c}{Second training seed} \\
\cmidrule(lr){5-7}\cmidrule(l){8-10}
Nested axes & & MACs & ms & worse & non-inf. & $\Delta$\fid & worse & non-inf. & $\Delta$\fid \\
\midrule
\tabinput tab/gen/supp_axes_mid_body.tex
\bottomrule
\end{tabular}
\end{table*}

\begin{table*}[tp]
\makeatletter\def\@captype{figure}\makeatother\centering
\includegraphics[width=\textwidth]{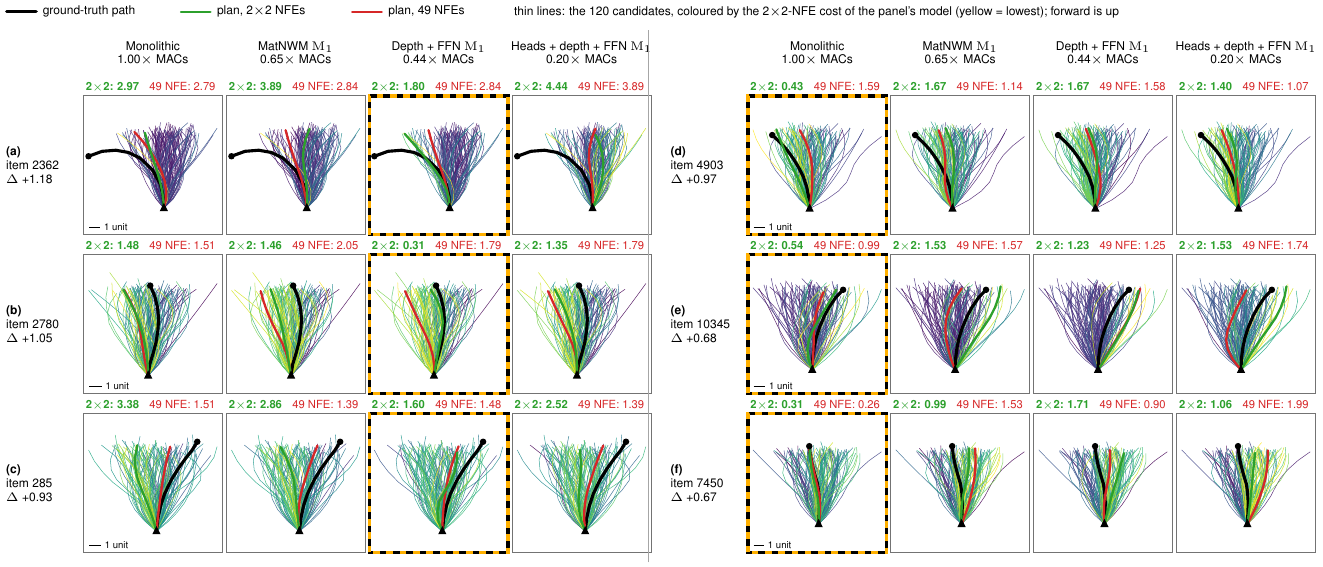}
\caption{\textbf{\cem plans on \recon with nested depth and heads} (second training seed, evaluation seed 0). The items are ranked, not random: selection as in \cref{fig:s-axes-gen}, applied to the \ate\ of the two-draw $2{\times}2$-NFE planner ($2{\times}S{=}3$) on the 500 items. (a)--(c): the three items that Depth + FFN \M{1} wins by the largest margin; (d)--(f): the three that \mono wins by the largest margin. All panels of a row show the same item and the same 120 candidates; the numbers above a panel are its \ate\ at $2{\times}2$ and at 49 NFEs.}
\label{fig:s-axes-plan}
\end{table*}

\begin{table*}[tp]
\makeatletter\def\@captype{figure}\makeatother\centering
\includegraphics[width=\textwidth]{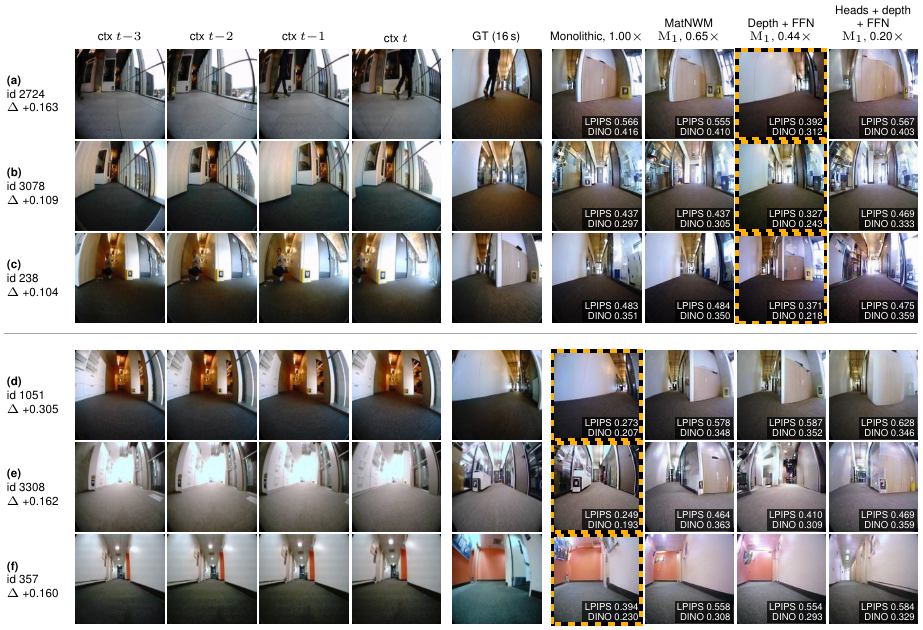}
\caption{\textbf{Direct 16\,s predictions on \sacson with nested depth and heads} ($S{=}50$; second training seed, evaluation seed 0). The items are ranked, not random. A model wins an anchor when its \lpips is the lowest of the four models; the margin $\Delta$ is its distance to the second-lowest. (a)--(c): the three of the 3{,}416 anchors that Depth + FFN \M{1} wins by the largest margin; (d)--(f): the three that \mono wins by the largest margin. Columns: four context frames, the ground truth, and the four models with their MACs relative to \mono. Insets: \lpips and DINO token distance to the ground truth. Dashed frame: the model the row was selected for.}
\label{fig:s-axes-gen}
\end{table*}

\clearpage
\section{Nesting Beyond the Feed-Forward Layers}
\label{sec:s-axes}

The main paper lists as a limitation that \paper nests only the feed-forward layers, which bounds its smallest sub-network at 0.65$\times$ the MACs of \mono. We test whether this bound comes from nested training or from the choice of axis. Five further models keep the feed-forward nesting and additionally nest the number of blocks (depth), the attention heads, or the hidden width (the hidden size of the blocks, \cref{sec:s-impl}), alone or together with depth. Each is trained with two training seeds under the recipe and 50-epoch schedule of \paper. These models go beyond the FFN-only \paper of the main paper and change no claim of the main paper.

\paragraph{Protocol.}
Each model is compared with \paper at the same granularity and with \mono, both of the same training seed, over twelve planning cells (three datasets at four sampler settings) and three evaluation seeds for \M{1} and \M{4} (\cref{tab:s-axes}). \fid, latency, and the intermediate granularities \M{2} and \M{3} (\cref{tab:s-axes-mid}) use one evaluation seed. Two training runs per model are two draws and do not estimate the variation between training seeds.

\paragraph{Depth.}
Nesting depth lowers \M{1} from 0.65$\times$ to 0.44$\times$ the MACs of \mono without a measured planning loss: on each training seed, \M{1} is non-inferior to \mono in 12/12 cells and significantly worse than \paper \M{1} in 0/12, at a \fid cost of $+1.1$ (primary seed) and $+0.6$ (second seed) at 4\,s (\cref{fig:s-axes-compute}). It is the only one of the five models with this result on both training seeds. \cref{fig:s-axes-gen,fig:s-axes-plan} show the cases in which it gains most over the other models, next to those in which \mono gains most.

\paragraph{Attention Heads and Hidden Width.}
Nesting the attention heads reaches 0.27$\times$; its \M{1} is non-inferior in 12/12 cells and significantly worse than \paper \M{1} in 1/12 on each seed, at a \fid cost of $+2.2$ and $+2.6$. Combining heads with depth (0.20$\times$) is non-inferior in 8/12 cells on the primary training seed, where it misses the margin in all four \sacson cells, and in 12/12 on the second. This combination is therefore sensitive to the training seed; at \M{2} (one evaluation seed) it is again non-inferior in 8/12 cells on the primary seed and in 12/12 on the second. Nesting the hidden width costs the most: at 0.20$\times$ and 0.14$\times$, \M{1} is non-inferior in at most 9/12 cells on either seed and \fid rises by $+5.5$ to $+7.7$. We hypothesize that the normalization layers, which are shared across sub-networks of different hidden width, cause this loss; a normalization per hidden width is untested.

\section{Negative Results}
\label{sec:s-neg}
\begin{table}[t]
\centering
\scriptsize
\caption{\textbf{Negative results} (\sacson unless stated; lower is better; \ate\ at evaluation seed 0 so all variants share identical candidates). The anchored-source prior, which initializes the flow from a noised copy of the last context frame instead of pure noise, is worse at every granularity and budget and collapses at $S{=}5$ (4 NFEs). Screening the 120 candidates with \M{1} and re-scoring the best 24 with \M{4} ties the single-granularity planners at 1.24$\times$ the time of \paper \M{1} (1.16$\times$ that of \paper \M{4}). Because shared attention accounts for roughly 60\,\% of MACs, \M{1}'s 35\,\% MAC reduction yields a 6\,\% forward-pass latency reduction under explicit attention (184 vs.\ 197\,ms/sample at 49 NFEs, batch 120) and 10--15\,\% under PyTorch fused SDPA (74.5 vs.\ 87.7\,ms/sample). Three-seed comparisons for the standalone \M{1} model are in \cref{sec:s-nesting}.}
\label{tab:negatives}
\setlength{\tabcolsep}{3pt}
\resizebox{\linewidth}{!}{%
\begin{tabular}{@{}lcccc@{}}
\toprule
 & & \lpips & \fid & \ate\ (seed 0) \\
Variant & $S$\,/\,NFE & 4\,s & 4\,s & \sacson / \recon \\
\midrule
\paper \M{4} ($\sigma{=}1$, reference)       & 50\,/\,49 & 0.333 & 17.1 & 2.81 / 1.38 \\
Anchored source prior, $\sigma{=}0.5$       & 50\,/\,49 & 0.387 & 28.0 & 3.46 / 1.42 \\
Anchored source prior, $\sigma{=}0.5$       & 5\,/\,4   & 0.380 & 51.0 & -- \\
Two-stage planner \M{1}$\to$\M{4}, $k{=}24$ & 50\,/\,49 & --    & --   & 2.80 / 1.37 \\
Independently trained \M{1}, same budget    & 50\,/\,49 & 0.334 & 17.7 & 2.83 / 1.36 \\
\bottomrule
\end{tabular}}
\end{table}

Two variants did not survive our experiments (\cref{tab:negatives}). An anchored source prior that starts the flow from a noised copy of the last context frame ($\sigma{=}0.5$) instead of pure noise is worse at every granularity and budget ($+0.054$ \lpips at 49 NFEs and a collapse of \fid to 51.0 at four NFEs on \sacson; \ate\ 3.46 against 2.81). Because attention is shared across granularities, \M{1}'s 35\,\% MAC reduction yields a 6\,\% forward-pass latency reduction under explicit attention (184 vs.\ 197\,ms per sample at 49 NFEs, batch 120). Under homogeneous PyTorch fused SDPA it yields 10--15\,\% across $S\in[2,50]$ (182.4 vs.\ 214.7\,ms per batch-120 forward; 4.24 vs.\ 4.71\,GB peak CUDA memory), rising to 24\,\% under \texttt{torch.compile} ($100$ vs.\ $132$\,ms; \cref{tab:s-sdpa-shard}).

\paragraph{Position of the Monolithic Baseline.}
The monolithic baseline lies in the regime of the published models. Its 0.323 / 0.361 \lpips and 16.5 / 17.8 \fid on \sacson at 49 NFEs ($S{=}50$, 4 / 16\,s) are 0.012--0.020 \lpips and 1.4--1.9 \fid behind published \raenwm (0.303 / 0.349 and 15.1 / 15.9) and far ahead of \nwm (0.407 / 0.470), while its planning error (1.35, 2.85, 1.01 at seed 0; 1.36, 2.84, 1.04 over three seeds) is on par with or below published \raenwm (1.36, 2.91, 1.14), whereas unpaired 1\,B \nwm ($3\times 250{=}750$ DDPM steps/frame on full splits) reaches 1.13 on \recon vs.\ 4.12 on \sacson. We attribute the remaining synthesis gap to the data-preparation and block-level architectural differences listed in \cref{sec:s-impl}. Training-seed variation is too small to account for it: the second training seed (\cref{sec:s-seed1}) changes the three-dataset mean by at most 0.0009 \lpips and DINO distance and by 0.17--0.44 \fid across \mono, \M{4}, and \M{1} at 49 and four NFEs (\cref{tab:s-seed1-gen}), and on \sacson at 49 NFEs the seed-1 baseline reaches 0.323 / 0.360 \lpips and 16.7 / 17.5 \fid. The released \raenwm checkpoint has not been evaluated in our harness, so the published rows of Tables~1 and~2 are unpaired with ours.

\section{Second Training Seed}
\label{sec:s-seed1}

\begin{table*}[t]
\centering\scriptsize
\caption{\textbf{Planning with the second training seed.} \ate\ and translational \rpe\ (mean\,$\pm$\,sd over evaluation seeds 0--2, 500 items per dataset) of the final 50-epoch checkpoints of training seed 1, with the primary runs (training seed 0) for reference. $S$ grid points execute $S{-}1$ NFEs (in parentheses); $2{\times}3$ averages two $S{=}3$ noise draws offline and is scored by \ate\ only (seed-0 two-draw \mono in \cref{tab:s-seeds}). The sd covers evaluation seeds only.}
\label{tab:s-seed1-plan}
\setlength{\tabcolsep}{5pt}
\begin{tabular}{@{}lc ccc ccc@{}}
\toprule
 & & \multicolumn{3}{c}{\ate} & \multicolumn{3}{c}{\rpe} \\
\cmidrule(lr){3-5}\cmidrule(l){6-8}
Model & $S$ (NFEs) & \recon & \sacson & \scand & \recon & \sacson & \scand \\
\midrule
\tabinput tab/gen/supp_t2_seed1_plan_body.tex
\bottomrule
\end{tabular}
\end{table*}

\begin{table*}[t]
\centering\scriptsize
\caption{\textbf{Direct prediction with the second training seed.} \lpips, DINO distance, and \fid at 4\,s and 16\,s on the full test splits (4{,}262 / 3{,}416 / 1{,}883 samples) for the primary runs (training seed 0) and the seed-1 runs at $S{=}50$ (49 NFEs) and $S{=}5$ (4 NFEs), with the unweighted mean over the three datasets. Four decimals are shown because the two seeds differ only in the third or fourth.}
\label{tab:s-seed1-gen}
\setlength{\tabcolsep}{3.6pt}\renewcommand{\arraystretch}{0.95}
\begin{tabular}{@{}llc cc cc cc cc cc cc@{}}
\toprule
 & & & \multicolumn{6}{c}{Training seed 0 (primary runs)} & \multicolumn{6}{c}{Training seed 1} \\
\cmidrule(lr){4-9}\cmidrule(l){10-15}
 & & & \multicolumn{2}{c}{\lpips} & \multicolumn{2}{c}{DINO} & \multicolumn{2}{c}{\fid} & \multicolumn{2}{c}{\lpips} & \multicolumn{2}{c}{DINO} & \multicolumn{2}{c}{\fid} \\
Dataset & Model & $S$ (NFEs) & 4\,s & 16\,s & 4\,s & 16\,s & 4\,s & 16\,s & 4\,s & 16\,s & 4\,s & 16\,s & 4\,s & 16\,s \\
\midrule
\tabinput tab/gen/supp_t2_seed1_gen_body.tex
\bottomrule
\end{tabular}
\end{table*}

\begin{table*}[t]
\centering\scriptsize
\caption{\textbf{Paired \ate\ differences for the second training seed} (first minus second configuration; negative favors the first). Per-item differences are averaged over evaluation seeds 0--2 and bootstrapped over the 500 items of each dataset (1{,}000 resamples, 95\,\% interval; \sig: excludes zero). Blocks~(a)--(c) compare configurations within one training seed, block~(d) pairs the two training seeds of the same model and $S$, and block~(e) gives the seed-1 minus seed-0 difference of four contrasts.}
\label{tab:s-seed1-ci}
\setlength{\tabcolsep}{6pt}\renewcommand{\arraystretch}{0.95}
\begin{tabular}{@{}l lll@{}}
\toprule
Comparison & \recon & \sacson & \scand \\
\midrule
\tabinput tab/gen/supp_t2_seed1_ci_body.tex
\bottomrule
\end{tabular}
\end{table*}

\paper and the monolithic baseline were each trained a second time with training seed 1 (provenance in \cref{sec:s-impl}), and their final 50-epoch EMA checkpoints were evaluated under the unchanged protocol: planning on the same 500 items per dataset at evaluation seeds 0--2 for $S\in\{50,5,3,2\}$ and for the two-draw $S{=}3$ planner (two noise draws averaged offline, as in \cref{tab:s-seeds}), and direct prediction on all 9{,}561 test samples at $S\in\{50,5\}$. \cref{tab:s-seed1-plan} lists \ate\ and \rpe, \cref{tab:s-seed1-gen} generation, and \cref{tab:s-seed1-ci} the paired \ate\ differences (per-item differences averaged over the three evaluation seeds, then bootstrapped over items with 1{,}000 resamples).

\paragraph{Planning Within Training Seed 1.}
The planning comparisons of the main paper reproduce on the second seed. \M{1} minus \mono at a matched step count is not significant on \recon at any $S$ ($|\Delta|\le 0.019$) or on \sacson at $S\le 5$; \M{1} is ahead on \sacson at $S{=}50$ ($-0.065$\sig, upper bound $-0.0005$) and on \scand at $S\in\{5,3,2\}$ ($-0.033$\sig, $-0.075$\sig, $-0.091$\sig). Every upper bound of block~(a), including the two-draw planner, lies below 5\,\% of the seed-1 \monoat{50} error (0.069, 0.145, and 0.052). The seed-0 deficit of \M{1} on \sacson at $S\le 3$ ($+0.053$ at $S{=}3$ and $+0.102$\sig\ at $S{=}2$) does not recur ($+0.002$ and $-0.042$, neither significant). Reducing 49 to four NFEs lowers \ate\ on \recon / \sacson by $-0.087$\sig\ / $-0.100$\sig\ for \M{1} and $-0.069$\sig\ / $-0.128$\sig\ for \mono (seed 0: $-0.101$\sig\ / $-0.087$\sig\ and $-0.067$\sig\ / $-0.081$\sig). The two-draw planner $2{\times}\at{1}{3}$ beats \monoat{50} of the same seed by $-0.138$\sig, $-0.249$\sig, and $-0.037$\sig\ on \recon, \sacson, and \scand, and the second draw helps both models on \sacson ($-0.094$\sig\ for \M{1}, $-0.075$\sig\ for \mono) and on neither of the other datasets. On \scand, \mono worsens at two and one NFEs ($+0.029$\sig\ and $+0.038$\sig\ against 49 NFEs).

\paragraph{Variation Between Training Seeds.}
Block~(d) of \cref{tab:s-seed1-ci} pairs the two training seeds of one model on identical items, candidates, and sampler noise. For \M{1}, 14 of 15 intervals include zero; the exception is \recon at $S{=}5$ ($+0.024$\sig, lower bound $+0.001$). For \mono, no difference on \recon or \scand is significant, but on \sacson the seed-1 model is worse at $S{=}50$ ($+0.071$\sig), $S{=}3$ ($+0.090$\sig\ [$+0.023$, $+0.152$]), and $S{=}2$ ($+0.078$\sig), with $+0.024$ at $S{=}5$. This is the shift of up to 0.09 \ate\ cited in the limitations of the main paper; the \sacson gaps among \mono, \M{4}, and \M{1} at matched NFEs in Table~2 (at most 0.05) lie within it. The contrast of 1-NFE \M{1} against 49-NFE \mono on \sacson consequently differs between seeds ($-0.027$ at seed 0, $-0.164$\sig\ at seed 1; difference $-0.137$\sig\ [$-0.235$, $-0.048$], block~(e)), which combines the $+0.071$\sig\ shift of \monoat{50} with a non-significant $-0.066$ shift of \at{1}{2}. The other three contrasts of block~(e) do not differ significantly between seeds on any dataset; the largest difference is four-NFE \M{1} against 49-NFE \mono on \sacson ($-0.076$ [$-0.157$, $+0.002$]). The two seeds of each model also differ in micro-batch size and attention kernel during training (\cref{sec:s-impl}), so block~(d) measures their combined effect, and two runs per model do not estimate a between-seed variance.

\paragraph{Generation.}
In the three-dataset mean, seed 1 stays within 0.0009 \lpips and DINO distance of seed 0 for all three models at both schedules, and its \fid is higher by 0.17--0.44 in all 12 cells (\cref{tab:s-seed1-gen}). Per dataset the differences are at most 0.0027 in \lpips and DINO distance and between $-0.35$ and $+1.20$ in \fid, the largest on \scand. The ordering \mono, \M{4}, \M{1} of the three-dataset mean holds for every metric at both schedules on both seeds (4\,s \fid at $S{=}50$: 21.54 / 22.11 / 23.75 at seed 1 and 21.30 / 21.74 / 23.51 at seed 0), and the \fid gap between \M{1} and \mono (2.2--4.6 on both seeds) is at least five times the largest between-seed difference.

\section{Same-Recipe Deterministic Regressor}
\label{sec:s-reg}

\begin{table*}[t]
\centering\scriptsize
\caption{\textbf{Same-recipe DINO-Reg planning.} Top: \ate\ and translational \rpe\ (mean\,$\pm$\,sd over evaluation seeds 0--2, 500 items per dataset) of the regressor after 25, 47, and 50 epochs of one training run and of three reference planners from the primary runs. Bottom: paired differences (first minus second, epochs in parentheses; per-item mean over the three evaluation seeds, 1{,}000-resample bootstrap over items, 95\,\% interval; \sig: excludes zero). The regressor uses one network evaluation per frame and no sampler noise.}
\label{tab:s-reg-plan}
\setlength{\tabcolsep}{5pt}\renewcommand{\arraystretch}{0.95}
\begin{tabular}{@{}l ccc ccc@{}}
\toprule
 & \multicolumn{3}{c}{\ate} & \multicolumn{3}{c}{\rpe} \\
\cmidrule(lr){2-4}\cmidrule(l){5-7}
Planner & \recon & \sacson & \scand & \recon & \sacson & \scand \\
\midrule
\tabinput tab/gen/supp_t2_reg_plan_body.tex
\bottomrule
\end{tabular}\\[4pt]
\begin{tabular}{@{}l lll@{}}
\toprule
Paired difference [95\,\% CI] & \recon & \sacson & \scand \\
\midrule
\tabinput tab/gen/supp_t2_reg_plan_ci_body.tex
\bottomrule
\end{tabular}
\end{table*}

\begin{table*}[t]
\centering\scriptsize
\caption{\textbf{Same-recipe DINO-Reg by trajectory curvature.} Three-seed \ate\ on the quartiles of \cref{tab:s-turns} (125 items each; range of $\max_{t\le 8}|y_t|$ in step-size units) for the \cem prior mean, four reference planners from the primary runs ($@S$: grid size; $2{\times}@3$: two-draw \at{1}{3}), and the regressor after 25, 47, and 50 epochs, followed by paired differences of the 50-epoch regressor with 95\,\% bootstrap intervals (\sig: excludes zero).}
\label{tab:s-reg-turns}
\setlength{\tabcolsep}{3pt}
\resizebox{\textwidth}{!}{%
\begin{tabular}{@{}ll c c ccc ccc lll@{}}
\toprule
 & & & \mono & \multicolumn{3}{c}{\paper \M{1}} & \multicolumn{3}{c}{DINO-Reg (epochs)} & \multicolumn{3}{c}{Paired $\Delta$ of DINO-Reg at 50 epochs [95\,\% CI]} \\
\cmidrule(lr){4-4}\cmidrule(lr){5-7}\cmidrule(lr){8-10}\cmidrule(l){11-13}
Dataset & Quartile ($\max|y|$) & Prior & $@50$ & $@5$ & $@2$ & $2{\times}@3$ & 25 & 47 & 50 & $-$\,\at{1}{5} & $-$\,\at{1}{2} & $-$\,$2{\times}\at{1}{3}$ \\
\midrule
\tabinput tab/gen/supp_t2_reg_turns_body.tex
\bottomrule
\end{tabular}}
\end{table*}

\begin{table*}[t]
\centering\scriptsize
\caption{\textbf{Same-recipe DINO-Reg generation.} One-shot direct prediction at 4\,s and 16\,s on the full test splits (one network evaluation, no noise) after 25, 47, and 50 epochs, against \paper \M{1} at $S{=}50$ (49 NFEs). \emph{Mean}: unweighted over the three datasets.}
\label{tab:s-reg-gen}
\setlength{\tabcolsep}{6pt}
\begin{tabular}{@{}ll cc cc cc@{}}
\toprule
 & & \multicolumn{2}{c}{\lpips} & \multicolumn{2}{c}{DINO} & \multicolumn{2}{c}{\fid} \\
\cmidrule(lr){3-4}\cmidrule(lr){5-6}\cmidrule(l){7-8}
Dataset & Model & 4\,s & 16\,s & 4\,s & 16\,s & 4\,s & 16\,s \\
\midrule
\tabinput tab/gen/supp_t2_reg_gen_body.tex
\bottomrule
\end{tabular}
\end{table*}

\begin{table*}[t]
\centering\scriptsize
\caption{\textbf{One-shot against chained token error of the same-recipe DINO-Reg} on a seeded subset of 1{,}024 anchors per dataset. Token DINO distance (\emph{Tok.\ D.}) and latent MSE of a one-shot prediction (\emph{1-shot}) and of the autoregressive rollout frame nearest in time (\emph{chain}); $\Delta$: chained minus one-shot Tok.\ D.; NFEs per frame in parentheses. Per-cell 95\,\% interval half-widths of Tok.\ D.\ are at most 0.005, and cells are compared without a paired test. On \recon, the one-shot cells of \M{1} at $S{=}50$ cover 1{,}017 valid anchors and the chained cells of the 25-epoch regressor 1{,}023.}
\label{tab:s-reg-tok}
\setlength{\tabcolsep}{3.2pt}
\begin{tabular}{@{}l ccc cc ccc cc ccc cc@{}}
\toprule
 & \multicolumn{5}{c}{\recon} & \multicolumn{5}{c}{\sacson} & \multicolumn{5}{c}{\scand} \\
\cmidrule(lr){2-6}\cmidrule(lr){7-11}\cmidrule(l){12-16}
 & \multicolumn{3}{c}{Tok.\ D.} & \multicolumn{2}{c}{Lat.\ MSE} & \multicolumn{3}{c}{Tok.\ D.} & \multicolumn{2}{c}{Lat.\ MSE} & \multicolumn{3}{c}{Tok.\ D.} & \multicolumn{2}{c}{Lat.\ MSE} \\
\cmidrule(lr){2-4}\cmidrule(lr){5-6}\cmidrule(lr){7-9}\cmidrule(lr){10-11}\cmidrule(lr){12-14}\cmidrule(l){15-16}
Model, $S$ (NFEs) & 1-shot & chain & $\Delta$ & 1-shot & chain & 1-shot & chain & $\Delta$ & 1-shot & chain & 1-shot & chain & $\Delta$ & 1-shot & chain \\
\midrule
\tabinput tab/gen/supp_t2_reg_tok_body.tex
\bottomrule
\end{tabular}
\end{table*}

Section~5.4 of the main paper argues that one NFE returns approximately the conditional mean, and Table~2 lists a regressor that targets this mean directly (\emph{Same-recipe DINO-Reg}). The regressor is the 315.7\,M monolithic architecture trained with the recipe of \cref{sec:s-impl} (training seed 0; same data, optimizer, 50-epoch schedule, and EMA), with the flow-matching loss replaced by an L2 regression onto the normalized target tokens: the network receives a zero tensor in place of the noisy latent together with a constant flow time, and its output is read as the predicted tokens. A prediction costs one network evaluation per frame and uses no sampler noise (960 evaluations per decision), so the three evaluation seeds re-draw only the 120 candidates. Table~2 reports the final 50-epoch checkpoint; to follow training we also evaluate the EMA checkpoints of the same run after 25 and 47 epochs. The reference planners are the primary \paper \M{1} at $S{=}2$ and $S{=}5$ (one and four NFEs) and \mono at $S{=}50$ (49 NFEs).

\paragraph{Planning and Training Progression.}
At 50 epochs the regressor reaches $2.234\pm0.005$, $3.090\pm0.033$, and $1.223\pm0.024$ \ate\ on \recon, \sacson, and \scand, with \rpe\ of $0.544\pm0.002$, $0.755\pm0.004$, and $0.318\pm0.005$ (\cref{tab:s-reg-plan}). It is worse than 1-NFE \M{1} by $+0.985$\sig, $+0.280$\sig, and $+0.227$\sig, than four-NFE \M{1} by $+0.967$\sig, $+0.342$\sig, and $+0.213$\sig, and than 49-NFE \mono by $+0.871$\sig, $+0.253$\sig, and $+0.185$\sig; \rpe\ is worse in all nine comparisons ($+0.030$\sig\ to $+0.181$\sig). On \recon and \scand its \ate\ also exceeds that of the \cem prior mean (1.64 and 0.83, \cref{tab:s-calib}). Planning error rises during training. Within the run, \recon \ate\ goes from 1.753 at 25 epochs to 2.198 at 47 and 2.234 at 50, a paired increase of $+0.480$\sig\ [$+0.414$, $+0.545$] from 25 to 50 epochs and of $+0.036$\sig\ over the last three; \sacson rises by $+0.126$\sig\ and \scand is unchanged ($-0.006$). One-shot generation improves over the same epochs (\cref{tab:s-reg-gen}), so longer training within this recipe does not close the planning gap.

\paragraph{Curvature.}
\cref{tab:s-reg-turns} stratifies the same items by the quartiles of \cref{tab:s-turns}. On \recon the gap to four-NFE \M{1} grows with curvature, from $+0.134$\sig\ on Q1 to $+0.539$\sig, $+1.214$\sig, and $+1.983$\sig\ on Q2--Q4, and the regressor exceeds the \cem prior mean in every quartile (3.686 against 2.68 on Q4). Training leaves Q1 nearly unchanged (1.232, 1.205, and 1.210 at 25, 47, and 50 epochs) and raises Q4 from 2.786 to 3.638 and 3.686. On \sacson and \scand the regressor is worse in every quartile without a monotone trend. All 36 quartile intervals against four-NFE \M{1}, 1-NFE \M{1}, and $2{\times}\at{1}{3}$ exclude zero. A predictor that averages over turn directions would produce the \recon pattern; we did not test this mechanism directly.

\paragraph{Generation and Token-Space Error.}
Decoded predictions of the regressor are far from those of the flow models: the three-dataset mean 4\,s / 16\,s \fid is 103.2 / 123.5 against 23.5 / 25.6 for \M{1} at 49 NFEs (\cref{tab:s-reg-gen}). At 4\,s its \fid (115.5, 98.8, and 95.2 on \recon, \sacson, and \scand) and \lpips (0.46, 0.39, and 0.48) are nevertheless below those of 1-NFE \M{1} (172.5, 142.0, and 169.7; 0.51, 0.44, and 0.55; \cref{tab:s-gen-controls}), so the regressor is the better single-evaluation predictor and the worse planner. \cref{tab:s-reg-tok} compares one-shot and chained token error on a seeded subset of 1{,}024 anchors per dataset. In one shot at 2\,s the regressor has a token DINO distance of 0.175 against 0.178 for 1-NFE \M{1} on \recon and 0.133 against 0.146 on \sacson, and is higher on \scand (0.274 against 0.208). After nine chained steps (2.25\,s) it is higher on \recon (0.228 against 0.212) and the \sacson margin closes (0.178 against 0.176, overlapping intervals). The increase from one-shot to chained error is $+0.053$ / $+0.045$ / $+0.050$ for the regressor against $+0.035$ / $+0.030$ / $+0.040$ for 1-NFE \M{1} and $+0.031$ / $+0.019$ / $+0.019$ at 49 NFEs. No per-anchor paired test was run on these cells; they are compared through their per-cell 95\,\% intervals (half-width at most 0.005). On \scand the one-shot error of the regressor itself increases with training (2\,s latent MSE of 0.559, 0.614, and 0.621 at 25, 47, and 50 epochs). The one-shot accuracy of the regressor on \recon and \sacson therefore does not carry over to the chained rollouts that planning scores.

\section{Additional Qualitative Results}
\label{sec:s-qual}

The items in this section are ranked by a stated rule and are not random draws. \cref{fig:s-roll-sacson,fig:s-roll-recon,fig:s-roll-scand} show, for each dataset, the six test anchors with the largest late-horizon \lpips gap of the \M{1} rollout over the \M{4} rollout among anchors on which \M{4} is accurate, the worst cases for \M{1} relative to \M{4}, and add direct predictions of \M{1} at 49, 4, and 2 NFEs (few-NFE autoregressive rollouts were not run). \cref{fig:s-cem-recon,fig:s-cem-sacson} show the Q4 planning items with the largest three-seed \ate\ reduction from 49-NFE \M{1} to the $2{\times}2$-NFE planner, and \cref{fig:s-cem-counter} the three Q4 items per dataset with the largest increase; \cref{fig:s-cem-qual} is the random draw from the same quartile.

\clearpage

\begin{figure*}[p]
\centering
\RolloutPage{sacson}
\caption{\textbf{Ranked rollout panels on \sacson: the worst cases for \M{1} relative to \M{4}.} The items are ranked, not random. Among the 1{,}024 test anchors per dataset with stored \M{1} rollout frames, we keep those on which the 49-NFE \M{4} rollout has late \lpips (mean over the stored frames from 12.25 to 16\,s) within its best 40\,\%, early \lpips (3.25--5.25\,s) within its best 50\,\%, and late \lpips at most 0.01 above the monolithic model's (164 / 163 / 208 anchors on \sacson{} / \recon{} / \scand), and sort them by the late-\lpips gap of \M{1} over \M{4}. Panels (a)--(f) are ranks 1--6; (e) is the item of Fig.~1a. Rows: ground truth, 4\,fps autoregressive rollouts of \M{4} and \M{1} at 49 NFEs, and direct prediction of \M{1} at 49, 4, and 2 NFEs. Columns give the time after the last context frame. The ground-truth and rollout rows show the stored rollout frames at 4.25\,s and 8.25\,s, the direct-prediction rows show their 4\,s and 8\,s targets one frame earlier, and the 16\,s column is the same instant in every row. Model rollout frames are resized from $256^2$ to $224^2$ pixels. At 16\,s the \M{4} rollout keeps the scene layout of the ground truth in all six panels and the \M{1} rollout has drifted to a different layout, least clearly in (d). In direct prediction \M{1} keeps the coarse geometry at 4 NFEs and loses texture at 2 NFEs, most visibly in (b) and (d).}
\label{fig:s-roll-sacson}
\end{figure*}

\begin{figure*}[p]
\centering
\RolloutPage{recon}
\caption{\textbf{Ranked rollout panels on \recon.} The items are ranked, not random: ranks 1--6 under the rule of \cref{fig:s-roll-sacson} (largest late-\lpips gap of the \M{1} rollout over the \M{4} rollout among the 163 anchors on which \M{4} is accurate), the worst cases for \M{1} relative to \M{4}. Rows, columns, and frame timing as in \cref{fig:s-roll-sacson}. At 16\,s the \M{4} rollout keeps the ground-truth layout and the \M{1} rollout drifts in (a), (b), (d), and (f). The panels do not show this in (c), where neither rollout reproduces the distance to the building, and in (e), where both rollouts retain the tree and the building. In direct prediction \M{1} keeps the coarse geometry at 4 NFEs in every panel. At 2 NFEs thin structures fade in (a) and (b), and (c)--(f) stay close to the 4-NFE frames at this scale.}
\label{fig:s-roll-recon}
\end{figure*}

\begin{figure*}[p]
\centering
\RolloutPage{scand}
\caption{\textbf{Ranked rollout panels on \scand.} The items are ranked, not random: ranks 1--6 under the rule of \cref{fig:s-roll-sacson} (largest late-\lpips gap of the \M{1} rollout over the \M{4} rollout among the 208 anchors on which \M{4} is accurate), the worst cases for \M{1} relative to \M{4}. Rows, columns, and frame timing as in \cref{fig:s-roll-sacson}. The ranked gaps are smaller on \scand than on the other two datasets and are visible at 16\,s only in (c), where the \M{1} rollout loses the red-paved lane that \M{4} keeps, and in (b), where it loses the blue fence. In (a), (d), (e), and (f) both rollouts keep the ground-truth layout and the panels do not separate the two granularities. In direct prediction \M{1} keeps the coarse geometry at 4 NFEs. At 2 NFEs the pedestrians in (c) smear and the pavement in (d) loses texture, and the other panels stay close to the 4-NFE frames at this scale.}
\label{fig:s-roll-scand}
\end{figure*}

\begin{figure*}[p]
\centering
\includegraphics[width=0.93\textwidth]{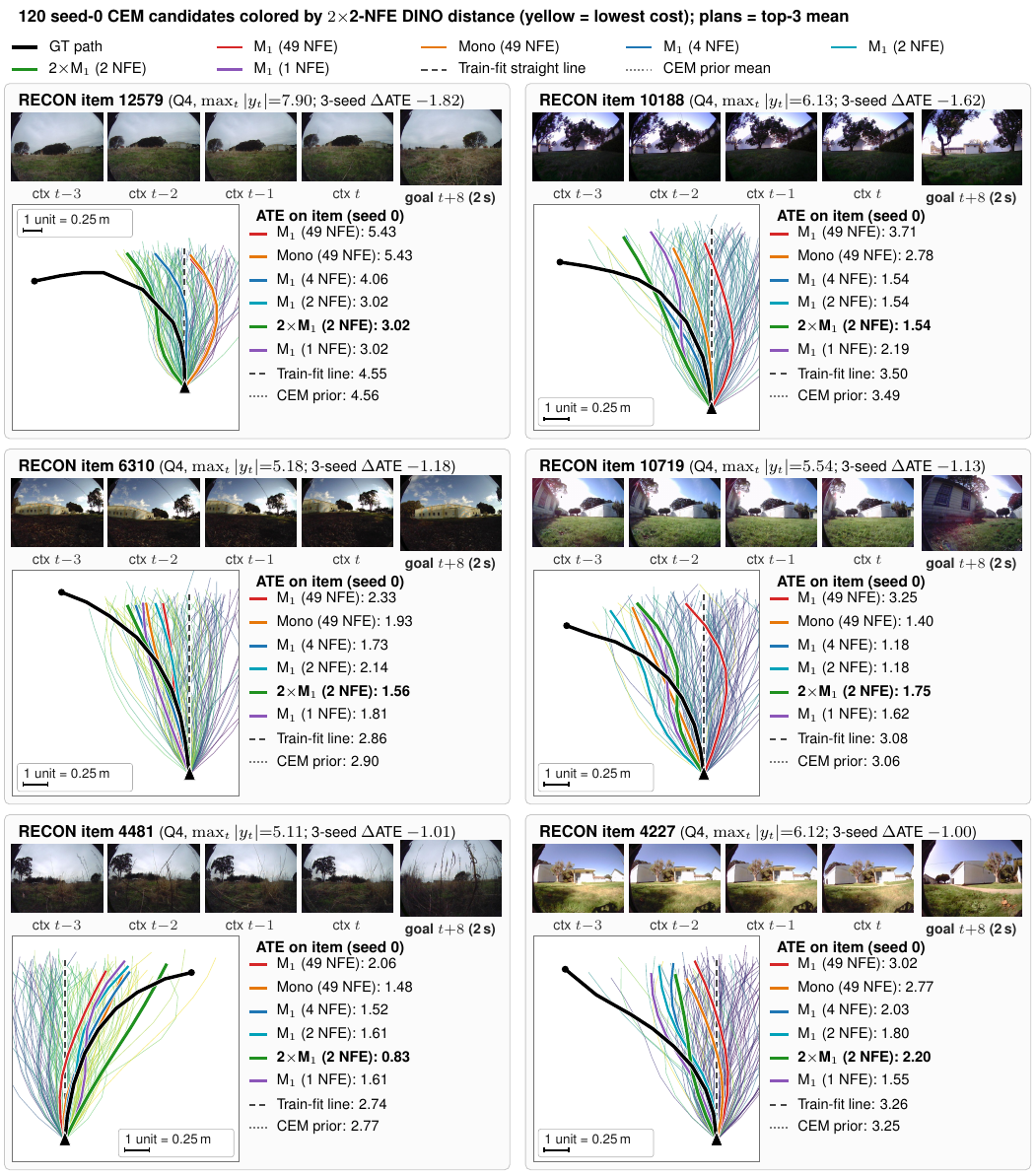}
\caption{\textbf{Ranked \cem planning panels on \recon: largest gains of the $2{\times}2$-NFE planner.} The items are ranked, not random; \cref{fig:s-cem-qual} is the random draw. Among the 125 Q4 items of \cref{tab:s-turns} we take those on which two averaged 2-NFE draws of \M{1} ($2{\times}2$ NFEs; $2{\times}\mathrm{M}_1$ in the legend) have lower \ate\ than 49-NFE \M{1} in each of the three evaluation seeds, exclude the items of \cref{fig:s-cem-qual}, and show the six with the largest three-seed mean \ate\ reduction in reading order (3-seed $\Delta$\ate\ in each title). Panels show seed 0 in the style of \cref{fig:s-cem-qual}. Over the whole quartile the three-seed mean \ate\ falls from 1.95 to 1.65 (\cref{tab:s-turns}), 97 of the 125 items have a positive three-seed mean gain, and 47 gain in all three seeds. On seed 0 the six items go from 5.43 to 3.02, 3.71 to 1.54, 2.33 to 1.56, 3.25 to 1.75, 2.06 to 0.83, and 3.02 to 2.20 \ate. In every panel the $2{\times}2$-NFE plan bends further toward the recorded path than the 49-NFE plan.}
\label{fig:s-cem-recon}
\end{figure*}

\begin{figure*}[p]
\centering
\includegraphics[width=0.93\textwidth]{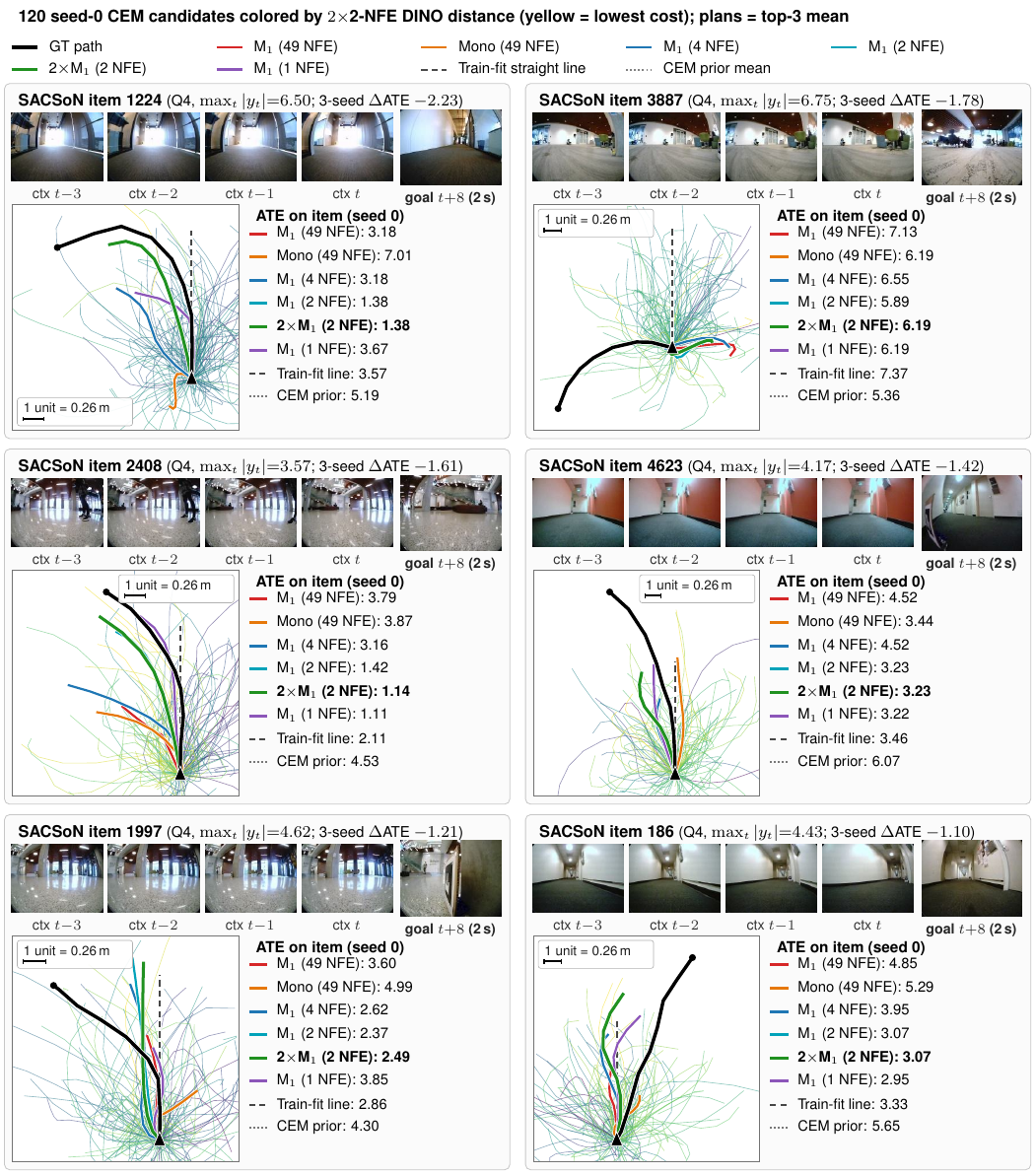}
\caption{\textbf{Ranked \cem planning panels on \sacson: largest gains of the $2{\times}2$-NFE planner.} The items are ranked, not random; \cref{fig:s-cem-qual} is the random draw. Selection as in \cref{fig:s-cem-recon}: among the 125 Q4 items (default speed prior), the six with the largest three-seed mean \ate\ reduction from 49-NFE \M{1} to $2{\times}2$ NFEs among items that gain in each of the three seeds, excluding the items of \cref{fig:s-cem-qual}; seed 0 is shown. Over the whole quartile the three-seed mean \ate\ falls from 3.87 to 3.74 (\cref{tab:s-turns}), 76 of the 125 items have a positive three-seed mean gain, and 26 gain in all three seeds. On seed 0 the six items go from 3.18 to 1.38, 7.13 to 6.19, 3.79 to 1.14, 4.52 to 3.23, 3.60 to 2.49, and 4.85 to 3.07 \ate. As in \cref{fig:s-cem-qual}, the short proposal fan of the default speed prior leaves a large residual error where the recorded path is long (item 3887).}
\label{fig:s-cem-sacson}
\end{figure*}

\begin{figure*}[p]
\centering
\includegraphics[width=0.93\textwidth]{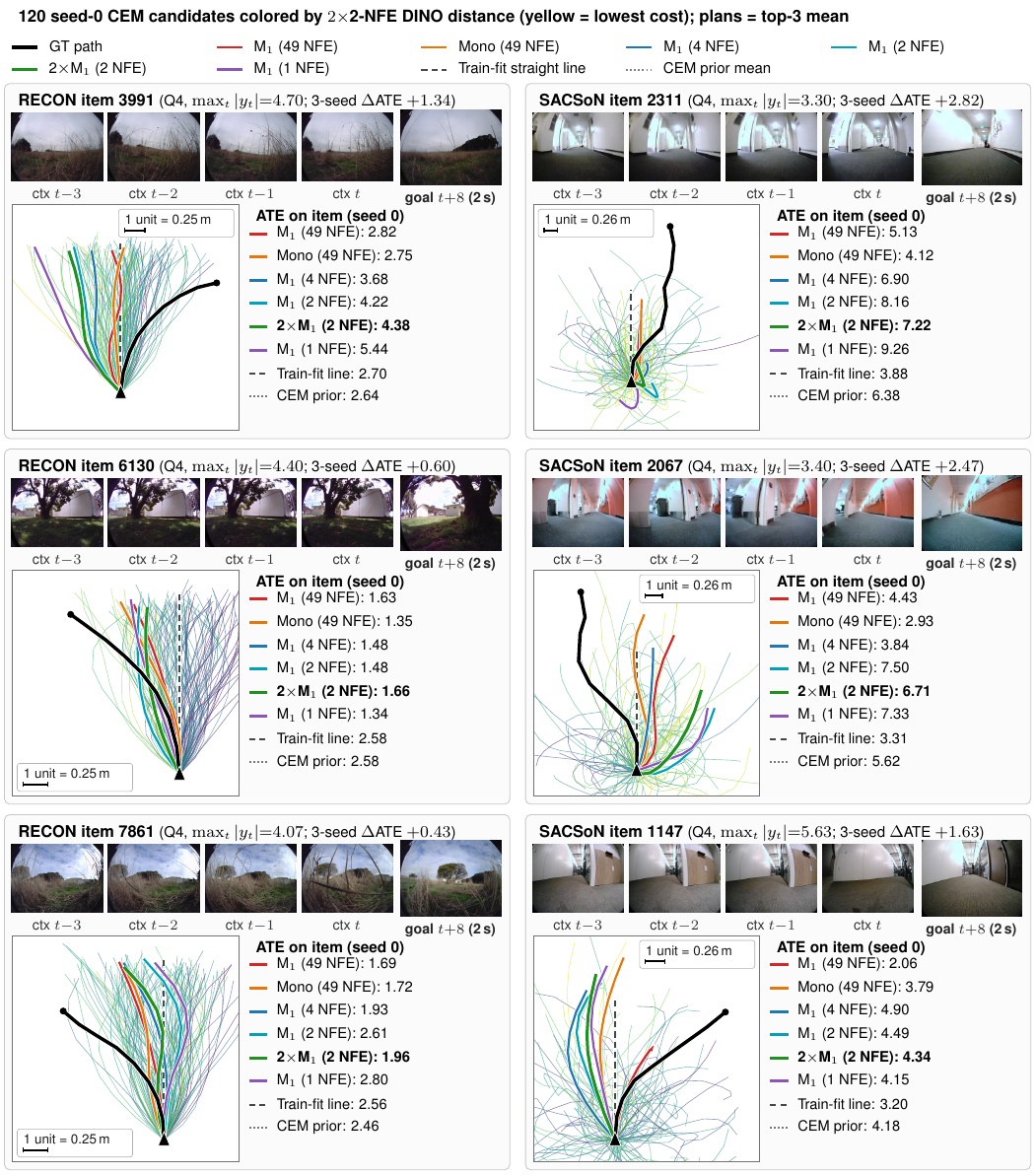}
\caption{\textbf{Ranked \cem planning panels: counter-examples.} The items are ranked, not random: the three Q4 items per dataset with the largest three-seed mean \ate\ increase from 49-NFE \M{1} to $2{\times}2$ NFEs (left column: \recon; right column: \sacson; seed 0 shown, style of \cref{fig:s-cem-qual}). All six are worse under $2{\times}2$ NFEs in each of the three seeds. On seed 0 they go from 2.82 to 4.38, 1.63 to 1.66, and 1.69 to 1.96 \ate\ on \recon and from 5.13 to 7.22, 4.43 to 6.71, and 2.06 to 4.34 on \sacson. Over the whole quartile, 28 of the 125 \recon items and 49 of the 125 \sacson items have a negative three-seed mean gain, and 11 and 8 lose in all three seeds.}
\label{fig:s-cem-counter}
\end{figure*}

\end{document}